\documentclass[preprint,12pt]{elsarticle}

 \usepackage{float} 
\usepackage{graphicx}
\usepackage{algorithm, algorithmic} 
\usepackage{amsmath}
\usepackage{amssymb}
\usepackage{dsfont}
\usepackage[english]{babel}
\usepackage[latin1]{inputenc}
\usepackage[T1]{fontenc}
\usepackage{multicol}
\usepackage{subcaption}

\usepackage{array}

\usepackage[table]{xcolor}

\usepackage{multirow} 

\usepackage{natbib}
\biboptions{numbers,sort&compress}

\graphicspath{{Figures/}
{Figures/dppm/}
{Figures/dppm/data_plots/}
{Figures/dppm/results_plot/}
{Figures/dppm/results_plot/celeux-two-components-results/}
{Figures/dppm/results_plot/real-data-results/}
{Figures/dppm/results_plot/sensitivity/}
{Figures/FiguresResults-nips4b/}
{Figures/FiguresResults-nips4b/IHMM_Whale_demo_NIPS/}
{Figures/FiguresResults-nips4b/IHMM_Whale_demo_NIPS_lI/}
{Figures/FiguresResults-nips4b/IHMM_Whale_demo_NIPS_lkB/}
	     }
\newcommand{\bA}{\mathbf{A}}

\newcommand{\bD}{\mathbf{D}}

\newcommand{\bH}{\mathbf{H}}

\newcommand{\bI}{\mathbf{I}}

\newcommand{\bx}{\mathbf{x}}
\newcommand{\bX}{\mathbf{X}}

\newcommand{\bz}{\mathbf{z}}

\newcommand{\bsx}{\boldsymbol{x}}
\newcommand{\bsX}{\boldsymbol{X}}

\newcommand{\cN}{\mathcal{N}}
\newcommand{\cM}{\mathcal{M}}
\newcommand{\cI}{\mathcal{I}}
\newcommand{\cW}{\mathcal{W}}

\newcommand{\cB}{\mathcal{B}}
\newcommand{\cD}{\mathcal{D}}

\newcommand{\cG}{\mathcal{G}}

\newcommand{\diff}{\mathrm{d}}
\newcommand{\bspi}{\boldsymbol{\pi}}
\newcommand{\bsmu}{\boldsymbol{\mu}}

\newcommand{\bsalpha}{\boldsymbol{\alpha}}

\newcommand{\bsSigma}{\boldsymbol{\Sigma}}

\newcommand{\bsLambda}{\boldsymbol{\Lambda}}
\newcommand{\bstheta}{\boldsymbol{\theta}}
\newcommand{\bsTheta}{\boldsymbol{\Theta}}

\newcommand{\Identity}{\textbf{I}}

\newcommand{\R}{\mathbb{R}}
	
\newcommand{\N}{\mathcal{N}}

\newcommand{\tr}{\mathrm{tr}}

\journal{Elsevier}

\begin{document}
\begin{frontmatter}
\title{Dirichlet Process Parsimonious Mixtures for clustering}

\author[UNICAEN-LMNO]{Fa\"icel Chamroukhi \corref{cor1}} \ead{faicel.chamroukhi@unicaen.fr}
\author[UNICAEN-LMNO]{Marius Bartcus} 
\author[LSIS-AMU,LSIS-UTLN]{Herv\'e Glotin}  
\cortext[cor1]{Corresponding author: Fa\"icel Chamroukhi\\ Normandie Univ, UNICAEN, UMR CNRS LMNO, Department of Mathematics and Computer Science, 14000 Caen, France \\Tel: +33(0) 2 31 56 73 67  \, ; \, Fax: +33(0) 2 31 56 73 20 }

\address[UNICAEN-LMNO]{Normandie Univ, UNICAEN, UMR CNRS LMNO, Department of Mathematics and Computer Science, 14000 Caen, France}
\address[LSIS-AMU]{Aix Marseille Universit\'e, CNRS, ENSAM, LSIS, UMR 7296, 13397 Marseille, France}
\address[LSIS-UTLN]{Universit\'e de Toulon, CNRS, LSIS, UMR 7296, 83957 La Garde, France} 

\begin{abstract}
The parsimonious Gaussian mixture models,  which  exploit an eigenvalue decomposition of the group covariance matrices of the Gaussian mixture, have shown their success  in particular in cluster analysis.  Their estimation is in general performed by maximum likelihood estimation and has also been considered from a parametric Bayesian prospective. We propose new Dirichlet Process Parsimonious mixtures (DPPM) which represent a Bayesian nonparametric formulation of these parsimonious Gaussian mixture models. The proposed DPPM models are Bayesian nonparametric parsimonious mixture models that allow to simultaneously infer the model parameters, the optimal number of mixture components and the optimal parsimonious mixture structure from the data. We develop a Gibbs sampling technique for maximum a posteriori (MAP) estimation of the developed DPMM models  and  provide a Bayesian model selection framework by using Bayes factors. We apply them to cluster simulated data and real data sets, and compare them to the 
standard parsimonious mixture models. The obtained results highlight the effectiveness of the proposed nonparametric parsimonious mixture models as a good nonparametric alternative for the parametric parsimonious models.
\end{abstract}


\end{frontmatter} 



\section{Introduction}
\label{sec:introduction}
Clustering is one of the essential tasks in statistics and machine learning. Model-based clustering, that is the clustering approach based on the parametric finite mixture model \citep{McLachlan2000FMM}, is one of the most popular and successful approaches in cluster analysis \citep{mclachlan-basford88,mbc_banfield_raftery93,Fraley2002-MBC}. 
The finite mixture model decomposes the density of the observed data as a weighted sum of a finite number of $K$ component densities. Most often, the used model for multivariate real data is the finite Gaussian mixture model (GMM) in which each mixture component is Gaussian. This chapter will be focusing on Gaussian mixture modeling for multivariate real data. 

In \cite{mbc_banfield_raftery93} and \cite{celeux-and-govaert-parsimoniousGMM-95}, the authors developed a parsimonious GMM clustering approach by exploiting an eigenvalue decomposition of the group covariance matrices  of the GMM components, which provides a wide range of very flexible models with different clustering criteria. 
It was also demonstrated in \cite{Fraley2002-MBC}   that the parsimonious mixture model-based clustering framework provides very good results in density estimation as well as in cluster and discriminant analyses.

In model-based clustering using GMMs, the parameters of the Gaussian mixture are usually estimated in a maximum likelihood estimation (MLE) framework by maximizing the observed data likelihood. This is usually performed by the EM algorithm \citep{dlr,McLachlanEM2008} or EM extensions \citep{McLachlanEM2008}.
The parameters of the parsimonious Gaussian mixture models can also be estimated in a MLE framework by using the EM algorithm \citep{celeux-and-govaert-parsimoniousGMM-95}.  

However, a possible issue in the MLE approach using the EM algorithm for normal mixtures is that it may fail due to singularities or degeneracies, as highlighted namely in \cite{Stephens-thesis-97,snoussi-djafari-penalized-likelihood2001, snoussi-djafari-degeneracy2005,FraleyAndRaftery-2005} and \cite{FraleyAndRaftery-2007}.
The Bayesian estimation methods for mixture models have lead to intensive research in the field for dealing with the problems encountered in MLE for mixtures 
\citep{DieboltAndRobert1994,
EscobarANDWest-95-BayesianMixtures,
RobertBayesianChoiceBook2007,
RichardsonANDGreen97,
Stephens-thesis-97,
Bensmail-model-based-clust97,
Bensmail-and-meluman-MBC-03,
Marin2005Bayes-modeling-inference-mixtures,
bayesdataanalysis-2006} which rely on a Bayesian formulation of the the mixture model. They allow to avoid these problems by replacing the MLE by the maximum a posterior (MAP) estimator. This is namely achieved by introducing a regularization over the model parameters via prior parameter distributions, which are assumed to be uniform in the case of MLE.
%

The MAP estimation for the Bayesian Gaussian mixture is performed by maximizing the  posterior parameter distribution.   This can be performed, in some situations by an EM-MAP scheme as in \cite{FraleyAndRaftery-2005} and \cite{FraleyAndRaftery-2007} where the authors proposed an EM algorihtm for estimating Bayesian parsimonious Gaussian mixtures.
However, the common estimation approach in the case of Bayesian mixtures is still the one based on Bayesian sampling such as Markov Chain Monte Carlo (MCMC), namely Gibbs sampling \citep{DieboltAndRobert1994,Stephens-thesis-97,Bensmail-model-based-clust97} when the number of mixture components $K$ is known, or by reversible jump MCMC introduced by \cite{Green95reversiblejump, RichardsonANDGreen97} and \cite{Stephens-thesis-97}, when $K$ is unknown.
%
%
The flexible eigenvalue decomposition of the group covariance matrix described previously was also exploited in Bayesian parsimonious model-based clustering  by  \cite{Bensmail-model-based-clust97,Bensmail-and-meluman-MBC-03} where the authors used a Gibbs sampler for the model inference. 
%
%

For these model-based clustering approaches, the number of mixture components is usually assumed to be known. 
Another issue in the finite mixture model-based clustering approach, including the MLE approach as well as the MAP approach, is therefore the one of selecting the optimal number of mixture components, that is the problem of model selection. 
The model selection is in general performed through a two-fold strategy by selecting the best model from pre-established inferred model candidates.
For the MLE approach, the choice of the optimal number of mixture components can be performed via penalized log-likelihood criteria such as the Bayesian Information Criterion (BIC) \citep{BIC}, the Akaike Information Criterion (AIC) \citep{AIC}, the Approximate Weight of Evidence (AWE)  criterion \citep{mbc_banfield_raftery93}, or the Integrated Classification Likelihood criterion (ICL) \citep{ICL}, etc. 
For the MAP approach, this  can still be performed via modified penalized log-likelihood criteria such as a modified version of BIC \citep{FraleyAndRaftery-2007} computed for the posterior mode, and  more generally the Bayes factors \citep{BayesFactors-KassAndRaftery-1995} as in \cite{Bensmail-model-based-clust97} for parsimonious mixtures. Bayes factors are indeed the natural Bayesian criterion for model selection and comparison in the Bayesian framework and for which  the criteria such as BIC, AWE, etc represent indeed approximations.
There is also Bayesian extensions for mixture models that analyze mixtures with unknown number of components, for example as mentioned before the one of \cite{RichardsonANDGreen97} using RJMCMC and the one of 
\cite{Stephens98bayesiananalysis-mixtures,Stephens-thesis-97} using the birth and death process. They are referred to as fully Bayesian mixture models \citep{RichardsonANDGreen97} as they consider the number of mixture components as a parameter to be inferred from the data, jointly with the mixture model parameters, based on the posterior distributions.

However, these standard finite mixture models, including the non-Bayesian and the Bayesian ones, are parametric and may not be well adapted in the case of unknown and complex data structure.  
Recently, the Bayesian-non parametric (BNP) formulation of mixture models, that goes back to \cite{DP-Ferguson73} 
and \cite{Antoniak74-DPM}, have took much attention as a nonparametric alternative for formulating mixtures. The BNP methods \citep{RobertBayesianChoiceBook2007, bayes-non-param-princ-practic2010}  have indeed recently become popular due to their flexible modeling capabilities and advances in inference techniques, in particular for mixture models,  by using namely MCMC sampling techniques \citep{Neal2000MCMC-DPM, Rasmussen2000} 
or variational inference ones \citep{variational-dpm-blei-jordan2006}.
 BNP methods for clustering, including Dirichlet Process Mixtures (DPM) and Chinese Restaurant Process (CRP) mixtures \citep{DP-Ferguson73, Antoniak74-DPM, pitman95, Wood2008, Gershman-Blei-tutoBNP-2012} %
  which can be represented as infinite Gaussian mixture models as in \cite{Rasmussen2000},  provide a principled way to overcome the issues in standard model-based clustering and classical Bayesian mixtures for clustering.
They are fully Bayesian approaches that offer a principled alternative to jointly infer the number of mixture components (i.e clusters) and the mixture parameters, from the data. By using general processes as priors, they allow to avoid the problem of singularities and degeneracies of the MLE, and to simultaneously infer the optimal number of clusters from the data, in a one-fold scheme, rather than in a two-fold approach as in standard model-based clustering.
They also avoid assuming restricted functional forms and thus allow the complexity and accuracy of the inferred models to grow as more data is observed.  
They also represent a good alternative to the difficult problem of model selection in parametric mixture models.
Note that the term  non-parametric does not mean that there are no parameters, it rather means that one would have more and more parameters, as more data are observed.

In this paper, we present a new BNP formulation of the Gaussian mixture with the eigenvalue decomposition of the group covariance matrix of each Gaussian component which has proven its flexibility in cluster analysis for the parametric case  \citep{mbc_banfield_raftery93, celeux-and-govaert-parsimoniousGMM-95, Fraley2002-MBC, Bensmail-model-based-clust97}.
We develop new Dirichlet Process mixture models with parsimonious covariance structure, which results in Dirichlet Process Parsimonious Mixtures (DPPM). They represent a Bayesian nonparametric formulation of these parsimonious Gaussian mixture models. The proposed DPPM models are Bayesian parsimonious mixture models with a Dirichlet Process prior and thus provide a principled way to overcome the issues encountered in the parametric Bayesian and non-Bayesian case and allow to automatically and simultaneously infer the model parameters and the optimal model structure from the data, from different models, going from simplest spherical ones to the more complex standard general one. 
We develop a Gibbs sampling technique for maximum a posteriori (MAP) estimation of the various models  and  provide an unifying framework for model selection and models comparison by using namely Bayes factors, to simultaneously select the optimal number of mixture components and the best parsimonious mixture structure.
The proposed DPPM are more flexible in terms of modeling and their use in clustering, and automatically infer the number of clusters from the data. 

%
%
 
The paper is organized as follows. Section \ref{sec: previous work} describes and discusses previous work on model-based clustering. Then, Section \ref{sec: proposed DPPM} presents the proposed models and the  learning technique. In Section \ref{sec: experiments}, we give experimental results to evaluate the proposed models on simulated data and real data. Finally, Section \ref{sec: conclusion} is devoted to a discussion and concluding remarks.

\section{Parametric model-based clustering}
\label{sec: previous work}
Let $\bX = (\bsx_1,\ldots,\bsx_n)$ be a sample of $n$ i.i.d observations in $\R^d$, and let $\bz = (z_1,\ldots,z_n)$ be the corresponding unknown cluster labels where $z_i \in \{1,\hdots,K\}$ represents the cluster label of the $i$th data point $\bsx_i$, $K$ being the  possibly unknown number of clusters. 

\subsection{Model-based clustering}
Parametric Gaussian clustering, also called model-based clustering  \citep{mclachlan-basford88,Fraley2002-MBC}, is based on the finite GMM \citep{McLachlan2000FMM} in which the probability density function of the data is given by:
    \begin{equation}
     p(\bsx_i|\bstheta) = \sum_{k=1}^{K}\pi_k \ \N(\bsx_i|\theta_k)
   \label{eq: GMM}
    \end{equation}where the $\pi_k$'s are the mixing proportions, $\bstheta_k=(\bsmu_k,\bsSigma_k)$ are respectively the mean vector and the covariance matrix for the $k$th Gaussian component density
    and 
    $$\bstheta= (\pi_1,\hdots,\pi_{K-1},\bsmu^T_1,\hdots,\bsmu^T_K,\text{vech}(\bsSigma_1)^T,\hdots,\text{vech}(\bsSigma_K)^T)^T$$ is the GMM parameter vector. 
    From a generative point of view, the generative process of the data for the finite mixture model can be stated as follows. First, a mixture component $z_i$ is sampled independently from a Multinomial distribution given the mixing proportions $\bspi = (\pi_1,\ldots,\pi_K)$.
 Then, given the mixture component $z_i=k$, and the corresponding parameters $\bstheta_k$, the individual $\bsx_i$ is  generated independently from a Gaussian with parameters $\bstheta_k$, that is:
\begin{eqnarray}
    \label{eq:mm_generative}
      z_i &\sim& \cM(\bspi) \\
      \bsx_i|\theta_{z_i} &\sim& \N(\bsx_i|\theta_{z_i}).
    \end{eqnarray}
The mixture model parameters $\bstheta$ is usually estimated in a Maximum Likelihood Estimation (MLE) framework by maximizing the observed data likelihood     (\ref{eq: GMM likelihood}):
  \begin{equation}
     L(\bstheta|\bX) = \prod_{i=1}^{n}\sum_{k=1}^{K}\pi_k \ \N(\bsx_i|\theta_k).
   \label{eq: GMM likelihood}
    \end{equation}via the EM algorithm \citep{dlr,McLachlanEM2008} or EM extensions \citep{McLachlanEM2008}.
    
\subsection{Bayesian model-based clustering}

As mentioned in the introduction, the MLE approach using the EM algorithm for normal mixtures may fail in some situations due to singularities or degeneracies \citep{Stephens-thesis-97, FraleyAndRaftery-2005,FraleyAndRaftery-2007}.
The Bayesian approach of mixture models avoids the problems associated with the MLE via a MAP estimation framework by maximizing the  posterior parameter distribution
  \begin{equation}
     p(\bstheta|\bX) = p(\bstheta)L(\bstheta|\bX), 
   \label{eq: GMM posterior}
    \end{equation}
    $p(\bstheta)$ being a chosen prior distribution over the model parameters $\bstheta$.
The prior distribution in general takes the following form for the GMM:
\begin{equation*}
    p(\bstheta) = p(\bspi|\bsalpha) p(\bsmu|\bsSigma,\bsmu_0,\kappa_0) p(\bsSigma|\bsmu,\bsLambda_0,\nu)
     = p(\bspi|\bsalpha)\prod_{k=1}^K  p(\bsmu_k|\bsSigma_k) p(\bsSigma_k).
    \label{eq:map_prior_prod} 
  \end{equation*}where $(\bsalpha,\bsmu_0,\kappa_0,\bsLambda_0,\nu_0)$ are hyperparameters. A common choice for the GMM is to assume conjugate priors, that is Dirichlet distribution for the mixing proportions as in \cite{RichardsonANDGreen97} and \cite{Ormonenti-IEEENN-98},  and a multivariate normal Inverse-Wishart prior distribution for the Gaussian parameters, that is a multivariate normal for the means and an Inverse-Wishart for the covariance matrices, for example as in \cite{Bensmail-model-based-clust97}, \cite{FraleyAndRaftery-2005} and \cite{FraleyAndRaftery-2007}.

From a generative point of view, to generate data from the Bayesian GMM, a first step is to sample the model parameters from the prior, that is to sample the mixing proportions from their conjugate Dirichlet prior distribution, and the mean vectors and the covariance matrices of the Gaussian components from the corresponding conjugate multivariate normal Inverse-Wishart prior. Then, the generative procedure remains the same as in the previously described generative process for the non-Bayesian finite mixture, and is summarized by the following steps:
\begin{eqnarray}
    \bspi|\bsalpha   &\sim & \cD(\bsalpha)\nonumber\\
      z_i|\bspi      &\sim& \cM(\bspi) \label{eq: Generative Bayesian GMM} \\
      \bstheta_{z_i}|G_0 &\sim& G_0 \nonumber\\
      \bsx_i|\bstheta_{z_i} &\sim& \N(\bsx_i|\bstheta_{z_i}) \nonumber
    \end{eqnarray}where $\bsalpha$ are hyperparameters of the Dirichlet prior distribution, and $G_0$ is a prior distribution for the parameters of the Gaussian component, that is
a multivariate Normal Inverse-Wishart: 
\begin{eqnarray}
  \bsSigma_k &\sim & \cI\cW (\nu_0, \Lambda_0) \\
  \bsmu_k|\bsSigma_k &\sim & \cN(\bsmu_0, \frac{\bsSigma}{\kappa_0})
  \end{eqnarray}where the $\cI\cW$ stands for the Inverse-Wishart distribution.  

The parameters $\bstheta$ of the Bayesian Gaussian mixture are estimated by MAP estimation by maximizing the  posterior parameter distribution (\ref{eq: GMM posterior}). 
The MAP estimation can still be performed  by EM, namely in the case of conjugate priors where the prior distribution is only considered for the parameters of the Gaussian components, as in \cite{FraleyAndRaftery-2005} and \cite{FraleyAndRaftery-2007}.
However, in general, the common estimation approach in the case the Bayesian GMM described above, is the one using Bayesian sampling such as MCMC sampling techniques, namely the Gibbs sampler 
\citep{Geyer1991, Neal93-MCMC, DieboltAndRobert1994, Bensmail-model-based-clust97, Ormonenti-IEEENN-98, Stephens-thesis-97}.

\subsection{Parsimonious Gaussian mixture models}
\label{ssec: parsimonious GMM}
The GMM clustering has been extended to parsimonious GMM clustering \citep{mbc_banfield_raftery93,celeux-and-govaert-parsimoniousGMM-95} by exploiting an eigenvalue decomposition of the group covariance matrices, which provides a wide range of very flexible models with different clustering criteria. 
In these parsimonious models, the group covariance matrix $\bsSigma_k$ for each cluster $k$ is decomposed  as
 \begin{equation}
   \bsSigma_k = \lambda_k \bD_k \bA_k \bD_k^T
   \label{eq: eig-value-decomp-Sigma}
  \end{equation}where $\lambda_k = |\bsSigma_k|^{1/d}$, $\bD_k$ is an orthogonal matrix of eigenvectors  of $\bsSigma_k$ and $\bA_k$ is a diagonal matrix with determinant 1 whose diagonal elements are the normalized eigenvalues of $\bsSigma_k$ in a decreasing order. As described in \cite{celeux-and-govaert-parsimoniousGMM-95}, the scalar $\lambda_k$ determines the volume of cluster $k$, $\bD_k$ its orientation and $\bA_k$ its shape. Thus, this decomposition leads to several flexible models going from simplest spherical models to the complex general one and hence is adapted to various clustering situations.
 
The parameters $\bstheta$ of the parsimonious Gaussian mixture models are estimated in a MLE framework by using the EM algorithm. The details of the EM algorithm for the different parsimonious finite GMMs are given in \cite{celeux-and-govaert-parsimoniousGMM-95}.
The parsimonious GMMs have also took much attention under the Bayesian prospective. For example, in 
 \cite{Bensmail-model-based-clust97}, the authors proposed a fully Bayesian formulation for inferring the previously described parsimonious finite Gaussian mixture models. This Bayesian formulation was applied in model-based cluster analysis \citep{Bensmail-model-based-clust97,Bensmail-and-meluman-MBC-03}. The model inference in this Bayesian formulation is performed in a MAP  estimation framework by using MCMC sampling techniques, see for example \citep{Bensmail-model-based-clust97,Bensmail-and-meluman-MBC-03}.
Another Bayesian regularization for the parsimonious GMM was proposed by  \cite{FraleyAndRaftery-2005,FraleyAndRaftery-2007} in which the maximization of the posterior can still be performed  by the EM algorithm in the MAP framework (EM-MAP).  

\subsection{Model selection in finite mixture models}
Finite mixture model-based clustering requires to specify the number of mixture components (i.e., clusters) and, in the case of parsimonious models,  the type of the model. The main issues in this parametric model are therefore the one of selecting the number of mixture components (clusters), and possibly the type of the model, that fit at best the data.  This problem  can be tackled by  penalized log-likelihood criteria  such as BIC \citep{BIC} or penalized classification log-likelihood criteria such as 
AWE \citep{mbc_banfield_raftery93} or 
ICL \citep{ICL}, etc, or more generally by using Bayes factors  \citep{BayesFactors-KassAndRaftery-1995} which provide a general way to select and compare models in (Bayesian) statistical modeling, namely in Bayesian mixture models.

Further, we consider the parsimonious GMMs (PGMMs)  mainly in a Bayesian non-parametric framework, instead of into a finite (Bayesian) mixture. This helps namely to tackle the problem of model selection from the non-parametric prospective. 

\section{Dirichlet Process Parsimonious Mixtures}
\label{sec: proposed DPPM}

\label{Lewis94estimatingbayesFactor}

The Bayesian and non-Bayesian finite mixture models described previously are however in general parametric and may not be well adapted to represent complex and realistic data sets. 
Recently,  the Bayesian-non parametric (BNP) mixtures, in particular the Dirichlet Process Mixture (DPM) \citep{DP-Ferguson73, Antoniak74-DPM, Wood2008, Gershman-Blei-tutoBNP-2012} or by equivalence the Chinese Restaurant Process (CRP) mixture \citep{Aldous1985, Pitman2002, Gershman-Blei-tutoBNP-2012}, which can be seen as an  infinite mixture model \citep{Rasmussen2000}, 
provide a principled way to overcome the issues in standard model-based clustering and classical Bayesian mixtures for clustering.
They are fully Bayesian approaches and offer a principled alternative to jointly infer the number of mixture components (i.e clusters) and the mixture parameters, from the data.
BNP mixture approaches for clustering  assume general process as prior on the infinite possible partitions, which is not restrictive as in classical Bayesian inference.  Such a prior can be a Dirichlet Process \citep{DP-Ferguson73,Antoniak74-DPM, Gershman-Blei-tutoBNP-2012} or, by equivalence, a Chinese Restaurant Process \citep{Pitman2002,Gershman-Blei-tutoBNP-2012}.   In the next section, we rely on the Dirichlet Process Mixture (DPM) formulation to derive the proposed Bayesian non-parametric formulation of the parsimonious models.

\subsection{Dirichlet Process Parsimonious Mixtures}

A Dirichlet Process (DP) \citep{DP-Ferguson73} is a distribution over distributions and has two parameters, the concentration parameter $\alpha_0>0$ and the base measure $G_0$. We denote it by $\text{DP}(\alpha, G_0)$.  
Assume there is a parameter $\tilde{\bstheta}_i$ following a distribution $G$, that is $\tilde\bstheta_i|G \sim G$. Modeling with DP means that we assume that the prior over $G$ is a DP, that is, $G$ is itself generated from a DP: $G\sim \text{DP}(\alpha, G_0)$. This can be summarized by the following generative process:
 \begin{eqnarray}
      \tilde\bstheta_i|G &\sim& G,  \ \forall i \in  {1,\ldots,n}\\
            G|\alpha, G_0 &\sim&  \text{DP}(\alpha, G_0)\cdot
\label{eq:DP}
\end{eqnarray}The DP has two properties \citep{DP-Ferguson73}. First, random distributions drawn from DP, that is $G \sim  \text{DP}(\alpha,G_0)$,  are discrete. 
Thus, there is a strictly positive probability of multiple observations taking identical values within the set  $(\tilde\bstheta_1,\cdots,\tilde\bstheta_n)$. 
Suppose we have a random distribution $G$ drawn from a DP followed by repeated draws $(\tilde\bstheta_1,\hdots, \tilde\bstheta_n)$ from that random distribution, 
\cite{BlackwellandMacQueen73} introduced a P{\'o}lya urn representation of the joint distribution of the random variables $(\tilde\bstheta_1,\hdots, \tilde\bstheta_n)$, that is
\begin{equation}
p(\tilde\bstheta_1,\hdots,\tilde\bstheta_n) = p(\tilde\bstheta_1)p(\tilde\bstheta_2|\tilde\bstheta_1)p(\tilde\bstheta_3|\tilde\bstheta_1,\tilde\bstheta_2)\hdots p(\tilde\bstheta_n|\tilde\bstheta_1,\tilde\bstheta_2,\hdots,\tilde\bstheta_{n-1}),
\label{eq:joint distribution DP}
\end{equation}which is obtained by marginalizing out the underlying random measure $G$:
\begin{equation}
p(\tilde\bstheta_1,\hdots,\tilde\bstheta_n|\alpha, G_0) = \int\left(\prod_{i=1}^n p(\tilde\bstheta_i|G)\right)\diff p(G|\alpha,G_0)
\end{equation}and results in the following  P{\'o}lya urn representation for the calculation of the predictive terms of the joint distribution (\ref{eq:joint distribution DP}):
\begin{eqnarray}
 \tilde\bstheta_i|\tilde\bstheta_1,...\tilde\bstheta_{i-1} &\sim &\frac{\alpha_0}{\alpha_0+i-1} G_0 + \sum\limits_{j=1}^{i-1}\frac{1}{\alpha_0+i-1}  \delta_{\tilde\bstheta_j}\\
 &\sim &\frac{\alpha_0}{\alpha_0+i-1} G_0 + \sum\limits_{k=1}^{K_{i-1}}\frac{n_k}{\alpha_0+i-1}  \delta_{\bstheta_k}
 \label{eq:Polya urn DP}
\end{eqnarray}
where $K_{i-1}$ 
 is the number of clusters after $i-1$ samples, $n_k$ denotes the number of times each of the parameters $\{\bstheta_k\}_{k=1}^\infty$ occurred in the set $\{\tilde\bstheta_i\}_{i=1}^n$.  
The DP therefore places its probability mass on a countability infinite collection of points, also called atoms, that is  an infinite mixture of Dirac deltas \citep{DP-Ferguson73,sethuramanDP1994,Gershman-Blei-tutoBNP-2012}:
\begin{equation}\label{eq:dp}
G = \sum_{k=1}^{\infty} \pi_k \delta_{\bstheta_k} \quad \bstheta_k|G_0 \sim G_0, \ k=1,2,...,
\end{equation}where $\pi_k$ represents the probability assigned to the $k$th atom, and the set satisfy $\sum_{k=1}^\infty \pi_k =1$, and $\bstheta_k$ is the location or value of that component (atom). These atoms are drawn independently from the base measure $G_0$.
Hence, according to the DP process, the generated parameters $\tilde\bstheta_i$ exhibit a clustering property, that is, they share repeated values with positive probability where the unique values of $\tilde\bstheta_i$ shared among the variables are independent draws for the base distribution $G_0$ \citep{DP-Ferguson73,Gershman-Blei-tutoBNP-2012}. 
The Dirichlet process therefore provides a very interesting approach for a clustering perspective, when we do not have a fixed number of clusters, in other words having an infinite mixture, say $K$ tends to infinity. 
Consider a set of observations $(\bsx_1,\ldots,\bsx_n)$ to be clustered. Clustering with DP 
adds a third step to the DP (\ref{eq:DP}), that is we assume that the random variables $\bsx_i$, given the  distribution parameters $\tilde\bstheta_i$ which are generated from a DP, are generated from a conditional distribution $f(.|\tilde\bstheta_i)$. This is the DP mixture  (DPM) model \citep{Antoniak74-DPM, Escobar1994, Wood2008, Gershman-Blei-tutoBNP-2012}. 
The DPM adds therefore a third step to the DP, that is the of generating random variables $\bsx_i$ given the  distribution parameters $\tilde\bstheta_i$.
The generative process of the DP Mixture (DPM) is therefore as follows:
\begin{eqnarray}
      G|\alpha, G_0 &\sim&  \text{DP}(\alpha, G_0) \\
      \tilde\bstheta_i|G &\sim& G \\
      \bsx_i|\tilde\bstheta_i &\sim& f(\bsx_i|\tilde\bstheta_i)
    \end{eqnarray}where $f(\bsx_i|\tilde\bstheta_i)$  is a cluster-specific density, for example a multivariate Gaussian density in the case of DP multivariate Gaussian mixture, in which $\tilde\bstheta_i$ is composed of a mean vector and a covariance matrix. In that case, the base measure $G_0$ corresponds to the prior parameters distribution which may be a multivariate normal Inverse-Wishart conjugate prior.
When $K$ tends to infinity, it can be shown that the finite mixture model (\ref{eq: GMM}) - (\ref{eq: Generative Bayesian GMM}) converges to a Dirichlet process mixture model  \citep{Ishwaren_dp2002, Neal2000MCMC-DPM, Rasmussen2000}. 
The Dirichlet process has a number of properties which make inference based on this nonparametric prior computationally tractable. 
It also has a interpretation in term of the CRP mixture \citep{Pitman2002, Gershman-Blei-tutoBNP-2012}
 which explicitly shows its suitability to clustering thanks to the integration of the hidden component labels $z_i$ in the generative process. 
 Indeed, the second property of the DP, that is the fact that random parameters drawn from a DP share identical values and thus exhibit a clustering property, connects the DP to the CRP. 
Consider a random distribution drawn from a DP $G \sim DP(\alpha,G_0)$ followed by repeated draws from that random distribution $\tilde\bstheta_i \sim G$, $\forall i \in  {1,\ldots,n}$. The structure of the shared values defines a partition of the integers from $1$ to $n$, and the distribution of this partition is a CRP \citep{DP-Ferguson73, Gershman-Blei-tutoBNP-2012}. This is defined in the following section.

\subsection{Chinese Restaurant Process parsimonious mixtures} 
Consider the unknown cluster labels $\bz=(z_1,\ldots,z_n)$ where  each value $z_i$ is an indicator random variable that 
represents the label of the unique value $\bstheta_{z_i}$ of $\tilde{\bstheta}_i$ such that $\tilde{\bstheta}_i = \bstheta_{z_i}$ for all $i \in \{1,\ldots,n\}$.
The CRP provides a distribution on the infinite partitions of the data, that is a distribution over the positive integers $1,\hdots,n$.  
Consider the following joint distribution of the unknown cluster assignments $(z_1,\hdots,z_n)$:
\begin{equation}
p(z_1,\hdots,z_n) = p(z_1)p(z_2|z_1)\hdots p(z_n|z_1,z_2,\hdots,z_{n-1})\cdot
\label{eq:joint distribution CRP}
\end{equation}
From the  P{\'o}lya urn distribution (\ref{eq:Polya urn DP}),  each predictive term of the joint distribution (\ref{eq:joint distribution CRP}) can be computed as:
\begin{equation}
\small{
p(z_i=k|z_1,...,z_{i-1};\alpha_0)
=\frac{\alpha_0}{\alpha_0+i-1} \delta(z_i,K_{i-1}+1) + \sum\limits_{k=1}^{K_{i-1}}\frac{n_k}{\alpha_0+i-1}  \delta(z_i,k)\cdot
 }
 \label{eq:Polya urn CRP}
\end{equation}where $n_k=\sum_{j=1}^{i-1} \delta(z_j,k)$ is  the number of indicator random variables taking the value $k$ after $i-1$ observations,  
and 
 $K_{i-1}+1$ is the previously unseen value.  
From this distribution, one can  therefore allow  assigning new data to possibly previously unseen (new) clusters as the data are observed, after starting with one cluster. 
The distribution on partitions induced by the sequence of conditional distributions in Eq. (\ref{eq:Polya urn CRP}) is commonly referred to as the Chinese Restaurant Process (CRP).
It can be interpreted as follows. 
Suppose there is a restaurant with an infinite number of tables and in which customers are entering and sitting at these tables. We assume that customers are social, so that the $i$th customer sits at table $k$ with probability proportional to the number of already seated customers $n_k$ ($k\leq K_{i-1}$ being a previously occupied table),  
  and may choose a new table ($k> K_{i-1}$, $k$ being a new table to be occupied) with a probability proportional to a small positive real number $\alpha$, which represents the CRP concentration parameter.  

In clustering with the CRP, customers correspond to data points and tables correspond to  clusters.
In CRP mixture, the prior $\text{CRP}(z_1,\hdots,z_{i-1};\alpha)$ (\ref{eq:Polya urn CRP}) is completed with a likelihood with parameters $\bstheta_{k}$ for  each table (cluster) $k$ (i.e., a multivariate Gaussian likelihood with mean vector and covariance matrix in the GMM case), and a prior distribution ($G_0$) for the parameters. For example, in the GMM case, one can use a conjugate multivariate normal Inverse-Wishart prior distribution for the mean vectors and the covariance matrices.
This process therefore corresponds to the fact that the $i$th customer sits at table $z_i=k$, chooses a dish (the parameter $\bstheta_{z_i}$) from the prior of that table (cluster).  
The CRP mixture can be summarized according to the following generative process.
\begin{eqnarray}
z_i &\sim & \text{CRP}(z_1,\hdots,z_{i-1};\alpha)\\
\bstheta_{z_i}|G_0 &\sim & G_0\\
\bsx_i|\bstheta_{z_i} &\sim & f(.|\bstheta_{z_i})\cdot
\end{eqnarray}where the CRP distribution is given by Eq. (\ref{eq:joint distribution CRP}), $G_0$ is a base measure (the prior distribution) and $f(\bsx_i|\bstheta_{z_i})$  is a cluster-specific density.
In the DPM and CRP mixtures with multivariate Gaussian components, the parameters $\bstheta$ of each cluster density are composed of a mean vector and a covariance matrix. In that case, a common base measure $G_0$ is a multivariate normal Inverse-Wishart conjugate prior. 

We note that in the proposed DP parsimonious mixture, or by equivalence, CRP parsimonious  mixture, the cluster covariance matrices are parametrized in terms of an eigenvalue decomposition to provide more flexible clusters with possibly different volumes, shapes and orientations. In terms of a CRP interpretation, this can be seen as a variability of dishes for each table (cluster).
We indeed use the eigenvalue value decomposition described in section \ref{ssec: parsimonious GMM} which until now has been considered only in the case of parametric finite mixture model-based clustering (eg. see \cite{celeux-and-govaert-parsimoniousGMM-95} and \cite{mbc_banfield_raftery93}), and Bayesian parametric finite mixture model-based clustering (eg. see \cite{Bensmail-model-based-clust97}, \cite{Bensmail-and-meluman-MBC-03}, \cite{FraleyAndRaftery-2005}, and \cite{FraleyAndRaftery-2007}).
We investigate twelve parsimonious models and implemented and experimented the following nine models, covering the three families of the mixture models: the general, the diagonal and the spherical family. The parsimonious models therefore go from the simplest spherical one to the more general full model. 
Table \ref{tab:DPPMs and priors} summarizes the considered parsimonious Gaussian mixture models, the corresponding prior distribution for each model and the corresponding number of free parameters for a mixture model with $K$ components for data of dimension $d$.
\begin{table}[H]
     \centering
  {\scriptsize
  \hspace*{-1cm}
  \begin{tabular}{|c|c|c|c|c|c|c|c|}
  \hline
\# & Model                              & Type    &  Prior           & Applied to         &  \# free parameters     \\ \hline \hline
1&$\lambda \Identity$                         & Spherical     &  $\cI\cG$                 & $\lambda$ & $\upsilon + 1$ \\       
2&$\lambda_k \Identity$                       & Spherical     &   $\cI\cG$                & $\lambda_k$  & $\upsilon + d$                     \\
\hline
3& $\lambda \bA$                              & Diagonal      &  $\cI\cG$                 & diagonal elements of $\lambda \bA$ & $\upsilon + d$ \\
4& $\lambda_k \bA$                            &  Diagonal     & $\cI\cG$                  &   diagonal elements of $\lambda_k \bA$ & $\upsilon + d + K - 1$             \\
 %
 \hline
5&  $\lambda \bD \bA \bD^T$                   & General       &  $\cI\cW$                 & $\bsSigma = \lambda \bD \bA \bD^T$  & $\upsilon + \omega$ \\
6& $\lambda_k \bD \bA \bD^T$                  &  General      &   $\cI\cG$ and $\cI\cW$   & $\lambda_k$ and $\bsSigma = \bD \bA \bD^T$ & $\upsilon + \omega + K - 1$    \\
7& $\lambda \bD \bA_k \bD^T$*   		& General  & $\cI\cG$  &    diagonal elements of $\lambda \bA_k$    &      $\upsilon + \omega + (K - 1)(d-1)$                   \\
8& $\lambda_k \bD \bA_k \bD^T$*			& General   &   $\cI\cG$  & diagonal elements of $\lambda_k \bA_k$& $\upsilon + \omega + (K - 1)d$\\
9& $\lambda \bD_k \bA \bD_k^T$               & General       &   $\cI\cG$                & diagonal elements of $\lambda \bA$  & $\upsilon + K\omega - (K - 1)d$     \\
10& $\lambda_k \bD_k \bA \bD_k^T$             & General       &    $\cI\cG$               & diagonal elements of $\lambda_k \bA$ & $\upsilon + K\omega - (K - 1)(d-1)$ \\
11& $\lambda \bD_k \bA_k \bD_k^T$* 		& General &   $\cI\cG$ and $\cI\cW$ 	& $\lambda$ and $\bsSigma_k = \bD_k \bA_k \bD_k^T$ &  $\upsilon + K\omega - (K - 1)$       \\
12& $\lambda_k \bD_k \bA_k \bD_k^T$            &General        &$\cI \cW$                  & $\bsSigma_k = \lambda_k \bD_k \bA_k \bD_k^T$  & $\upsilon + K\omega$ \\
  \hline
\end{tabular}
\hspace*{-1cm}}
\caption{\label{tab:DPPMs and priors}{\small Considered Parsimonious models,  the  associated prior for the covariance structure and the corresponding number of free parameters. $\cI$ denotes an inverse distribution, $\cG$ a Gamma distribution and $\cW$ a Wishart distribution.}}
\end{table}

 We used conjugate priors, that is Dirichlet distribution for the mixing proportions  \citep{RichardsonANDGreen97,Ormonenti-IEEENN-98},  and a multivariate Normal for the mean vectors and
and an Inverse-Wishart or an Inverse-Gamma prior for the covariance matrix depending on the parsimonious model as in \cite{FraleyAndRaftery-2007} and \cite{Bensmail-model-based-clust97}.

\subsection{Bayesian inference via Gibbs sampling}

 Given a sample of $n$ i.i.d observations $\bX=(\bsx_1,\ldots,\bsx_n)$ modeled by one of the proposed Dirichlet process parsimonious mixture models (DPPMs), the aim is to infer the number $K$ of latent clusters underlying the observed data, their parameters $\bsTheta = (\bstheta_1,\ldots,\bstheta_K)$ and the latent  cluster labels $\bz=(z_1,\ldots,z_n)$.  
 We developed an MCMC Gibbs sampling technique, as in \cite{Neal2000MCMC-DPM}, \cite{Rasmussen2000}, and \cite{Wood2008} for the Bayesian inference of the nonparametric parsimonious mixture models. 

The Gibbs sampler for mixtures performs in an iterative way as follows. Given an initial mixture parameters $\bstheta^{(0)}$, and the prior over the missing labels $\bz$ (here the CRP), the Gibbs sampler draws the missing labels $\bz^{(t)}$  from their posterior distribution $p(\bz|\bX,\bstheta^{(t)})$ at each iteration $t$, which is in this case a Multinomial distribution whose parameters are the posterior component probabilities. Then, given the completed data and the prior distribution $p(\bstheta)$ over the mixture parameters, the Gibbs sampler generates the mixture parameters $\bstheta^{(t+1)}$ from the corresponding posterior distribution $p(\bstheta|\bX,\bz^{(t+1)})$, which is in this conjugate prior case a multivariate Normal Inverse-Wishart, or a Normal-Inverse-Gamma distribution, depending on the parsimonious model. This Bayesian sampling procedure produces namely an ergodic Markov chain of samples $(\bstheta^{(t)})$ with a stationary distribution $p(\bstheta|\bsX)$. Therefore, after initial $M$ burn-in samples in $N$ Gibbs samples, the variables $(\bstheta^{(M+1)},...,\bstheta^{(N)})$, can be considered to be approximately distributed according to the posterior distribution $p(\bstheta|\bX)$. 
 The Gibbs sampler consists in sampling the couple $(\bsTheta,\bz)$ from their corresponding posterior distribution. The posterior distribution for $\bstheta_k$ given all the other variables is given by
 \begin{equation}
 p(\bstheta_k |\bz,\bX,\bsTheta_{-k},\alpha;H) \propto  \prod_{i|z_i=k} f(\bsx_i|z_i=k;\bstheta_k) p(\bstheta_k;H)
 \end{equation}where $\bsTheta_{-k} = (\bstheta_1,\ldots,\bstheta_{k-1},\bstheta_{k+1},\ldots,\bstheta_{K_{i-1}})$ and $p(\bstheta_k;H)$ is the prior distribution for $\bstheta_k$, that is $G_0$, with $H$ being the hyperparameters of the model.
 The cluster labels $z_i$ are similarly sampled from  the posterior distribution which is given, up to a constant, by:
  \begin{equation}
 p(z_i = k|\bz_{-i},\bX,\bsTheta,\alpha) \propto  f(\bsx_i|z_i;\bsTheta) p(z_i|\bz_{-i};\alpha)
 \end{equation}
 where $\bz_{-i} = (z_1,\ldots,z_{i-1},z_{i+1},\ldots,z_{n})$, 
and $p(z_i|\bz_{-i};\alpha)$ is the prior predictive distribution corresponds which to the CRP distribution  computed as in  Equation (\ref{eq:Polya urn CRP}).  
The prior distribution, and the resulting posterior distribution, for each of the considered models,  are close to those in \cite{Bensmail-model-based-clust97} and are provided in detail in  \ref{appendix}.

\subsubsection{Sampling the hyperparameter $\alpha$ of the DPPM}

The number of mixture components in the models depends on the concentration hyperparameter $\alpha$ of the Dirichlet Process \citep{Antoniak74-DPM}. We therefore choose to sample it to avoid fixing an arbitrary value for it. We follow the method introduced by \cite{EscobarANDWest-95-BayesianMixtures} which consists in 
sampling it by assuming a prior Gamma distribution $\alpha \sim \cG(a, b)$ with a shape hyperparameter $a>0$ and scale hyperparameter $b>0$. Then, a variable $\eta$ is introduced and sampled conditionally on $\alpha$ and the number of clusters $K_{i-1}$, according to a Beta distribution, that is,  $\eta|\alpha, K_{i-1} \sim \cB(\alpha+1, n)$. The resulting posterior distribution for the hyperparameter $\alpha$ is given by:
\begin{equation*}
\small{
 p(\alpha|\eta, K) \sim \vartheta_\eta \cG\left(a+K_{i-1}, b-\log\left(\eta\right)\right) + \left(1-\vartheta_\eta\right)\cG\left(a+K_{i-1}-1, b-\log\left(\eta\right)\right)
 }
 \label{eq: posterior of alpha}
\end{equation*}where the weights $\vartheta_\eta = \frac{a+K_{i-1}-1}{a+K_{i-1}-1+n(b-\log(\eta))}$.
The developed Gibbs sampler is summarized by the pseudo-code (\ref{algo: Gibbs for DPPM}).
\begin{algorithm}[!h]
{\small
 \caption{\label{algo: Gibbs for DPPM}Gibbs sampling for the proposed DPPM}
{\bf Inputs:} Data set $(\bx_1,\ldots,\bx_n)$ and \# Gibbs samples
\begin{algorithmic}[1]
\STATE Initialize the model hyperparameters $H$.
\STATE Start with one cluster $K_1=1, \bstheta_1=\{\bsmu_1,\bsSigma_1\}$
 \FOR{$t=2,\ldots,\#\text{samples}$}
\FOR{$i=1,\ldots,n$}
\FOR{$k=1,\ldots,K_{i-1}$}
\IF{$(n_k=\sum_{i=1}^Nz_{ik})-1=0$} 
   \STATE{Decrease $K_{i-1}=K_{i-1}-1$;\, let $\{\bstheta^{(t)}\}\leftarrow \{\bstheta^{(t)}\}\setminus\bstheta_{z_i}$}
\ENDIF
\ENDFOR
    \STATE Sample a cluster label $z^{(t)}_i$ from the posterior:\\
     $p( z_i|\bz_{\setminus z_i}, \bX,\bstheta^{(t)}, H) \propto p( \bx_i|z_i, \bstheta^{(t)}) \text{CRP}(\bz_{\setminus z_i}; \alpha)$
    \IF{$z^{(t)}_i = K_{i-1}+1$} 
      \STATE{Increase $K_{i-1}=K_{i-1}+1$ (We get a new cluster) and sample a new cluster parameter $\bstheta_{z_i}^{(t)}$ from the conjugate prior distribution $\cN\cI\cW(\bsmu_0,\kappa_0,\nu_0,\bsLambda_0)$}.
    \ENDIF
    \ENDFOR
%
 \FOR{$k=1,\ldots,K_{i-1}$}
        \STATE Sample the parameters $\bstheta_k^{(t)}$ from the posterior distribution.
       \ENDFOR
        \STATE Sample the hyperparameter $\alpha^{(t)} \sim p(\alpha^{(t)}|K_{i-1})$ from the posterior (\ref{eq: posterior of alpha})
   \STATE $\bz^{(t+1)}\gets\bz^{(t)}$
 \ENDFOR
 \end{algorithmic}
}
{\bf Outputs:} The parameters vector chain of the mixture $\hat{\bsTheta} = \{\bspi^{(t)}, \bsmu^{(t)}, \bsSigma^{(t)} \}, \ \forall t=1,\ldots, n_s$.
\end{algorithm}%
Finally, after a sufficiently large number of samples, the retained solution is the one corresponding to the posterior mode of the number of mixture components, that is the one that appears the most frequently during the sampling. 

\subsubsection{Complexity of the algorithm}
The complexity of the method is mainly related to the sampling of the labels $z_i$ and hence to the sample size and the number of components,  and model parameters $\bstheta_i$. 
 More specifically, the complexity related to each Gibbs sample is proportional to the current value of the number of mixture components $K$ and hence varies randomly from one iteration to another. Since asymptotically $K$ tends to $\alpha \log(n)$ when $n$ tends to infinity \citep{Antoniak74-DPM}, therefore, each sample requires $O(\alpha n \log(n))$ operations for sampling the class labels $z_i$. 
The parameter simulation (the mean vector and the covariance matrix) requires in the worst case (when the covariance matrix is full, that is a non-parsimonious model) approximatively $O\left(\alpha \log(n) \left(d + d^3 \right)\right)$. This gives us a complexity in $O\left(\alpha n \log(n) d^3\right)$.

\subsubsection{The label switching problem}

The statistical inference of the model parameters meaningful if the model is identifiable. 
It is well known that mixture models are not identifiable in the strict sense, but a weak identifiability up to a permutation can be established for them.
As discussed for example in \cite[Section 1.14]{McLachlan2000FMM}, this problem is not of concern in maximum likelihood fitting of mixtures via the EM algorithm.
However, identifiably in mixtures is of concern in the Bayesian framework where in the posterior simulation the mixture component labels can be interchanged from one sample to another. This problem is known as the label-switching problem. 
Different strategies were proposed in the literature to deal with this problem. 
One  simple way to deal with label switching is to impose constraints on the model parameters to force an  unique labeling in the MCMC sampling, and hence ensure identifiability. 
For example one may use ordering constrains on the parameters as in \cite{RichardsonANDGreen97} for the case of univariate Gaussian mixtures, e.g., constraints on the means, the variances, or the mixing proportions. This was also discussed in \cite{Marin2005Bayes-modeling-inference-mixtures}.  
However, \cite{Celeux99LabelSwitchingTechnicalReport,CeleuxJASA2000} showed that this strategy of forcing constrains on the model parameters is not efficient and, if it works, it does not scale to higher dimensions. 
Another approach is to post-process the posterior parameter samples by searching for the labels permutation that minimizes some loss function as in \cite{Stephens2000Jasa}.  
As discussed in \cite{Celeux99LabelSwitchingTechnicalReport} and \cite{CeleuxJASA2000},  while this  procedure  works  well,  it can  be  numerically  demanding as it is  an  offline  algorithm  needing  storing  significant  amount  of  data samples, and it is also restricted  to  the  limited  framework  of  Bayesian  analysis  of  latent  structure  models  with  conjugate  prior  distributions.  
\cite{Celeux99LabelSwitchingTechnicalReport,CeleuxJASA2000}  proposed a better solution in the same spirit of the one of  Stephens 
which consists of a  sequential k-means like  algorithm to cluster the posterior samples and which  has  several  advantages.  It  is  quite  simple,  not  specific  to  Bayesian  analysis  with  conjugate  prior  distributions  or  to  the  mixture  context,  and  it  is  not  numerically demanding.  
So what is suggested here is  to relabel the obtained posterior parameter samples when the label switching happens by the K-means-like algorithm of \cite{Celeux99LabelSwitchingTechnicalReport,CeleuxJASA2000}.

\subsection{Bayesian model comparison via Bayes factors}

This section provides the used strategy for model comparison, that is, the selection of the best model from the different parsimonious  models.
We use Bayes factors \citep{BayesFactors-KassAndRaftery-1995, BasuANDChib2003}  which provide a general way to compare models in (Bayesian) statistical modeling, and has been widely studied in the case of mixture models \citep{BayesFactors-KassAndRaftery-1995, Bensmail-model-based-clust97, Gelfand94, BradleyChib1995, BasuANDChib2003}. 
Suppose that we have two model candidates $M_1$ and $M_2$, if we assume that the two models have the same prior probability $p(M_1)=p(M_2)$, the Bayes factor   is given by
\begin{equation}
\label{eq:bf}
BF_{12} = \frac{p(\bX|M_1)}{p(\bX|M_2)}
\end{equation}which corresponds to the ratio between the marginal likelihood values of the two models $M_1$ and $M_2$. 
It is a summary of the evidence for model $M_1$ against model $M_2$ given the data $\bX$.
The marginal likelihood $p(\bX|M_m)$ for model $M_m, \ m\in \{1,2\}$,  also called the integrated likelihood, is given by
\begin{equation}
\label{eq:marginal likelihood}
p(\bX|M_m) = \int p(\bX|\bstheta_m, M_m) p(\bstheta_m|M_m) d\bstheta_m
\end{equation} where $p(\bX|\bstheta_m, M_m)$ is the likelihood of model $M_m$ with parameters $\bstheta_m$ and $p(\bstheta_m|M_m)$ is the prior density of the mixture parameters $\bstheta_m$ for model $M_m$.
As it is difficult to compute analytically the marginal likelihood (\ref{eq:marginal likelihood}), several approximations have been proposed to approximate it.  
 One of the most used approximations is the Laplace-Metropolis approximation \citep{Lewis94estimatingbayesFactor} given by
\begin{equation}\label{eq:laplace_marginal_likelihood}
 \hat{p}_{\text{Laplace}}(\bX|M_m) = (2\pi)^\frac{\nu_m}{2} |\hat{\bH}|^\frac{1}{2} p(\bX|\hat{\bstheta}_m, M_m) p(\hat{\bstheta}_m|M_m)
\end{equation} where $\hat{\bstheta}_m$ is the posterior estimation of $\bstheta_m$ (posterior mode) for model $M_m$,
$\nu_m$ is the number of free parameters of the mixture model $M_m$ as given in Table \ref{tab:DPPMs and priors}, and $\hat \bH$ is minus the inverse Hessian of the function 
$\log (p(\bX|\hat{\bstheta}_m,M_m) p(\hat{\bstheta}_m|M_m))$ evaluated at the posterior mode of $\bstheta_m$, that is $\hat{\bstheta}_m$. 
The matrix $\hat \bH$ is asymptotically equal to the posterior covariance matrix \citep{Lewis94estimatingbayesFactor}, and is computed as the sample covariance matrix of the posterior simulated sample. 
We note that, in the proposed DPPM models, as the number of components $K$ is itself a parameter in the model and is changing during the sampling, which leads to parameters with different dimension, we compute the Hessian matrix $\hat{\bH}$ in Eq.   (\ref{eq:laplace_marginal_likelihood}) by taking the posterior samples corresponding to the posterior mode of $K$. 
Once the estimation of Bayes factors is obtained, it can be interpreted as described in Table \ref{tab:bf_interpretation} as suggested by \cite{Jeffreys61}, see also \cite{BayesFactors-KassAndRaftery-1995}.
\begin{table}[!ht]
\centering
{\footnotesize
 \begin{tabular}{|c|c|c|}
\hline
BF$_{12}$ &$2 \log \text{BF}_{12}$ & Evidence for model $M_1$ \\
\hline
\hline
  $<1$ & $<0$ 		& Negative ($M_2$ is selected)\\
  $1-3$& $0-2$ 		& Not bad\\
  $3-12$& $2-5$ 	& Substantial\\
  $12-150$& $5-10$ 	& Strong\\
  $>150$& $>10$ 	& Decisive\\
\hline
\end{tabular}
\caption{\label{tab:bf_interpretation}{Model comparison using Bayes factors.}}
}
\end{table}

\section{Experiments}
\label{sec: experiments}
We perform experiments on both simulated and real data in order to evaluate our proposed DPPM models. We assess their flexibility in terms of modeling, their use for clustering and inferring the number of clusters from the data. 
We show how the proposed DPPM approach is able to automatically and simultaneously select the best model with the optimal number of clusters by using the Bayes factors, which is used to evaluate the results. We also perform comparisons with  the finite model-based clustering approach (as in \cite{Bensmail-model-based-clust97, FraleyAndRaftery-2007}), which will be abbreviated as PGMM approach.
We also use the Rand index to evaluate and compare the provided partitions, and the misclassification error rate when the number of estimated components equals the actual one.

For the simulations, we consider several situations of simulated data, from different models, and with different levels of cluster separations, in order to assess the efficiency of the proposed approach to retrieved the actual partition with the actual number of clusters.
We also assess the stability of our proposed DPPMs models regarding the choice of the hyperparameters values, by considering several situations and varying them. 
Then, we perform experiments on several real data sets and provide numerical results in terms of comparisons of the Bayes factors (via the log marginal likelihood values) and as well the Rand index and the misclassification error rate for data sets with known actual partition.
%
%
In the experiments, for each of the compared approaches and for each model, each Gibbs is run ten times with different initializations. Each Gibbs run generates 2000 samples for which 100 burn-in samples are removed. 
The solution corresponding to the highest Bayes factor, of those ten runs, is then selected.

\subsection{Experiments on simulated data}

\subsubsection{Varying the clusters shapes, orientations, volumes and separation}
\begin{figure*}[!ht]
\centering
{\footnotesize 
\hspace*{-1cm}
 \begin{tabular}{ccc}
  \includegraphics[scale=.25]{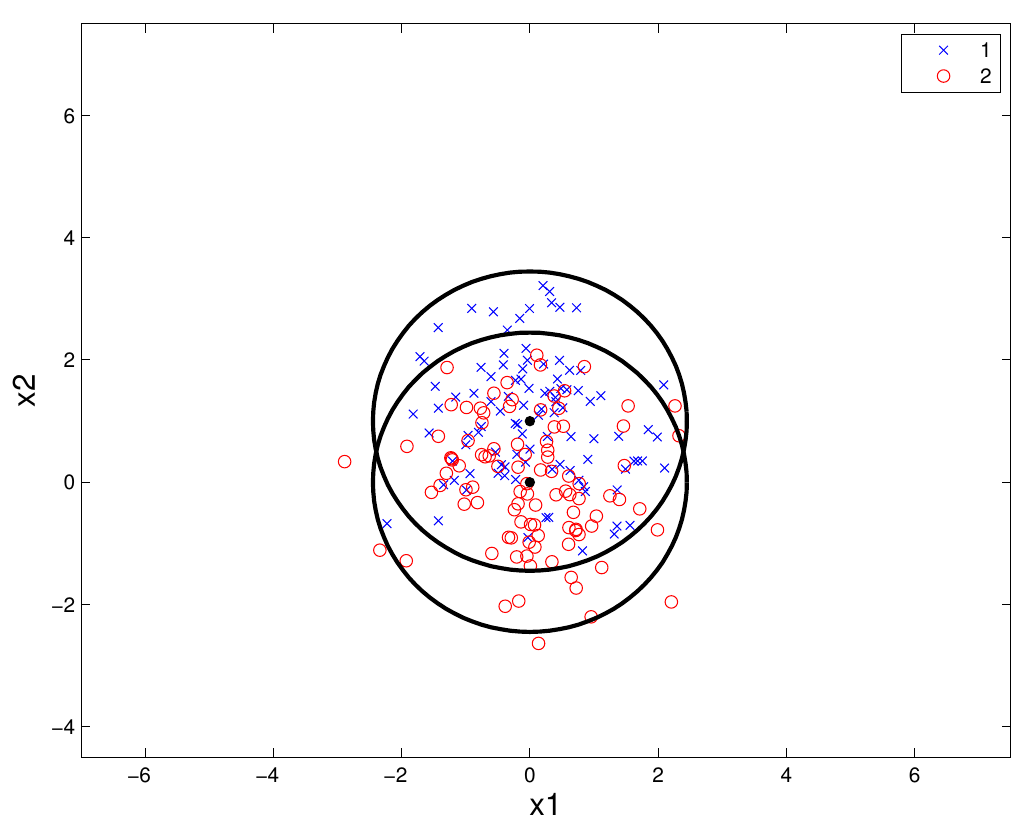}&
\includegraphics[scale=.25]{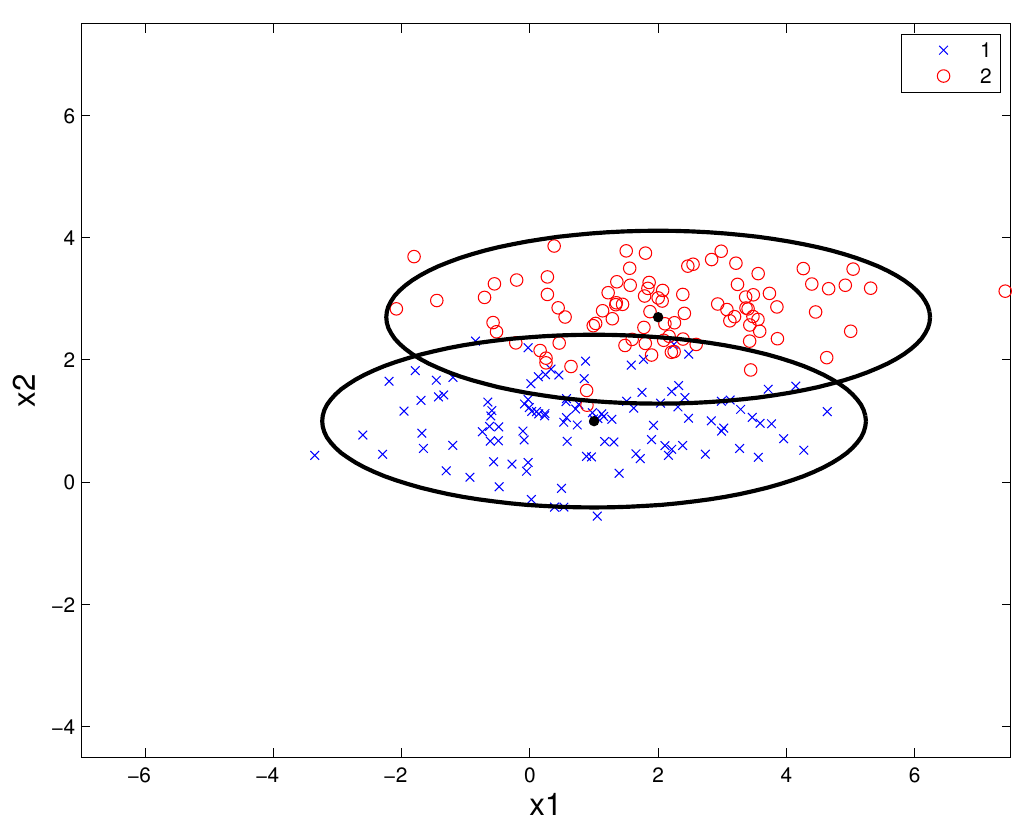}&
\includegraphics[scale=.25]{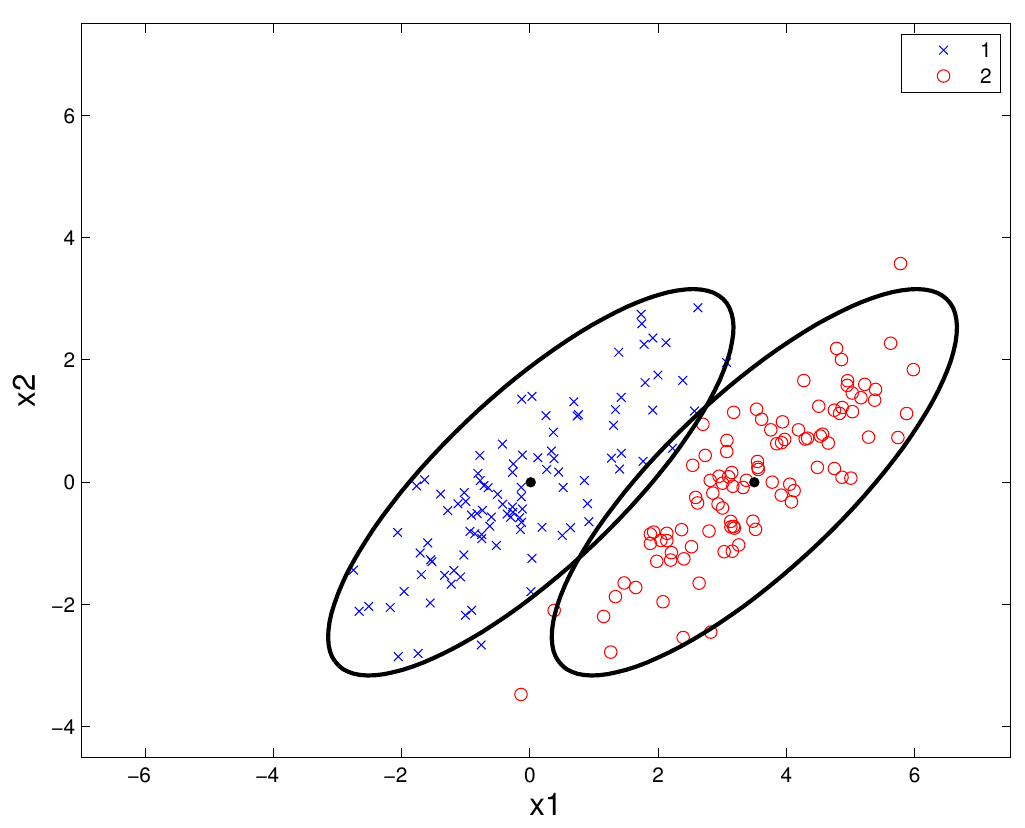}\\
 \end{tabular} 
 \hspace*{-1cm}
 \caption{\label{fig: two-class illustration with equal volume}{Examples of simulated data with the same volume across the mixture components: spherical model $\lambda \bI$ with poor separation (left), diagonal model $\lambda \bA$ with good separation (middle), and general model $\lambda \bD \bA \bD^T$  with very good separation (right).}}
}
\end{figure*}

\begin{figure*}[!ht]
\centering
{\footnotesize 
\hspace*{-1cm}
 \begin{tabular}{ccc}
\includegraphics[scale=.25]{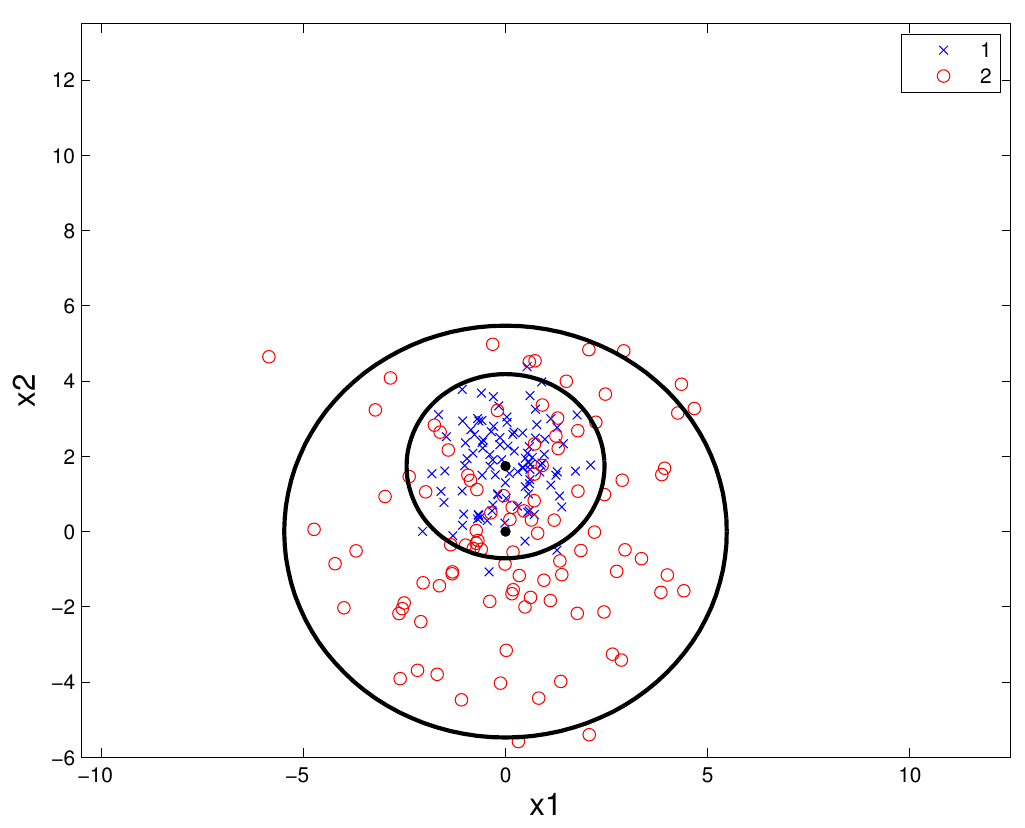}&
\includegraphics[scale=.25]{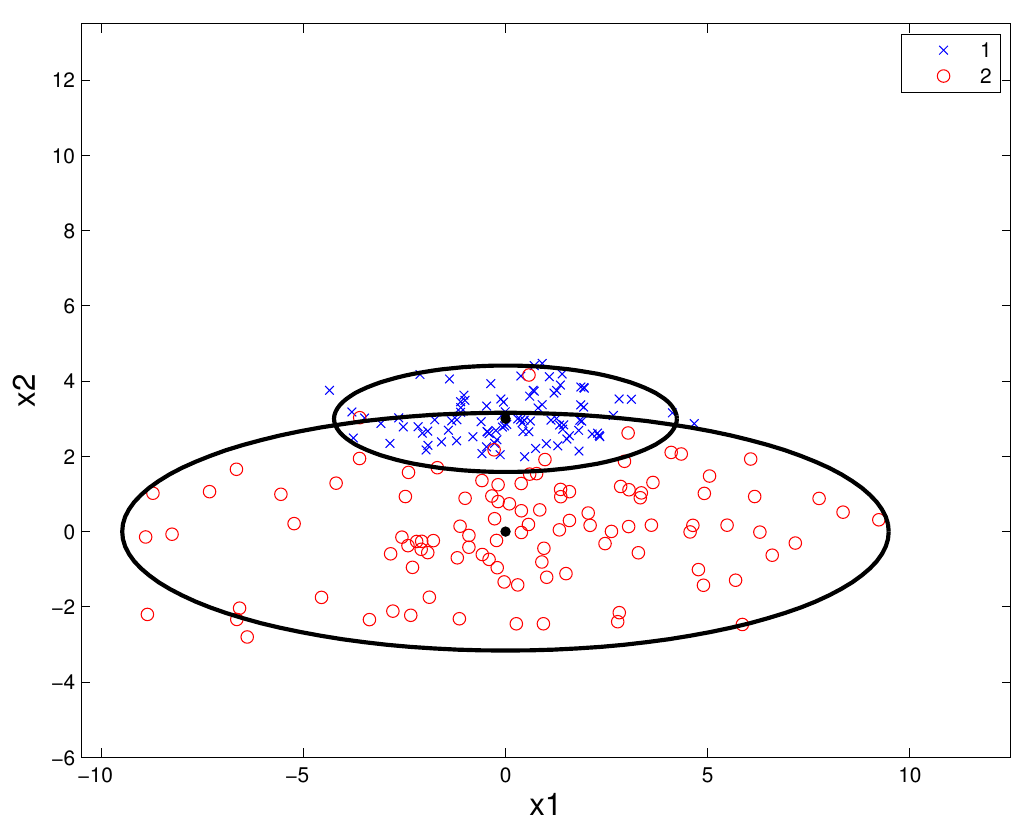}&
\includegraphics[scale=.25]{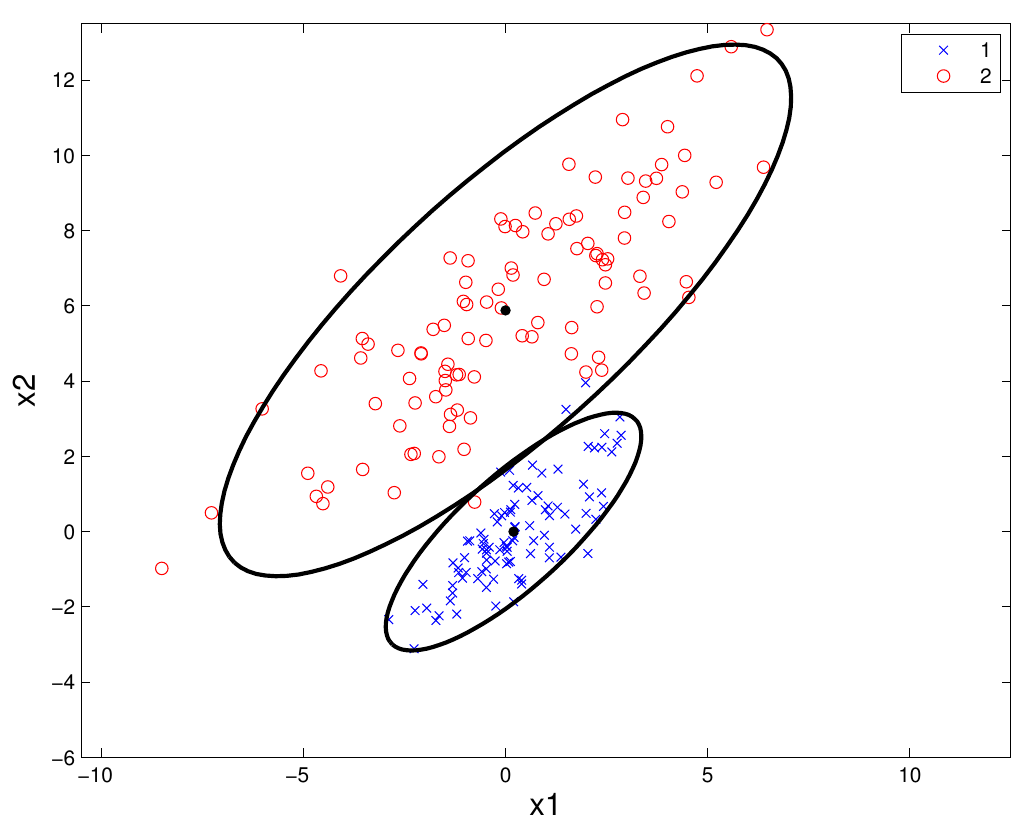}
 \end{tabular} 
 \hspace*{-1cm}
 \caption{\label{fig: two-class illustration with different volume}{Examples of simulated data with the volume changing across the mixture components: spherical model $\lambda_k \bI$ with poor separation (left), diagonal model $\lambda_k \bA$ with good separation (middle), and general model $\lambda_k \bD \bA \bD^T$  with very good separation (right).}}
 }
\end{figure*}

In this experiment, we apply the proposed models on simulated data generated according to different models, and with different level of mixture separation, going from poorly separated mixtures to very-well separated mixtures.
To simulate the data, we first consider an experimental protocol close to the one used by  \cite{celeux-and-govaert-parsimoniousGMM-95} where the authors considered the parsimonious mixture estimation within a MLE framework. This therefore allows to see how do the proposed Bayesian non-parametric DPPM perform compared to the standard parametric non-Bayesian one. 
We note however that in \cite{celeux-and-govaert-parsimoniousGMM-95} the number of components was known a priori and the problem of estimating the number of classes was not considered. 
We have performed extensive experiments involving all the models and many Monte Carlo simulations for several data structure situations. Given the variety of models, data structures, level of separation, etc, it is not possible to display all the results in the paper. We choose to perform in the same way as in the standard paper \cite{celeux-and-govaert-parsimoniousGMM-95} by selecting the results display, for the experiments on simulated data, for six models of different structures.
The data are generated from a two component Gaussian mixture in $\R^2$ with $200$ observations. 
The six different structures of the mixture that have been considered to generate the data are: two spherical models: $\lambda \bI$ and $\lambda_k \bI$, two diagonal models: $\lambda \bA$ and $\lambda_k \bA$ and two general models $\lambda \bD \bA \bD^T$ and $\lambda_k \bD \bA \bD^T$. 
Table (\ref{tab: generated data two-comp mixture}) shows the considered model structures and the respective model parameter values used to generate the data sets.
\begin{table}[!ht]
     \centering
{\footnotesize 
\begin{tabular}{|c|c|}
  \hline
Model& Parameters values            \\  
\hline   \hline
$\lambda \bI$              	& $\lambda = 1$\\
$\lambda_k \bI$           	& $\lambda_k=\{1, 5\}$ \\
$\lambda \bA$               	& $\lambda = 1;\ \bA = \text{diag}(3,1/3)$ \\
$\lambda_k \bA$              	& $\lambda_k=\{1, 5\};\ \bA = \text{diag}(3,1/3)$\\
$\lambda \bD \bA \bD^T$       & $\lambda = 1;\ \bD=\left[\frac{\sqrt{2}}{2} \ -\frac{\sqrt{2}}{2}; \frac{\sqrt{2}}{2} \ \ \frac{\sqrt{2}}{2}\right]$\\
$\lambda_k \bD \bA \bD^T$     & $\lambda_k=\{1, 5\};\ \bD=\left[\frac{\sqrt{2}}{2} \ -\frac{\sqrt{2}}{2}; \frac{\sqrt{2}}{2} \ \ \frac{\sqrt{2}}{2}\right]$\\
  \hline
\end{tabular}
\caption{\label{tab: generated data two-comp mixture}Considered two-component Gaussian mixture with different structures.}
}  
\end{table}Let us recall that the variation in the volume is related $\lambda$, the variation of the shape is related to $\bA$ and the variation of the orientation is related to $\bD$.  
Furthermore, for each type of model structure, we consider three different levels of mixture separation, that is: poorly separated, well separated, and very-well separated mixture. This is achieved by varying the following distance between the two mixture components $\varrho^2 = (\bsmu_1 - \bsmu_2)^T (\frac{\bsSigma_1 + \bsSigma_2}{2})^{-1} (\bsmu_1 - \bsmu_2)$. We consider the values $\varrho = \{1, 3, 4.5\}$. 
As a result, we obtain 18 different data structures with poorly ($\varrho = 1$), well ($\varrho = 3$) and very well ($\varrho = 4.5$) separated mixture components. 
As it is difficult to show the figures for all the situations and those of the corresponding results, in Figure \ref{fig: two-class illustration with equal volume}, 
we show for three models with equal volume across the mixture components, different data sets with varying level of mixture separation. 
Respectively, in Figure \ref{fig: two-class illustration with different volume}, we show for the models with varying  volume across the mixture components,  
different data sets with varying level of mixture separation. 

We compare the proposed DPPM to the parametric PGMM approach in model-based clustering \citep{bensmail-phd95,Bensmail-model-based-clust97,Bensmail-jasa1996}, for which the number of mixture components vary in the range $K=1,\ldots,5$ and the optimal number of mixture components was selected by using the Bayes factor (via the log marginal likelihood values).
%
For these data sets, the used hyperparameters are the following: $\bsmu_0$ was equal to the mean of the data, the shrinkage $\kappa_n = 5$, the degree of freedom $\nu_0 = d+2$, the scale matrix $\Lambda_0$ was equal to the empirical covariance matrix of the data, and the hyperparameter for the spherical models $s_0^2$ as the greatest eigenvalue of $\Lambda_0$.

\subsubsection{Obtained results}
%
%
%
%
Tables \ref{table:marginal likelihood generated data lI},
\ref{table:marginal likelihood generated data lA}
and
\ref{table:marginal likelihood generated data lDADT}
provide the obtained approximated log marginal likelihood values obtained by the PGMM and the proposed DPPM models, for, respectively, the equal (with equal clusters volumes) spherical data structure model  ($\lambda \bI$) and poorly separated mixture  ($\varrho=1$),
the equal diagonal data structure model ($\lambda \bA$) and good mixture separation ($\varrho=3$), 
and
the equal general data structure model ($\lambda \bD \bA \bD^T$) and very good mixture separation ($\varrho=4.5$).
%
%
Tables \ref{table:marginal likelihood generated data lkI},
\ref{table:marginal likelihood generated data lkA}
and
\ref{table:marginal likelihood generated data lkDADT}
provide the obtained approximated log marginal likelihood values obtained by the PGMM and the proposed DPPM models, for, respectively, the  different (with different clusters volumes) spherical data structure model ($\lambda_k \bI$) and poorly separated mixture  ($\varrho=1$),
the different diagonal data structure model ($\lambda_k \bA$) with good mixture separation ($\varrho=3$), 
and 
the different general data structure model ($\lambda_k \bD \bA \bD^T$) with very good mixture separation ($\varrho=4.5$). 
\begin{table}[!ht]
  {\scriptsize
   \centering
          \begin{tabular}{|c|c|c|c|c|c|c|c|}
    \hline &\multicolumn{2}{c|}{DPPM} &\multicolumn{5}{c|}{PGMM}\\
   \hline
   $\text{Model}$ &$\hat{K}$ & $\log\text{ML}$ & $K=1$ & $K=2$ & $K=3$ & $K=4$ & $K=5$ \\
  \hline
   \hline
 $\lambda \Identity$&2&-604.54&-633.88&-631.59&-635.07&-587.41&-595.63\\
 $\lambda_k \Identity$&2&\cellcolor{gray!25} \bf{-589.59}& -592.80&-589.88&-592.87&-593.26&-602.98\\
 $\lambda \bA$&2&-589.74&-591.67& -590.10&-593.04&-598.67&-599.75\\
 $\lambda_k \bA$&2&-591.65&-594.37&-592.46&-595.88&-607.01&-611.36\\
 $\lambda \bD \bA \bD^T$&2&-590.65& -592.20&\bf{-589.65}&-596.29&-598.63&-607.74\\
$\lambda_k \bD \bA \bD^T$&2&-591.77&-594.33&-594.89&-597.96&-594.49&-601.84\\
          \hline
   \end{tabular}
   \caption{\label{table:marginal likelihood generated data lI} Log marginal likelihood values obtained by the proposed DPPM and PGMM for the generated data with $\lambda \bI$ model structure and poorly separated mixture  ($\varrho=1$).}}
   \end{table}    
\begin{table}[!ht]
 {\scriptsize
  \centering
         \begin{tabular}{|c|c|c|c|c|c|c|c|}
   \hline &\multicolumn{2}{c|}{DPPM} &\multicolumn{5}{c|}{PGMM}\\
  \hline
  $\text{Model}$ &$\hat{K}$ & $\log\text{ML}$ & $K=1$ & $K=2$ & $K=3$ & $K=4$ & $K=5$ \\
  \hline
  \hline
$\lambda \Identity$&2&-730.31&-771.39&-702.38& -703.90&-708.71&-840.49\\
$\lambda_k \Identity$&2&-702.89&-730.26& -702.30&-704.68&-708.43&-713.58\\
$\lambda \bA$&2&\cellcolor{gray!25}\bf{-679.76}& -704.40&\bf{-680.03}&-683.13&-686.19&-691.93\\
$\lambda_k \bA$&2&-685.33&-707.26&-688.69&-696.46&-703.68&-712.93\\
$\lambda \bD \bA \bD^T$&2&-681.84&-693.44&-682.63&-688.39&-694.25&-717.26\\
$\lambda_k \bD \bA \bD^T$&2& -693.70&-695.81&-684.63&-688.17&-694.02&-695.75\\
 \hline
  \end{tabular}
  \caption{\label{table:marginal likelihood generated data lA} Log marginal likelihood values obtained by the proposed DPPM and the PGMM for the generated data with $\lambda \bA$ model structure and well separated mixture  ($\varrho=3$).}}
  \end{table}
 \begin{table}[!ht]
  {\scriptsize
   \centering
          \begin{tabular}{|c|c|c|c|c|c|c|c|}
    \hline &\multicolumn{2}{c|}{DPPM} &\multicolumn{5}{c|}{PGMM}\\
   \hline
   $\text{Model}$ &$\hat{K}$ & $\log\text{ML}$ & $K=1$ & $K=2$ & $K=3$ & $K=4$ & $K=5$ \\
  \hline
   \hline
 $\lambda \Identity$&2&-762.16&-850.66&-747.29&-746.09&-744.63&-824.06\\
 $\lambda_k \Identity$&2&-748.97&-809.46&-748.17&-751.08&-756.59&-766.26\\
 $\lambda \bA$&2&-746.05&-778.42&-746.32&-749.59&-753.64&-758.92\\
 $\lambda_k \bA$&2&-751.17&-781.31&-752.66&-761.02&-772.44&-780.34\\
 $\lambda \bD \bA \bD^T$&2&\bf{-701.94}&-746.11&\bf{-698.54}&-702.79&-707.83&-716.43\\
 $\lambda_k \bD \bA \bD^T$&2&-702.79&-748.36&-703.35&-708.77& -715.10&-722.25\\
  \hline
   \end{tabular}
   \caption{\label{table:marginal likelihood generated data lDADT} Log marginal likelihood values obtained by the proposed DPPM and PGMM for the generated data with $\lambda \bD \bA \bD^T$ model structure and very well separated mixture  ($\varrho=4.5$).}}
   \end{table}
%
%
\begin{table}[!ht]
 {\scriptsize
   \centering
         \begin{tabular}{|c|c|c|c|c|c|c|c|}
   \hline &\multicolumn{2}{c|}{DPPM} &\multicolumn{5}{c|}{PGMM}\\
  \hline
  $\text{Model}$ &$\hat{K}$ & $\log\text{ML}$ & $K=1$ & $K=2$ & $K=3$ & $K=4$ & $K=5$ \\
  \hline
  \hline
$\lambda \Identity$&3& -843.50&-869.52&-825.68&-890.26&-906.44&-1316.40\\
$\lambda_k \Identity$&2& \bf{-805.24}&-828.39& \bf{-805.21}&-808.43&-811.43&-822.99\\ 
$\lambda \bA$&2&-820.33&-823.55&-821.22&-825.58&-828.86&-838.82\\
$\lambda_k \bA$&2&-808.32&-826.34&-808.46&-816.65& -824.20&-836.85\\
$\lambda \bD \bA \bD^T$&2&   -824.00&-823.72&-821.92&-830.44&-841.22&-852.78\\
$\lambda_k \bD \bA \bD^T$&2&-821.29&-826.05&-803.96&-813.61&-819.66&-821.75\\
 \hline
  \end{tabular}
  \caption{\label{table:marginal likelihood generated data lkI}Log marginal likelihood values  and estimated number of clusters for the generated data with $\lambda_k \bI$ model structure and poorly separated mixture  ($\varrho=1$).}}
  \end{table}
\begin{table}[!ht]
  {\scriptsize
   \centering
          \begin{tabular}{|c|c|c|c|c|c|c|c|}
    \hline &\multicolumn{2}{c|}{DPPM} &\multicolumn{5}{c|}{PGMM}\\
   \hline
   $\text{Model}$ &$\hat{K}$ & $\log\text{ML}$ & $K=1$ & $K=2$ & $K=3$ & $K=4$ & $K=5$ \\
  \hline
   \hline
 $\lambda \Identity$&3&-927.01&-986.12&-938.65&-956.05&  -1141.00&-1064.90\\
 $\lambda_k \Identity$&3&-912.27&-944.87&-925.75&-911.31&-914.33&-918.99\\
 $\lambda \bA$&3&   -899.00&-918.47&-906.59&-911.13&-917.18&-926.69\\
 $\lambda_k \bA$&2&\cellcolor{gray!25}\bf{-883.05}&-921.44&-883.22&-897.99&-909.26& -928.90\\
 $\lambda \bD \bA \bD^T$&2&-903.43&-918.19&-902.23& -906.40&-914.35&-924.12\\
$\lambda_k \bD \bA \bD^T$&2&-894.05&-920.65&\bf{-876.62}&-886.86&-904.45&-919.45\\
          \hline
   \end{tabular}
   \caption{\label{table:marginal likelihood generated data lkA} Log marginal likelihood values obtained by the proposed DPPM and PGMM for the generated data with $\lambda_k \bA$ model structure and well separated mixture  ($\varrho=3$).}}
   \end{table}%
\begin{table}[!ht]
 {\scriptsize
  \centering
         \begin{tabular}{|c|c|c|c|c|c|c|c|}
   \hline &\multicolumn{2}{c|}{DPPM} &\multicolumn{5}{c|}{PGMM}\\
  \hline
  $\text{Model}$ &$\hat{K}$ & $\log\text{ML}$ & $K=1$ & $K=2$ & $K=3$ & $K=4$ & $K=5$ \\
  \hline
  \hline
$\lambda \Identity$&2&-984.33&-1077.20&-1021.60&-1012.30&  -1021.00&-987.06\\
$\lambda_k \Identity$&3&-963.45&-1035.80&-972.45&-961.91&-967.64&-970.93\\
$\lambda \bA$&2&-980.07&-1012.80&-980.92&-986.39&-992.05&-999.14\\
$\lambda_k \bA$&2&-988.75&-1015.90&-991.21&  -1007.00&-1023.70&-1041.40\\
$\lambda \bD \bA \bD^T$&3&-931.42&-984.93&-939.63&-944.89&-952.35&-963.04\\
$\lambda_k \bD \bA \bD^T$&2& \cellcolor{gray!25}\bf{-921.90}& -987.39&\bf{-921.99}&-930.61&-946.18&-956.35\\
 \hline
  \end{tabular}
  \caption{\label{table:marginal likelihood generated data lkDADT} Log marginal likelihood values obtained by the proposed DPPM and PGMM for the generated data with $\lambda_k \bD \bA \bD^T$ model structure and very well separated mixture  ($\varrho=4.5$).}}
  \end{table}

From these results,  we can see that, the proposed DPPM, in all the situations (except for the first situation in Table \ref{table:marginal likelihood generated data lI}) retrieves  the actual model, with the actual number of clusters. We can also see that, except for two situations, the selected DPPM model, has the highest log marginal likelihood value, compared to the PGMM. 
We also observe that the solutions provided by the proposed DPPM are, in some cases more parsimonious than those provided by the PGMM, and, in the other cases, the same as those provided by the PGMM.
For example, in Table \ref{table:marginal likelihood generated data lI}, which corresponds to data from poorly separated mixture, we can see that the proposed DPPM selects the spherical model $\lambda_k \bI$, which is more parsimonious than the general model $\lambda \bA$ selected by the PGMM, with a better  misclassification error (see Table \ref{table:misc error rate for generated data with equal cluster volume}). 
The same thing can be observed in Table \ref{table:marginal likelihood generated data lkA} where the proposed DPPM selects the actual diagonal model $\lambda_k \bA$, however the PGMM selects the general model $\lambda_k \bD \bA \bD^T$, while the clusters are well separated ($\varrho=3$).

Also, in terms of misclassification error, as shown in Table \ref{table:misc error rate for generated data with equal cluster volume},  the proposed DPPM models, compared to the PGMM ones, provide partitions with the lower miscclassification error, for situations with poorly, well or very-well separated clusters, and for  clusters with equal and different volumes (except for one situation).   
\begin{table}[!ht]
{\footnotesize
\centering
\begin{tabular}{|c|c|c|c|} 
\hline
PGMM & $48 \pm  8.05$ & $9.5\pm 3.68$  & $\bf{1\pm 0.80}$ \\ 
\hline
DPPM & $\bf{40 \pm 4.66}$ & $\bf{7 \pm 3.02}$ & $3 \pm 0.97$ \\ 
\hline
\end{tabular}
\caption{\label{table:misc error rate for generated data with equal cluster volume}Misclassification error rates obtained by the proposed DPPM and the PGMM approach. From left to right, the situations respectively shown in Table \ref{table:marginal likelihood generated data lI}, \ref{table:marginal likelihood generated data lA}, \ref{table:marginal likelihood generated data lDADT}}}
\end{table}

\begin{table}[!ht]
{\footnotesize
\centering
\begin{tabular}{|c|c|c|c|} 
\hline
PGMM & $23.5 \pm 2.89$ & $10.5 \pm 2.44$   &  $2 \pm 1.69$ \\ 
\hline
DPPM & $\bf{20.5 \pm 3.34}$ & $\bf{7 \pm 3.73}$  &  $\bf{1.5 \pm 0.79}$ \\ 
\hline
\end{tabular}
\caption{\label{table:misc error rate for generated data with different cluster volume}Misclassification error rates obtained by the proposed DPPM and the PGMM approach. From left to right, the situations respectively shown in Table \ref{table:marginal likelihood generated data lkI}, \ref{table:marginal likelihood generated data lkA}, \ref{table:marginal likelihood generated data lkDADT}}}
\end{table}

\begin{figure*}[!ht]
\centering
{\footnotesize
\hspace*{-1cm}
 \begin{tabular}{ccc}
  \includegraphics[scale=.25]{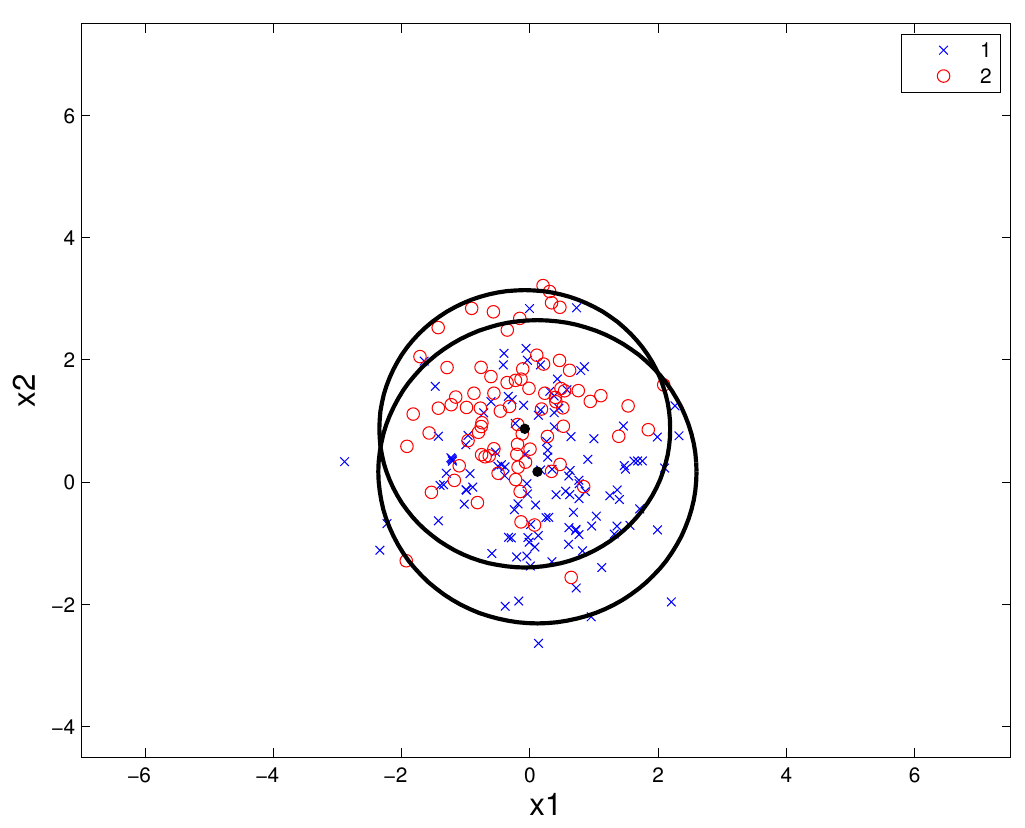}&
  \includegraphics[scale=.25]{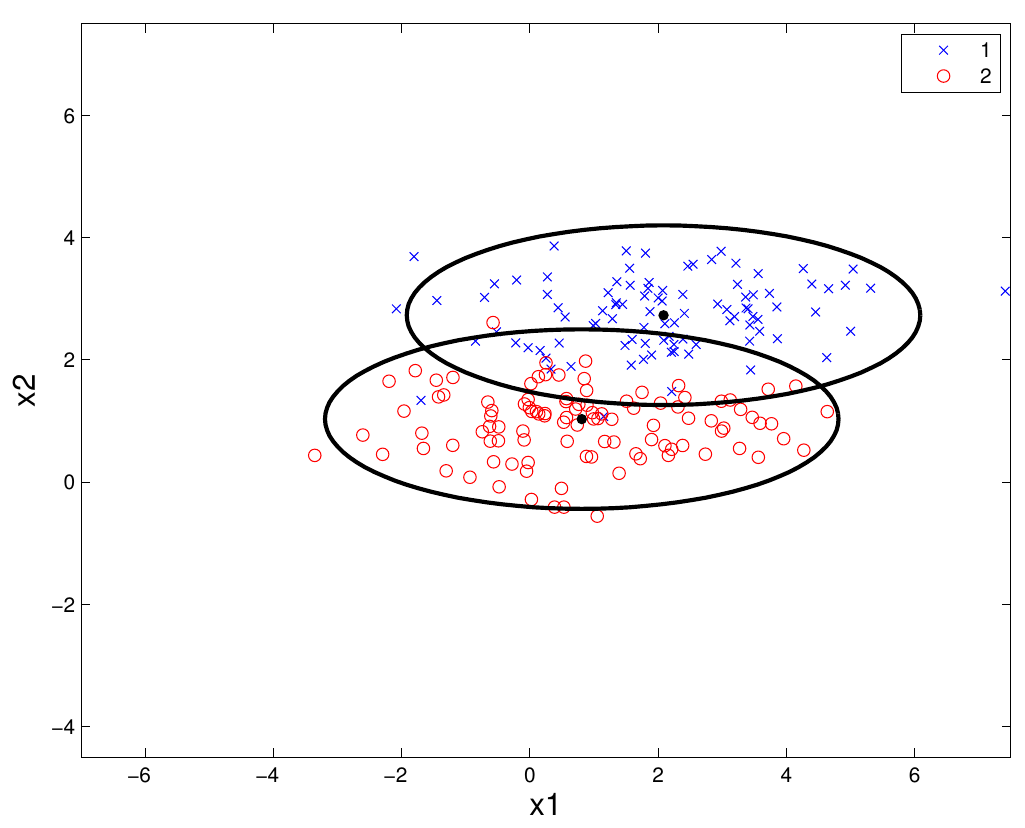}&
  \includegraphics[scale=.25]{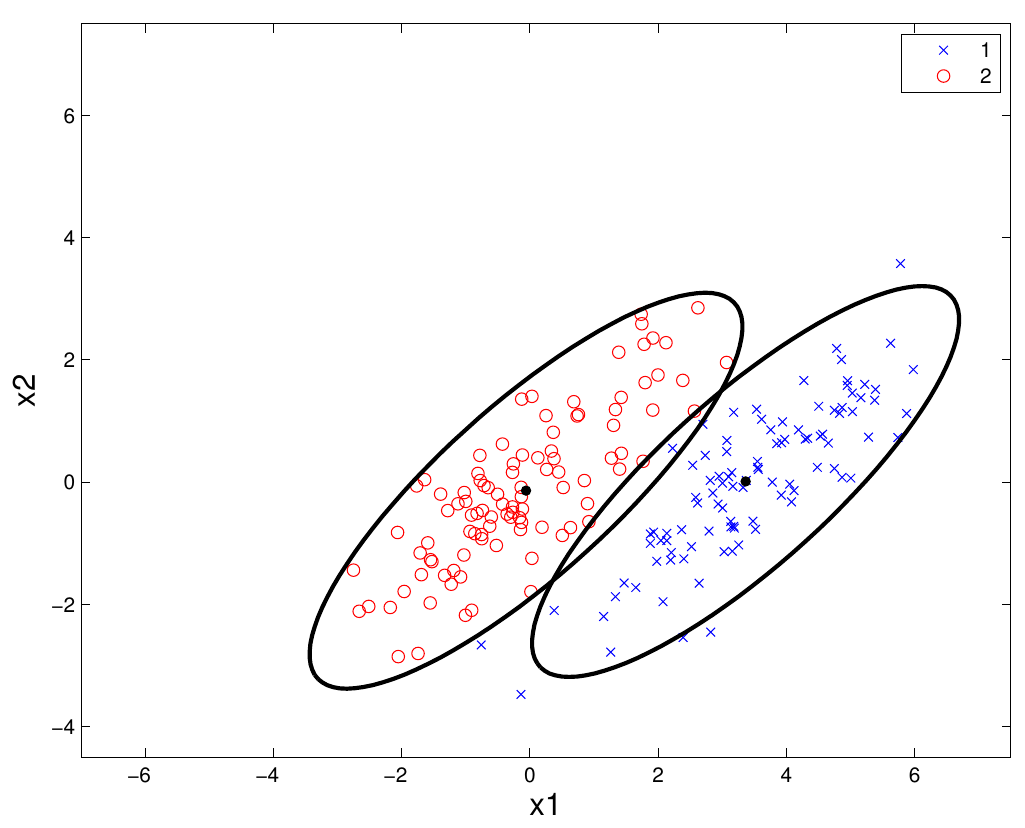}\\
  \includegraphics[scale=.25]{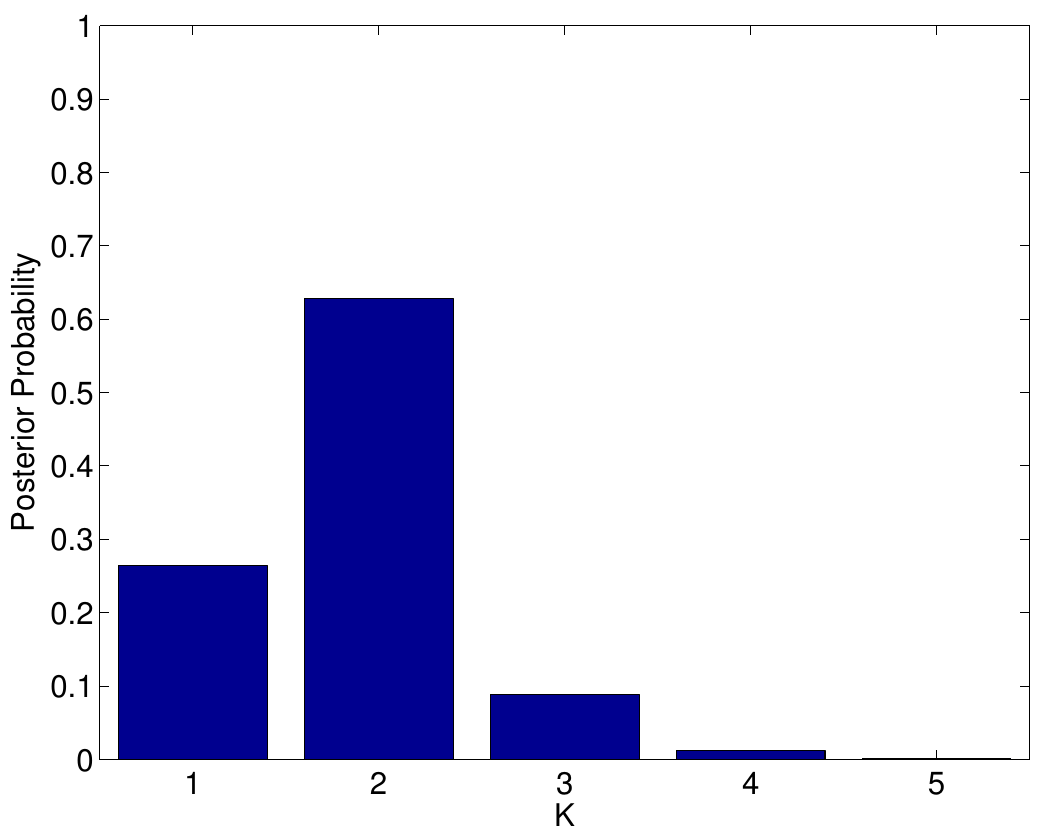}&
  \includegraphics[scale=.25]{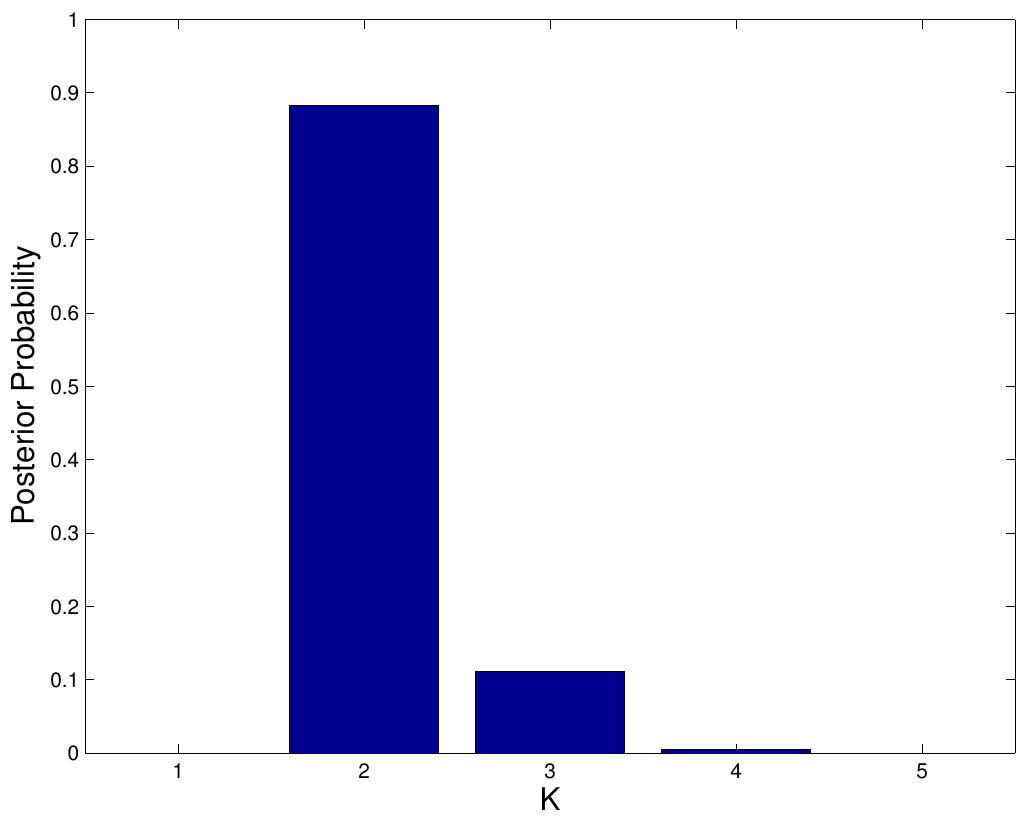}&
  \includegraphics[scale=.25]{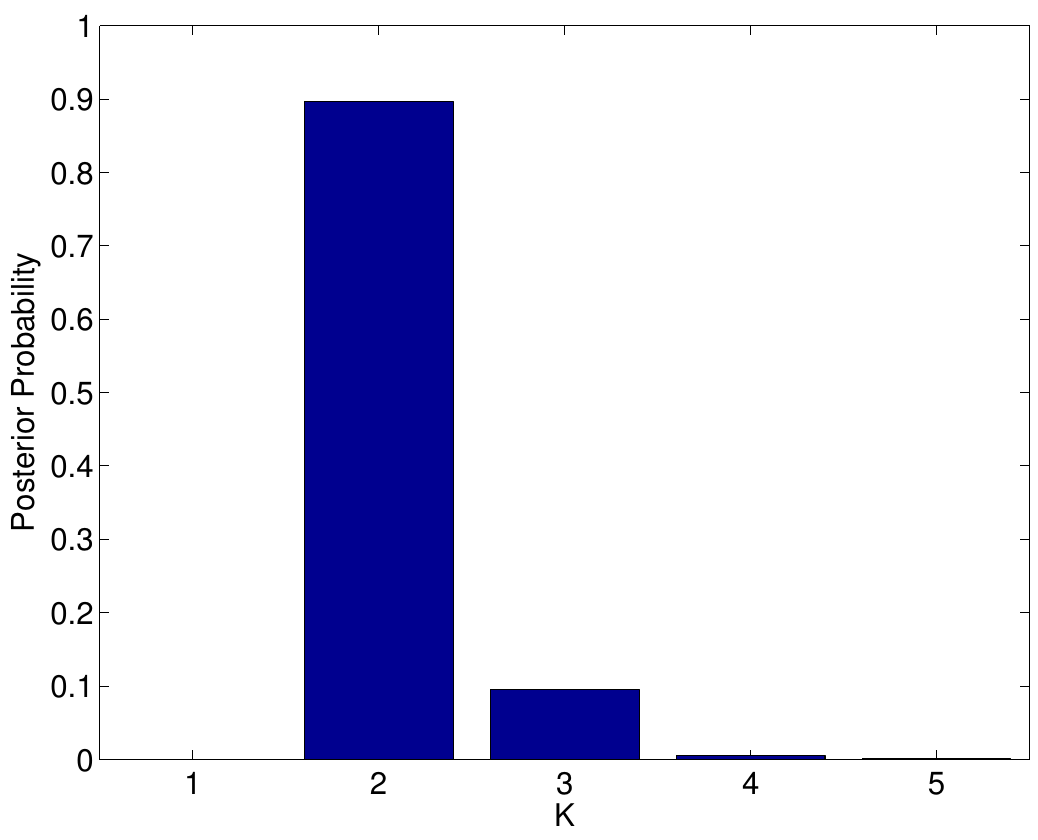}
\end{tabular}
\hspace*{-1cm}
\caption{\label{fig: best partitions for two-class data with equal volume}Partitions obtained by the  DPPM for the data sets in Fig. \ref{fig: two-class illustration with equal volume}.} 
}
\end{figure*}

\begin{figure*}[!ht]
\centering
{\footnotesize 
\hspace*{-1cm}
 \begin{tabular}{ccc}
    \includegraphics[scale=.25]{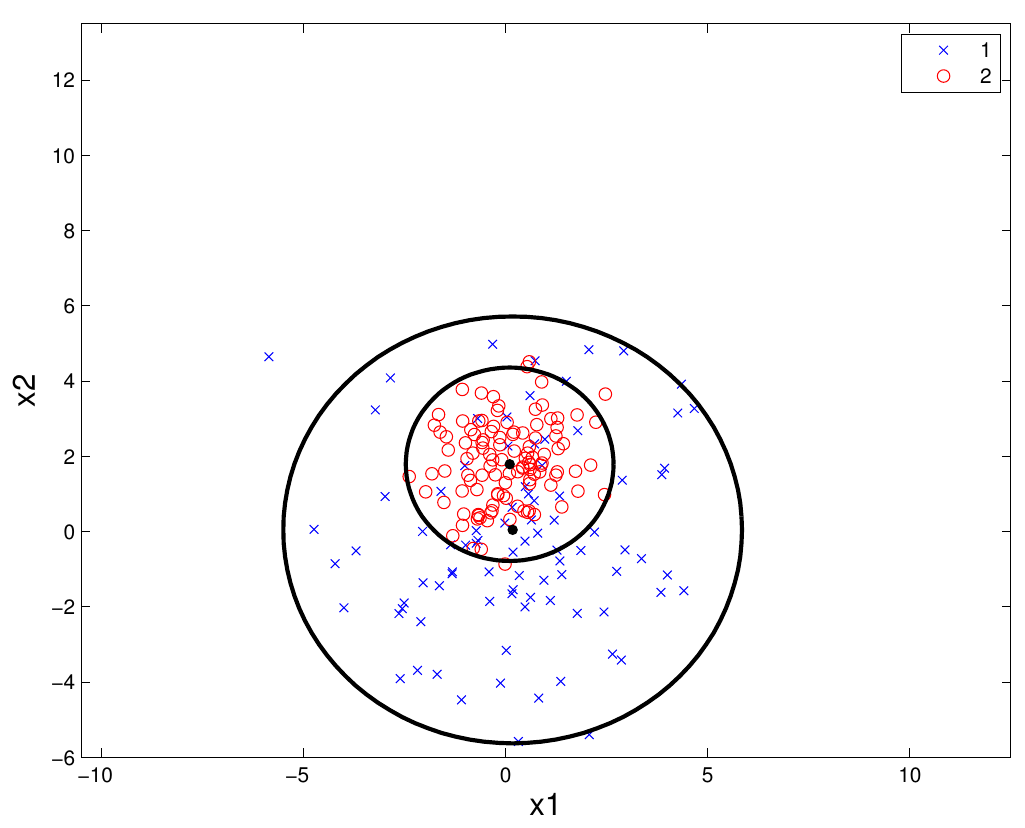}&
  \includegraphics[scale=.25]{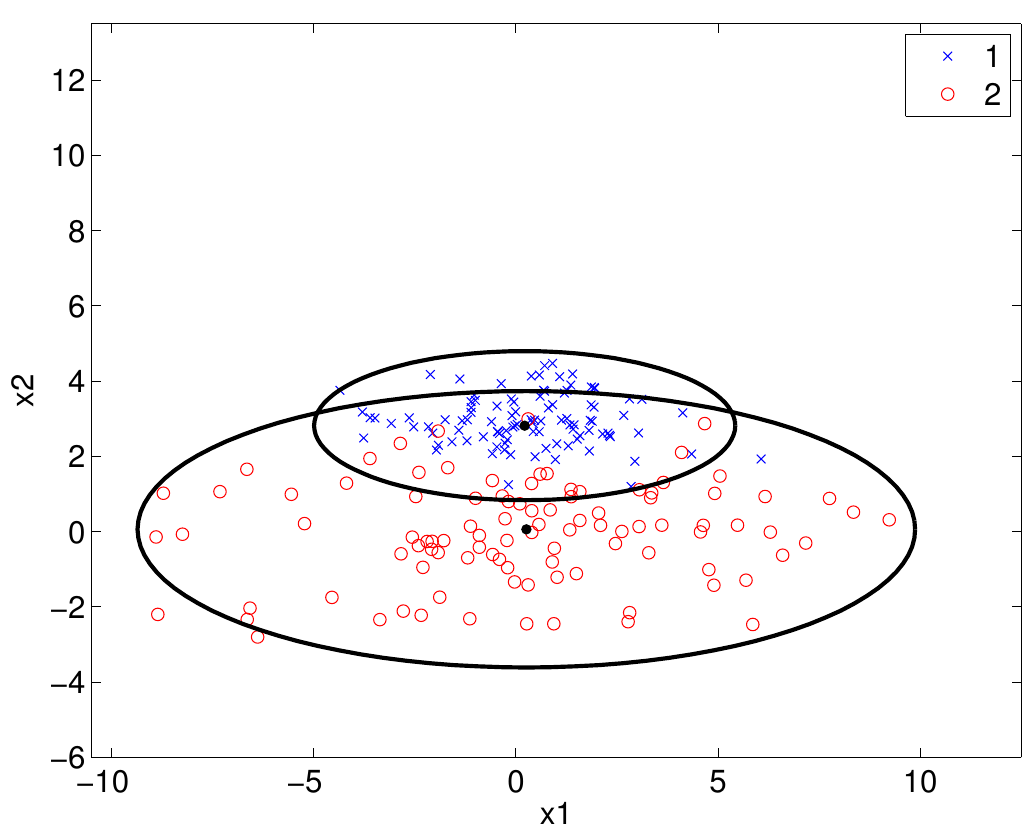}&
  \includegraphics[scale=.25]{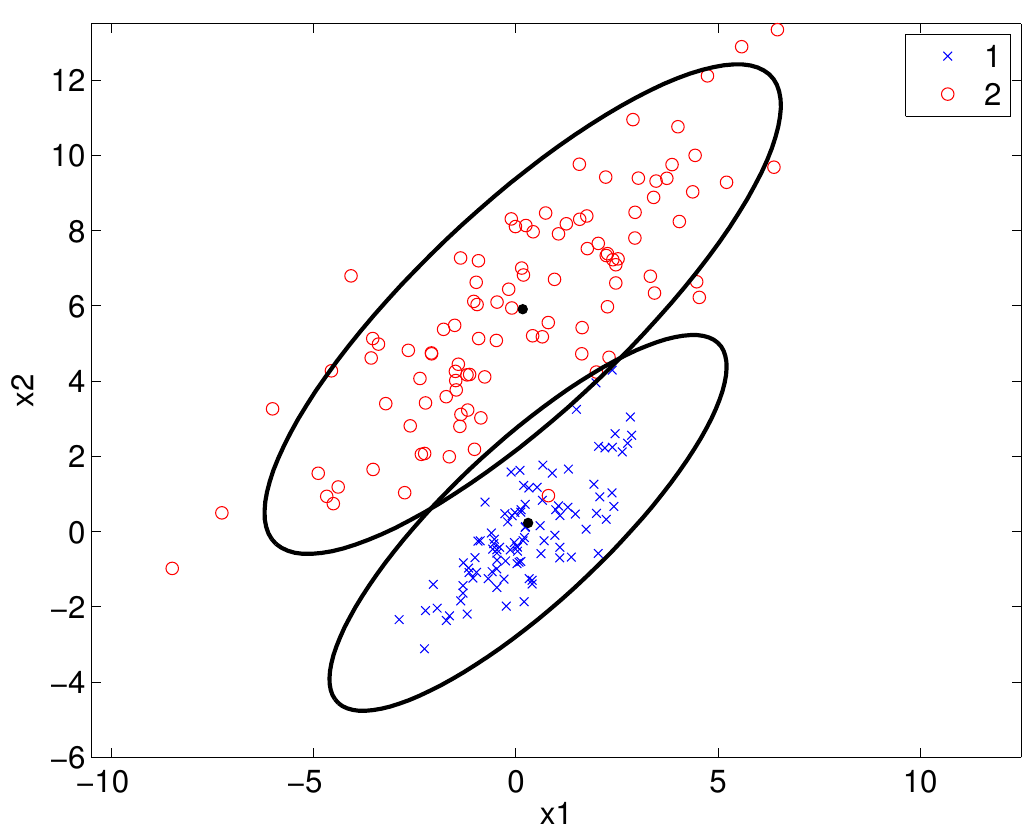}\\
   \includegraphics[scale=.25]{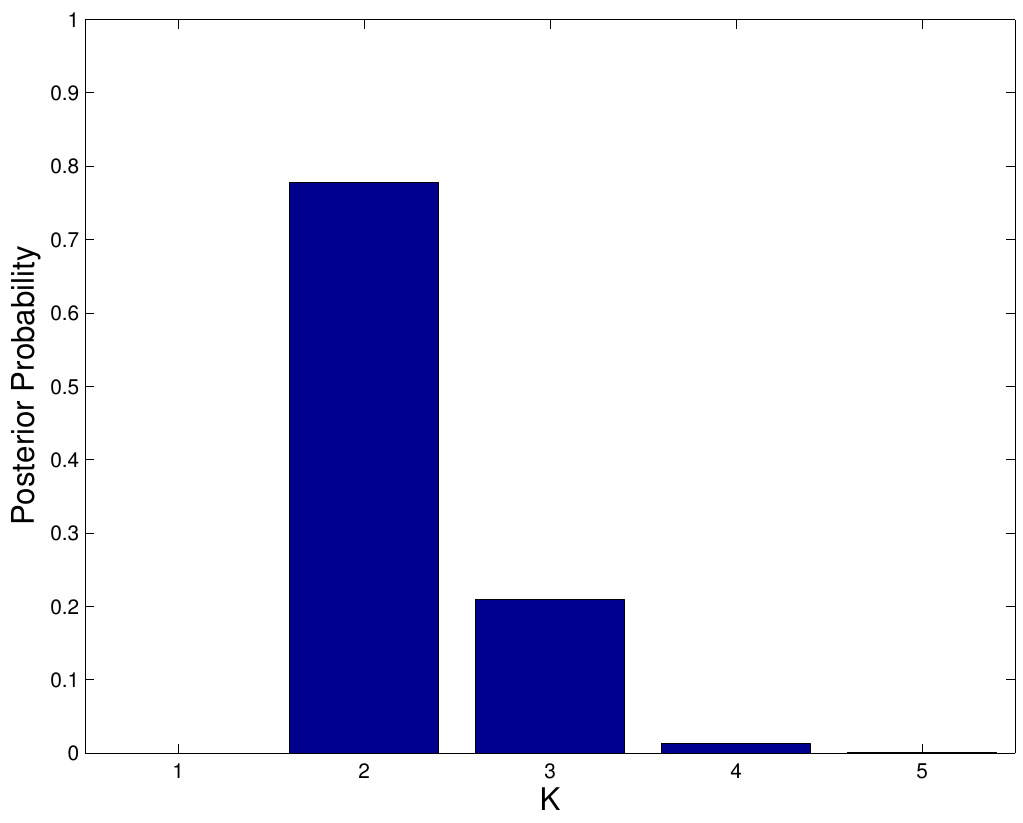}&
  \includegraphics[scale=.25]{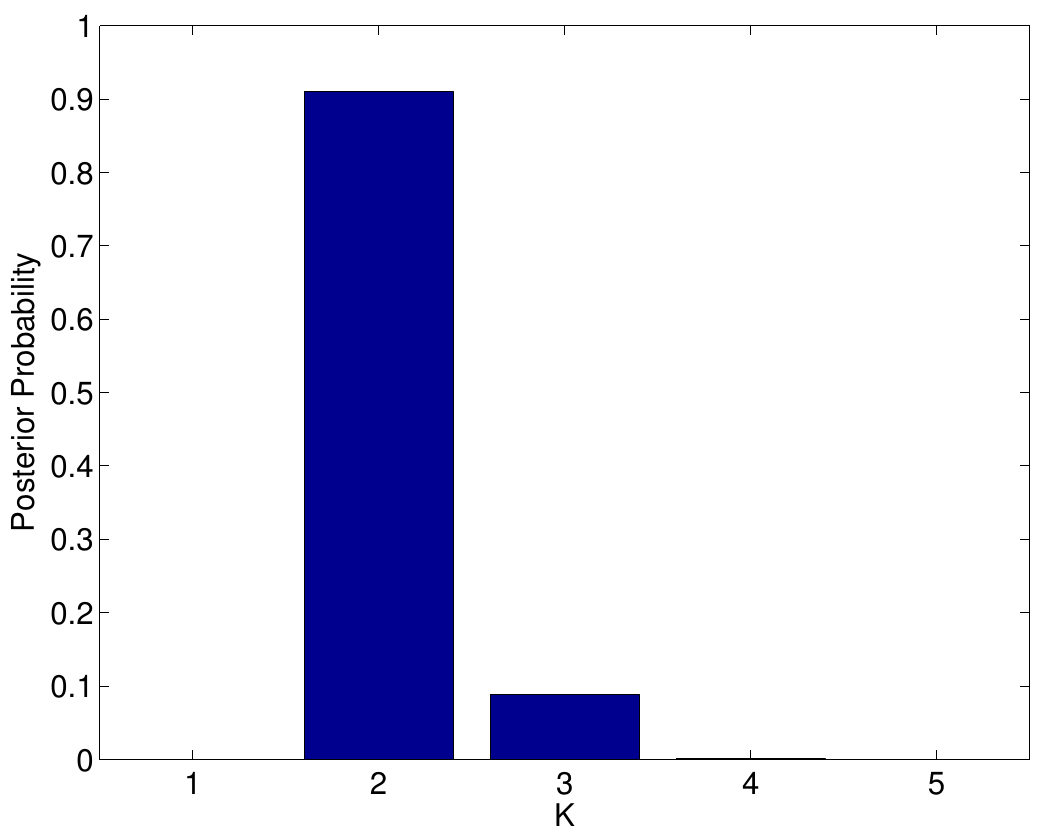}&
  \includegraphics[scale=.25]{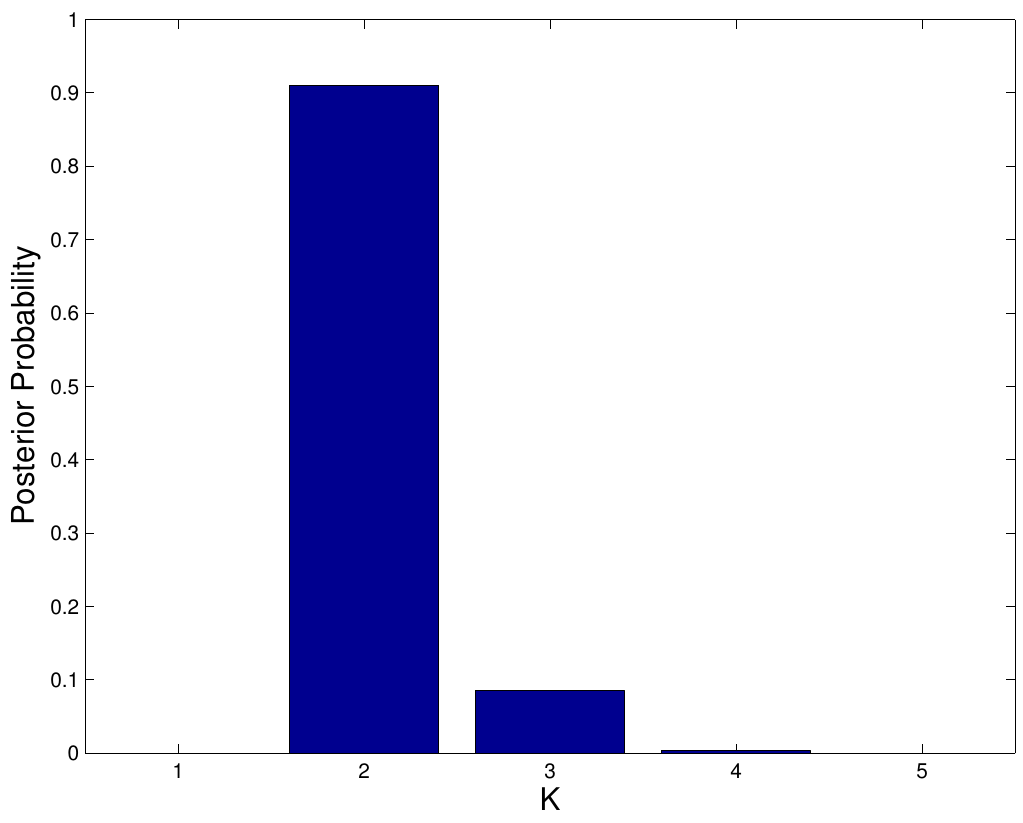}
 \end{tabular} 
 \hspace*{-1cm}
  \caption{\label{fig: best partitions for two-class data with different volume} Partitions obtained by the DPPM for the data sets in Fig. \ref{fig: two-class illustration with different volume}.}
}
 \end{figure*}

On the other hand, for the DPMM models, from the log marginal likelihood values shown  in Tables \ref{table:marginal likelihood generated data lI} to \ref{table:marginal likelihood generated data lkDADT}, we can see that the evidence of the selected model, compared to the majority of the other alternative is, according to Table \ref{tab:bf_interpretation},  in general decisive. Indeed, it can be easily seen that the value $2\log\text{BF}_{12}$ of the Bayes Factor between the selected model, and the other models, is more than 10, which corresponds to a decisive evidence for the selected model. 
Also, if we consider the evidence of the selected model, against the more competitive one, one can see from Table \ref{table:bf values for generated data with equal cluster volume} and Table \ref{table:bf values for generated data with different cluster volume}, that, for the situation  with very bad mixture separation, with clusters having the same volume, the evidence is not bad (0.3). However, for all the 
other situations, the optimal model is selected with an evidence 
going from an almost substantial evidence (a value of 1.7), to a strong and decisive evidence, especially for the models with different clusters volumes. 
We can also conclude that the models with different clusters volumes may work better in practice as highlighted by \cite{celeux-and-govaert-parsimoniousGMM-95}.
\begin{table}[!ht]
{\footnotesize
\centering
\begin{tabular}{|c|c|c|c|}
\hline 
$M_1$ vs $M_2$  & $\lambda_k \bI$ vs $\lambda \bA$ & $\lambda \bA$ vs $\lambda \bD \bA \bD^T$ & $\lambda \bD \bA \bD^T$ vs $\lambda_k \bD \bA \bD^T$\\
\hline
\hline
$2\log\text{BF}$ & 0.30 & 4.16  & 1.70 \\ 
\hline
\end{tabular}
\caption{\label{table:bf values for generated data with equal cluster volume}Bayes factor values obtained by the proposed DPPM by comparing the selected model (denoted $M_1$) and the one more competitive for it (denoted $M_2$). From left to right, the situations respectively shown in Table \ref{table:marginal likelihood generated data lI}, Table \ref{table:marginal likelihood generated data lA} and Table \ref{table:marginal likelihood generated data lDADT}}}
\end{table}
\begin{table}[!ht]
{\footnotesize
\centering
\begin{tabular}{|c|c|c|c|}
\hline 
$M_1$ vs $M_2$  & $\lambda_k \bI$ vs $\lambda_k \bA$ & $\lambda_k \bA$ vs $\lambda_k \bD \bA \bD^T $  & $\lambda_k \bD \bA \bD^T$ vs $\lambda \bD \bA \bD^T$ \\
\hline
\hline
$2\log\text{BF}$ & 6.16  & 22  & 19.04\\ 
\hline
\end{tabular}
\caption{\label{table:bf values for generated data with different cluster volume}Bayes factor values obtained by the proposed DPPM by comparing the selected model (denoted $M_1$) and the one more competitive for it (denoted $M_2$). From left to right, the situations respectively shown in Table \ref{table:marginal likelihood generated data lkI}, Table \ref{table:marginal likelihood generated data lkA} and Table (6) \ref{table:marginal likelihood generated data lkDADT}}}
\end{table} 
Finally, Figure (\ref{fig: best partitions for two-class data with equal volume}) shows the best estimated partitions for the data structures with equal volume  across the mixture components shown in Fig. \ref{fig: two-class illustration with equal volume} and the posterior distribution over the number of clusters.
%
One can see that for the case of clusters with equal volume, the diagonal family ($\lambda \bA$) with well separated mixture ($\varrho=3$) and the general family ($\lambda \bD \bA \bD^T$) with very well separated mixture ($\varrho=4.5$) data structure estimates a good number of clusters with the actual model. However, the equal spherical data model structure ($\lambda \bI$) estimates the $\lambda_k \bI$ model, which is also a spherical model.
%
%
%
%
Figure (\ref{fig: best partitions for two-class data with different volume}) shows the best estimated partitions for the data structures with different volume  across the mixture components shown in Fig. \ref{fig: two-class illustration with different volume} and the posterior distribution over the number of clusters.
%
One can see that for all of different data structure models: different spherical $\lambda_k \bI$, different diagonal $\lambda_k \bA$ and different general $\lambda_k \bD \bA \bD^T$, the proposed DPPM approach succeeded to estimate a good number of clusters equal to $2$ with an actual cluster structure. 

 \subsubsection{Stability with respect to the hyperparameters values}

\begin{figure}[!ht]
 \centering
 \includegraphics[scale=.5]{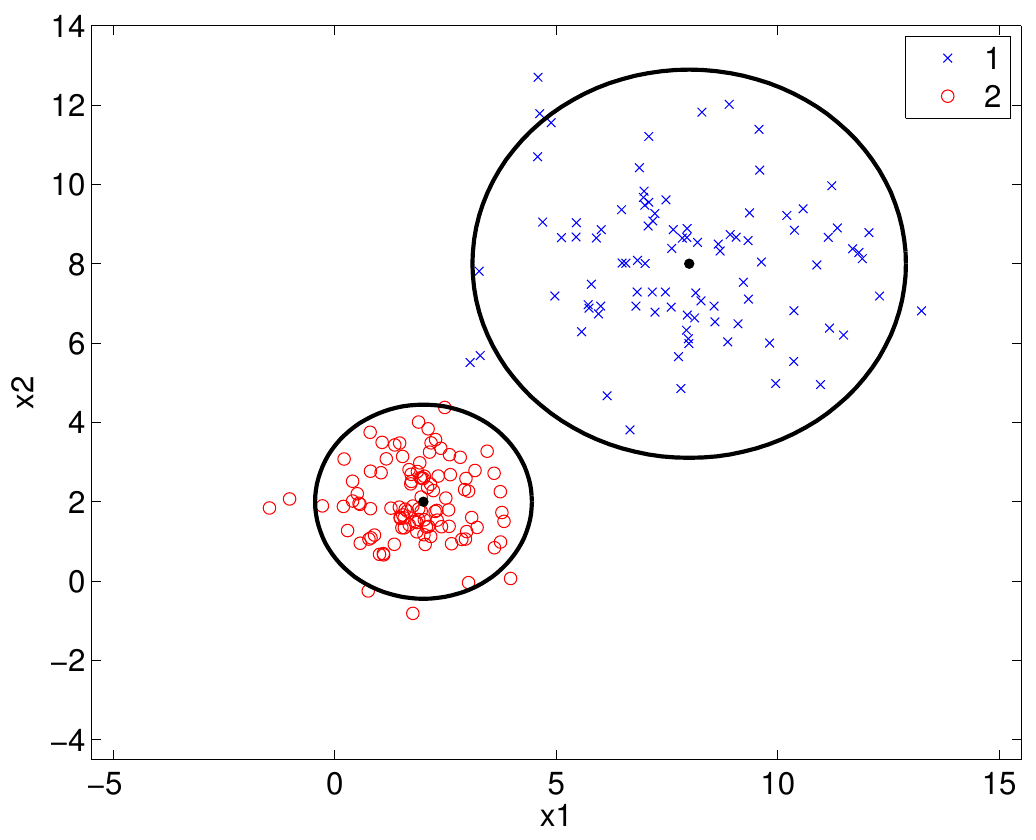}
 \caption{\label{fig: bensmail simulated 2 class data set}A two-class data set simulated according to  $\lambda_k \bI$, and the actual partition.}
 \end{figure}

\begin{figure*}[!ht]
\centering 
{\footnotesize
\hspace*{-1cm}
\begin{tabular}{cccc}
\includegraphics[scale=.2]{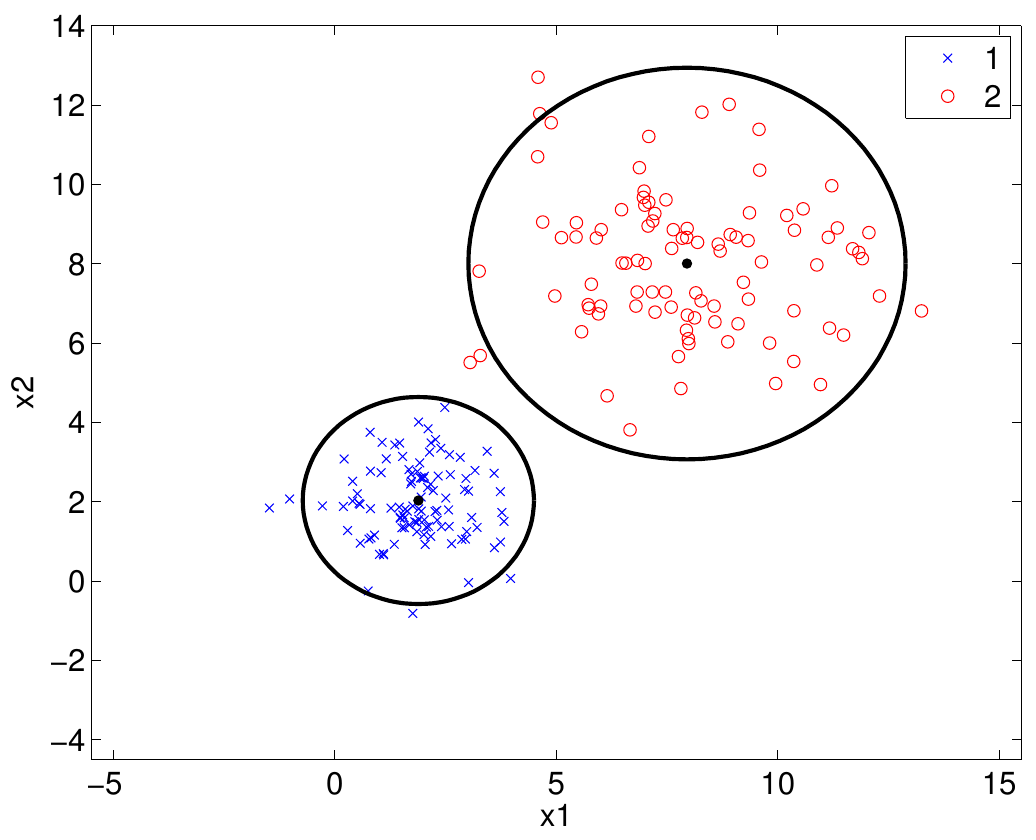} &
   \includegraphics[scale=.2]{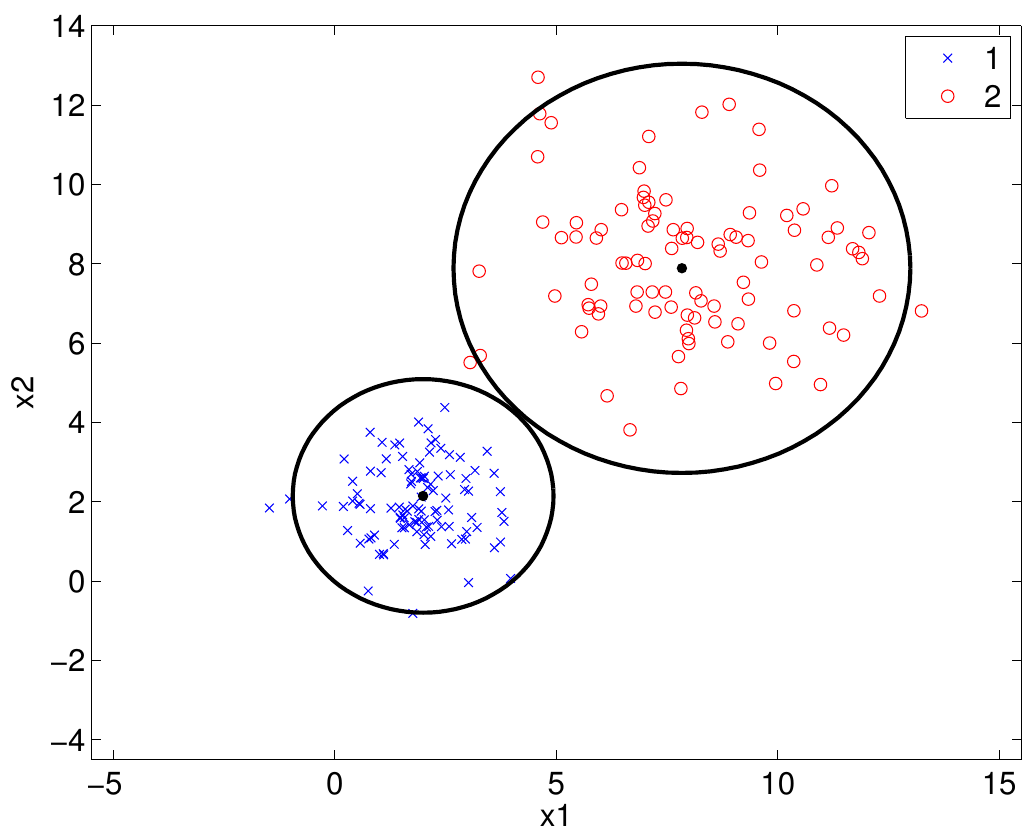} &
   \includegraphics[scale=.2]{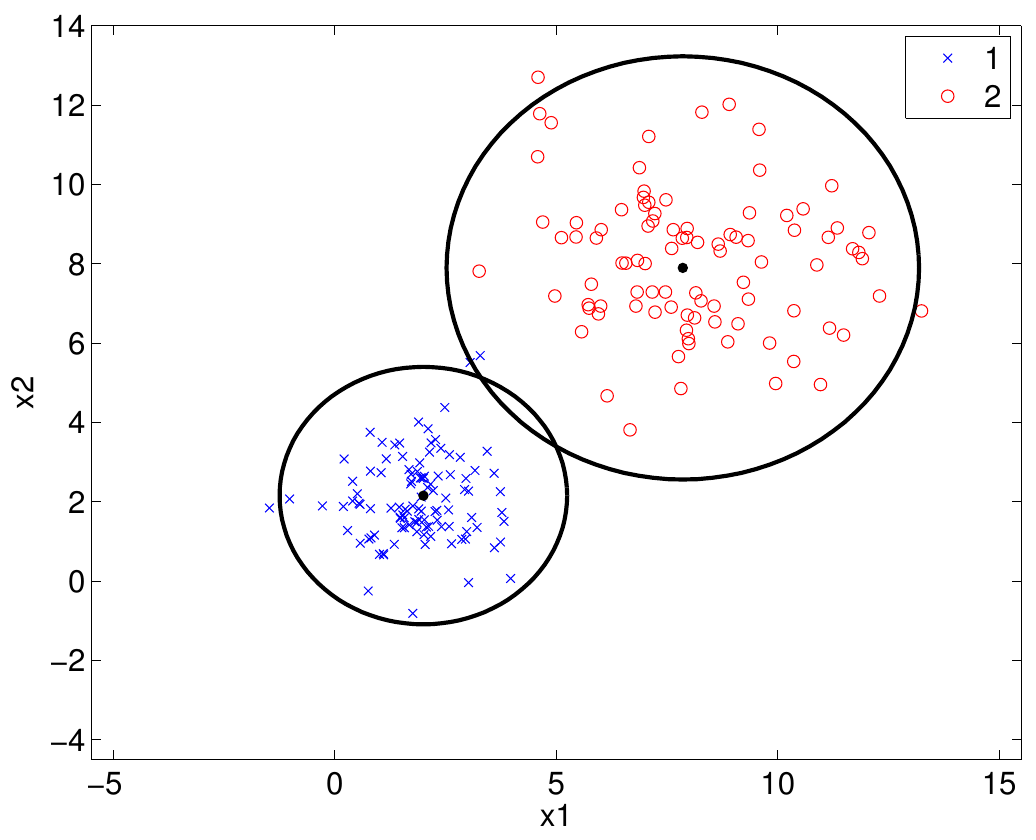} &
   \includegraphics[scale=.2]{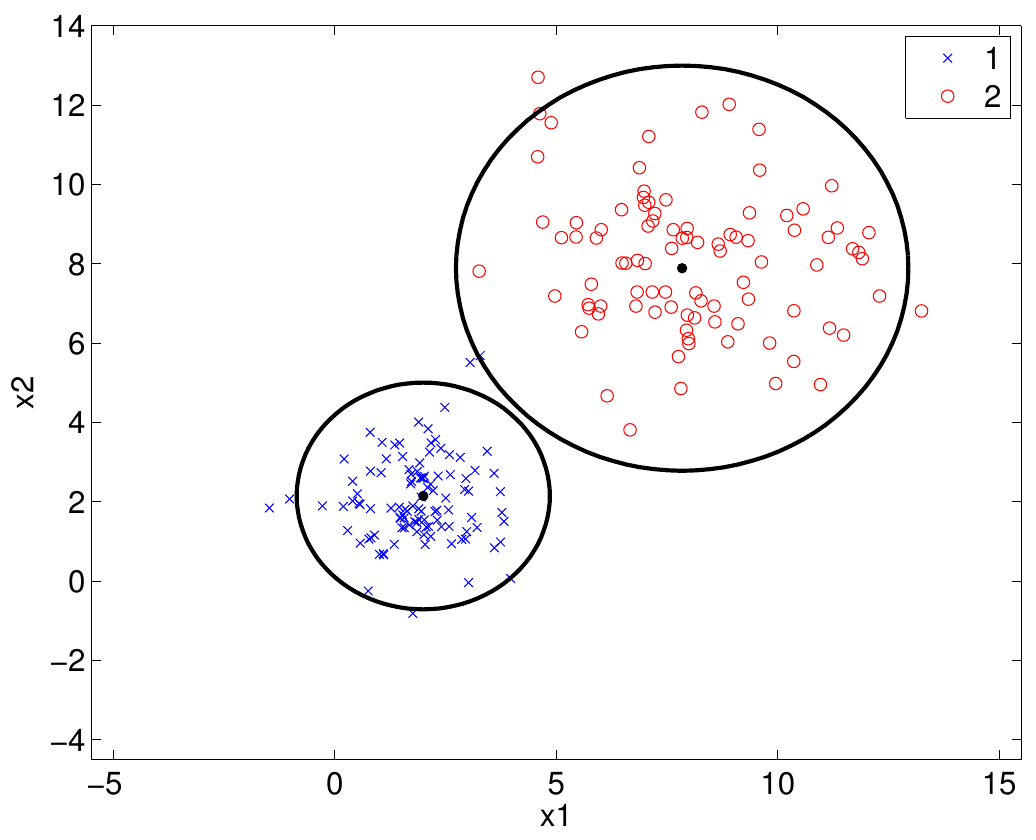}\\
  \includegraphics[scale=.2]{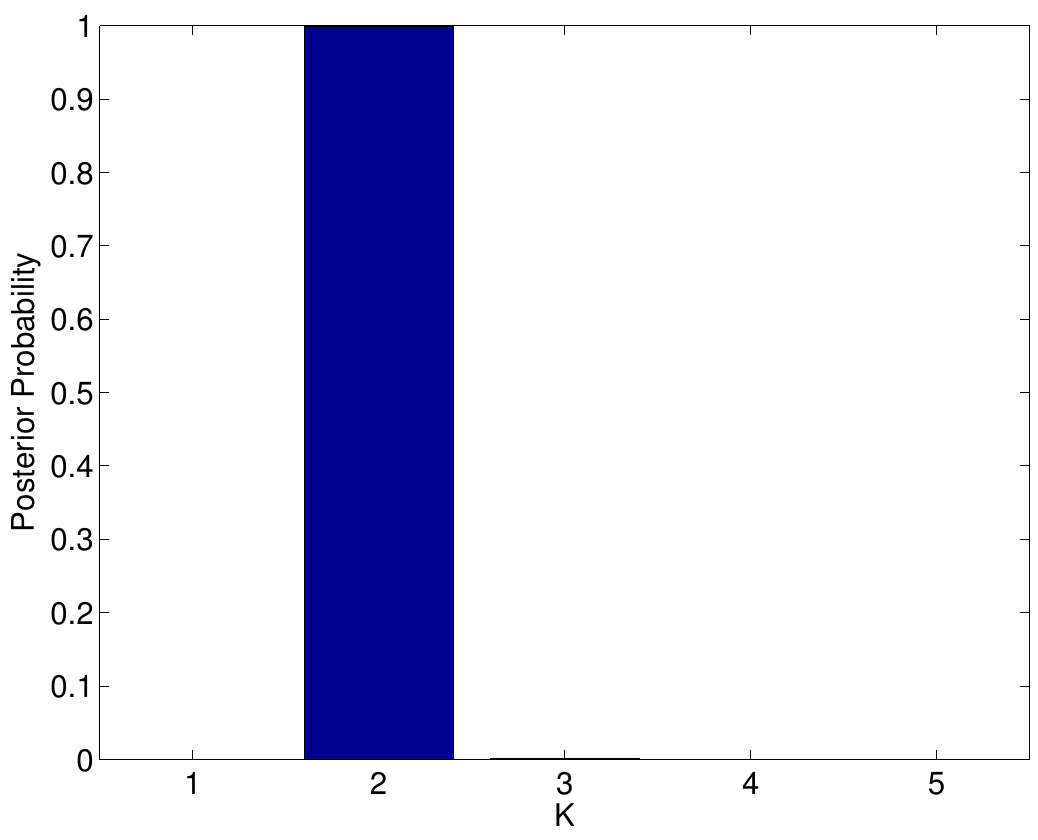}   &
  \includegraphics[scale=.2]{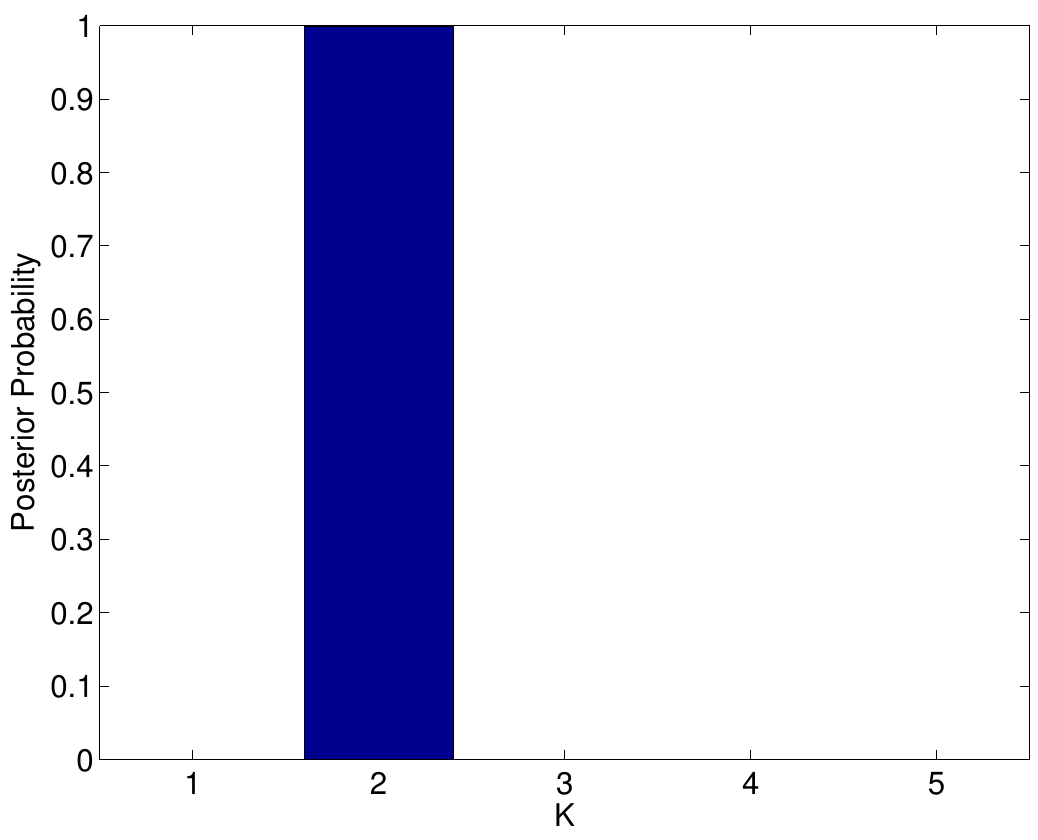}   &
  \includegraphics[scale=.2]{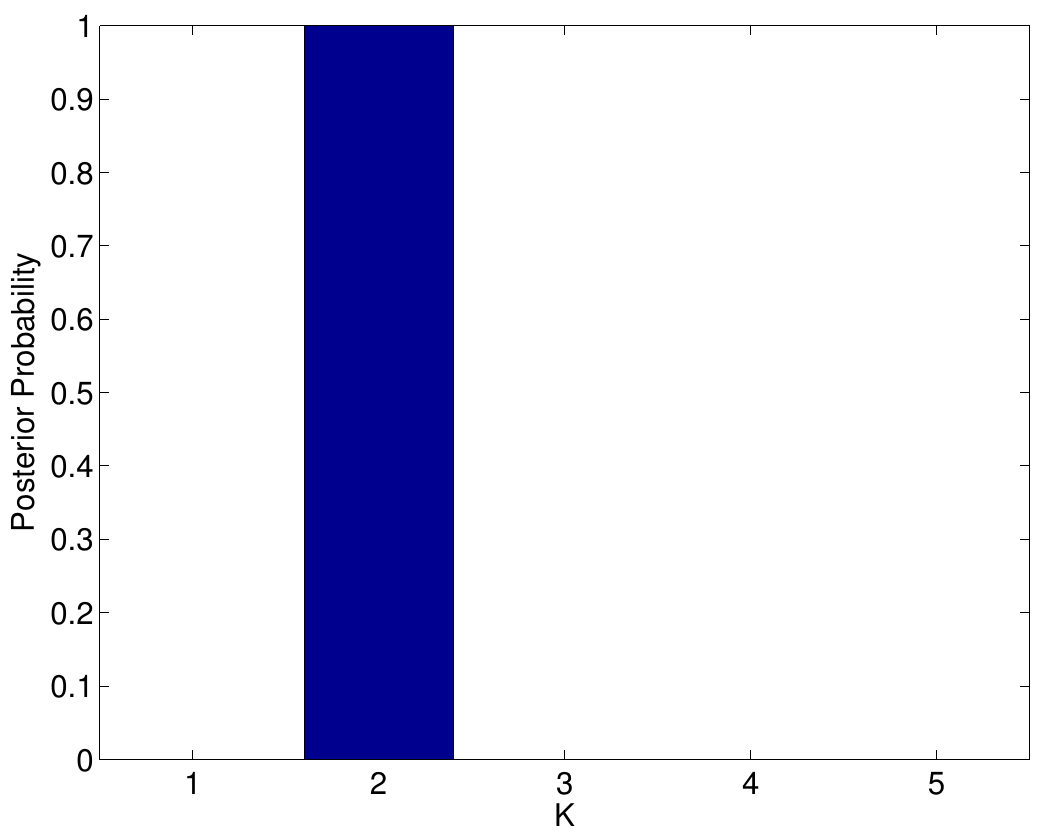} &
  \includegraphics[scale=.2]{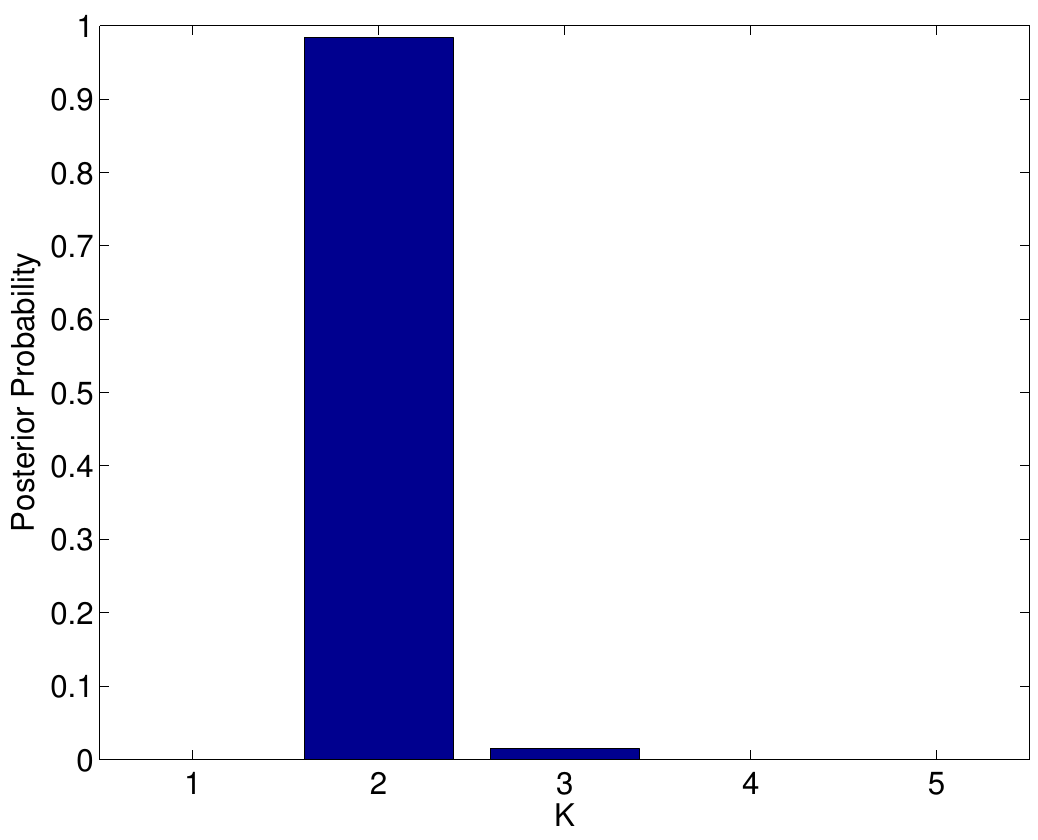}\\ 
  Situation 1 & Situation 2 & Situation 3 & Situation 4\\
  \end{tabular}
  \hspace*{-1cm}
   \caption{\label{fig: estimated partitions sensitivity 4 situations}{Best estimated partitions 
   obtained by the proposed $\lambda_k \bI$ DPPM for the four situations of of hyperparameters values. 
   }}
 }
\end{figure*}

In order to illustrate the effect of the choice of the hyperparameters values of the mixture on the estimations, 
we considered  two-class situations identical to those used in the parametric parsimonious mixture approach proposed in \cite{Bensmail-model-based-clust97}. 
The data set consists in a sample of $n=200$ observations from a two-component Gaussian mixture in $\R^2$ with the following parameters: $\pi_1=\pi_2=0.5$, $\bsmu_1 = (8,8)^T$ and $\bsmu_2 = (2,2)^T$, and two spherical covariances with different volumes $\bsSigma_1 = 4 \ \Identity_2$ and $\bsSigma_2 = \Identity_2$.
 In Figure (\ref{fig: bensmail simulated 2 class data set}) we can see a simulated data set from this experiment with the corresponding actual partition and density ellipses.
%
In order to assess the stability of the models with respect to the choice of the hyperparameters, 
we consider four situations with different hyperparameter values. In these situations,  the hyperparameters $\nu_0$ 
and $\bsmu_0$ are assumed to be the same for the four situations and their values are respectively  $\nu_0=d+2=4$ (related to the number of degrees of freedom) and $\bsmu_0$ is equal to the empirical mean vecotr of the data. 
We varied the two hyperparameters, $\kappa_0$ that controls the prior over the mean  and $s_0^2$ that controls the covariance. The considered four situations  are shown in Table \ref{tab: hyperparameters variation in simulation}. 
\begin{table}[ht!]
\centering
{\footnotesize
\hspace*{-1cm}
\begin{tabular}{|c||c|c|c|c|}
\hline
 Sit. & $1$ & $2$ & $3$ & $4$  \\
 \hline
 \hline
 $s_0^2$ & $\max(\text{eig}(\text{cov}(\bX)))$ & $\max(\text{eig}(\text{cov}(\bX)))$ & 4 $\max(\text{eig}(\text{cov}(\bX)))$ & $\max(\text{eig}(\text{cov}(\bX)))/4$\\
 \hline
 $\kappa_0$&$1$ & $5$ &$5$& $5$\\
 \hline
\end{tabular}
\hspace*{-1cm}
\caption{\label{tab: hyperparameters variation in simulation}{Four different situations the hyperparameters values.}}
}
\end{table}We consider and compare four models corresponding to the spherical, diagonal and general family, which are $\lambda \bI$, $\lambda_k \bI$, $\lambda_k \bA$ and $\lambda_k \bD \bA \bD^T$.
%
Table \ref{tab: logarithm of marginal likelihood 4 situation hyperparams} shows the obtained log marginal likelihood values for the four models for each of the situations of the hyperparameters. 
One can see that, for all the situations, the selected model is $\lambda_k \bI$, that is the one that corresponds to the actual model, and has the correct number of clusters (two clusters).
\begin{table}[ht]
 \centering
{\scriptsize
        \begin{tabular}{|c|c|c|c|c|c|c|c|c|}
     \hline
Model & \multicolumn{2}{c|}{$\lambda \bI$} & \multicolumn{2}{c|}{$\lambda_k \bI$} & \multicolumn{2}{c|}{$\lambda \bA$} & \multicolumn{2}{c|}{$\lambda_k \bD \bA \bD^T$} \\  
\hline
 Sit. 
  &$\hat{K}$&  $\log\text{ML}$& $\hat{K}$ &$\log\text{ML}$& $\hat{K}$ & $\log\text{ML}$ & $\hat{K}$ & $\log\text{ML}$ \\
 \hline
 \hline
1
& 2  &-919.3150& 2 & \bf{-865.9205} &3 &-898.7853 & 3& -885.9710\\
\hline
2
& 3 &-898.6422& 2 & \bf{-860.1917} & 2&-890.6766 &2 & -885.5094\\
\hline
3
& 2 &-927.8240& 2 & \bf{-884.6627} &2 &-906.7430 &2 & -901.0774\\
\hline
4
& 2 &-919.4910& 2 & \bf{-861.0925} &2 &-894.9835 &2 &-889.9267\\
 \hline
 \end{tabular}
 \caption{\label{tab: logarithm of marginal likelihood 4 situation hyperparams}Log marginal likelihood values for the proposed DPPM for 4 situations of hyperparameters values.}
 }
 \end{table}Also, it can be seen from Table \ref{table:bf values for generated data 4 situation hypermarams}, that the Bayes factor values ($2\log\text{BF}$), between the selected model, and the more competitive one, for each of the four situations, according to Table \ref{tab:bf_interpretation}, corresponds to a decisive   evidence of the selected model.
\begin{table}[!ht]
{\footnotesize
\centering
\begin{tabular}{|c|c|c|c|c|}
\hline 
 Sit. & 1 & 2 & 3 & 4\\
\hline
\hline
$2\log\text{BF}$ & 40.10 & 50.63  & 32.82 & 57.66 \\ 
\hline
\end{tabular}
\caption{\label{table:bf values for generated data 4 situation hypermarams}Bayes factor values for the proposed DPPM computed from Table \ref{tab: logarithm of marginal likelihood 4 situation hyperparams} 
by comparing the selected model ($M_1$, here in all cases $\lambda_k \bI$), and the one more competitive for it ($M_2$, here in all cases $\lambda_k \bD \bA \bD$).
}}
\end{table}  
These results confirm the stability of the DPPM with respect to the variation of the hyparameters values. 
Figure \ref{fig: estimated partitions sensitivity 4 situations} shows the best estimated partitions  obtained by the proposed DPPM for the generated data. 
Note that,  for the all the four situations, the estimated number of clusters is $2$, which corresponds to the mode of the posterior distribution of the number of clusters, with a probability very close to one.
 
\subsection{Experiments on real data}
To confirm the results previously obtained on simulated data, we have conducted several  experiments freely available real data sets: Iris, Old Faithful Geyser, 
Crabs 
and Diabetes 
whose characteristics are summarized in Table \ref{tab: used benchmarks}.
%
We compare the proposed DPPM models to the PGMM models. 
 \begin{table}[!ht]
 \centering
 {
 \begin{tabular}{l c c c}
 \hline
 Dataset & \# data ($n$) & \# dimensions ($d$) & True \#  clusters ($K$)\\
 \hline
\hline
Old Faithful Geyser& 272 & 2 & Unknown \\
Crabs& 200 & 5 &  2 \\
Diabetes & 145 & 3 & 3\\
Iris	& 150 & 4 & 3\\
\hline
 \end{tabular}
 \caption{\label{tab: used benchmarks}Description of the used real data sets.}
 }
 \end{table}


\vspace{-0.5cm}
\subsubsection{Old Faithful Geyser data set}
The Old Faithful geyser data set \citep{Azzalini1990} comprises $n=272$ measurements of the eruption of the Old Faithful geyser at Yellowstone National Park in the USA. Each measurement is bi-dimensional ($d=2$) and comprises the duration of the eruption and the time to the next eruption, both in minutes. While the number of clusters for this data set  is unknown, several clustering studies in the literature estimate at two, 
often interpreted as short and long eruptions.  
 
 We applied the proposed DPPM approach and the PGMM alternative to this data set (after standardization). For the PGMM, the value of $K$ varied from 1 to 6.  
Table \ref{table:marginal likelihood faithful data} reports the log marginal likelihood values obtained by the PGMM and the proposed DPPM for the Faithful Geyser data set. 
\begin{table}[!ht]
  \centering
{\scriptsize
\hspace*{-0.5cm}
\begin{tabular}{|c|c|c|c|c|c|c|c|c|}
   \hline &\multicolumn{2}{c|}{DPPM} &\multicolumn{6}{c|}{PGMM}\\
  \hline
  $\text{Model}$ &$\hat{K}$ & $\log\text{ML}$& $K=1$ & $K=2$ & $K=3$ & $K=4$ & $K=5$ & $K=6$\\
  \hline
  \hline
 $\lambda \Identity$		&2&-458.19&-834.75&-455.15&-457.56&-461.42&\bf{-429.66}&  -1665.00\\
 $\lambda_k \Identity$		&2&-451.11&-779.79&\bf{-449.32}&-454.22& -460.30&-468.66&-475.63\\
 $\lambda \bA$			&3&-424.23&-781.86&\bf{-445.23}&-445.61&-445.63&-448.93&-453.44\\
 $\lambda_k \bA$		&2&-446.22&-784.75&\bf{-461.23}&-465.94&-473.55& -481.20&-489.71\\
 $\lambda \bD \bA \bD^T$	&2&\cellcolor{gray!25}\bf{-418.99}&-554.33&\bf{-428.36}&-429.78&-433.36&-436.52&-440.86\\
 $\lambda_k \bD \bA \bD^T$	&2& -434.50&-556.83&\underline{\bf{-420.88}}&-421.96&-422.65&-430.09&-434.36\\
 $\lambda \bD_k \bA \bD_k^T$	&2&-428.96& -780.80&-443.51&\bf{-442.66}&-446.21& -449.40&-456.14\\
 $\lambda_k \bD_k \bA \bD_k^T$	&2&-421.49&-553.87&-434.37&\bf{-433.77}& -439.60&-442.56&-447.88\\
 
  \hline
  \end{tabular}
  \caption{\label{table:marginal likelihood faithful data}{\footnotesize Log marginal likelihood  values 
  for the Old Faithful Geyser data set.}}
  }
  \end{table}One can see that the parsimonious DPPM models estimate $2$ clusters except one model, which is the diagonal model with equal volume $\lambda \bA$ that estimates three clusters. 
  For a number of clusters varying from 1 to 6, the parsimonious PGMM models estimate two clusters at three exceptions, including the spherical model $\lambda \bI$ which  overestimates the number of clusters (provides 5 clusters). However, the solution provided by the proposed DPPM for the spherical model $\lambda \bI$
  is  more stable and estimates two clusters. 
It can also be seen that the best model with the highest value of the log marginal likelihood is the one provided by the proposed DPPM and corresponds to the general  model $\lambda \bD \bA \bD^T$ with equal volume and the same shape and orientation.   
On the other hand, it can also be noticed that, in terms of Bayes factors, the model $\lambda \bD \bA \bD^T$ selected by the proposed DPPM has a decisive evidence compared to the other models, and a strong evidence (the value of $2 \log \text{BF}$ equals $5$), compared to the most competitive one, which is in this case the model $\lambda_k \bD_k \bA \bD_k^T$. 

\begin{figure*}[!ht]
\centering
{\footnotesize 
\hspace*{-0.5cm}
\begin{tabular}{ccc} 
 \includegraphics[scale=.25]{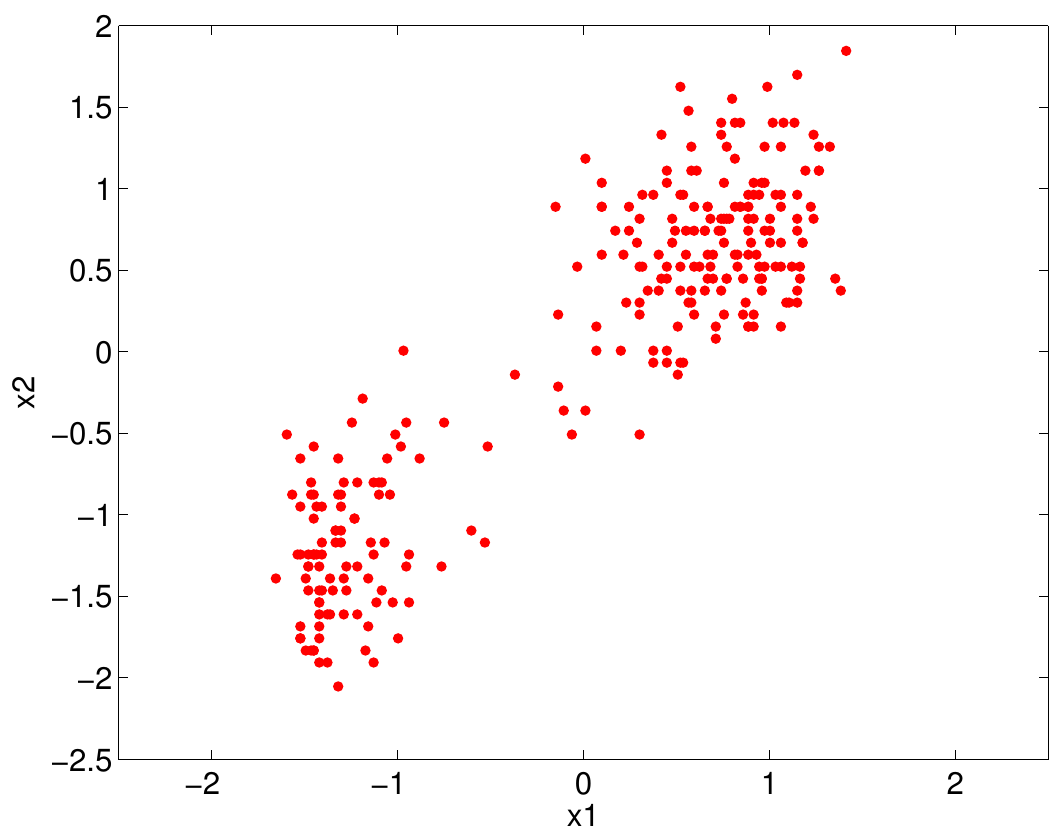} &
 \includegraphics[scale=.25]{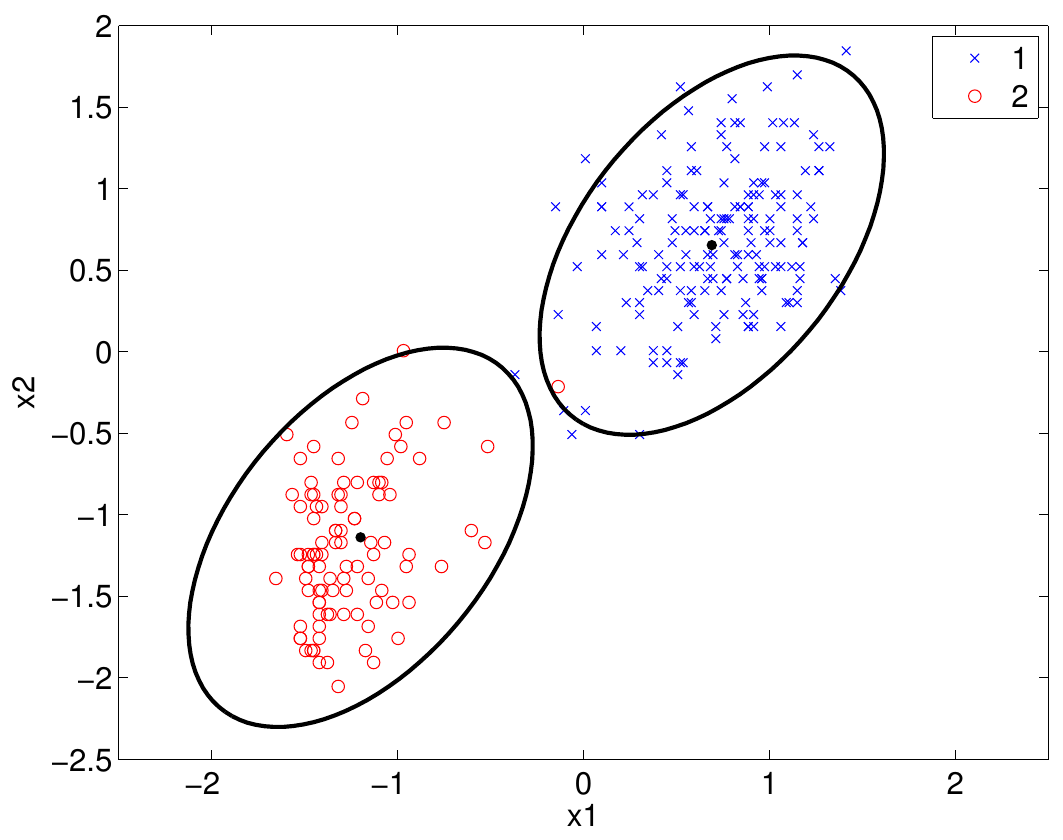} &
 \includegraphics[scale=.25]{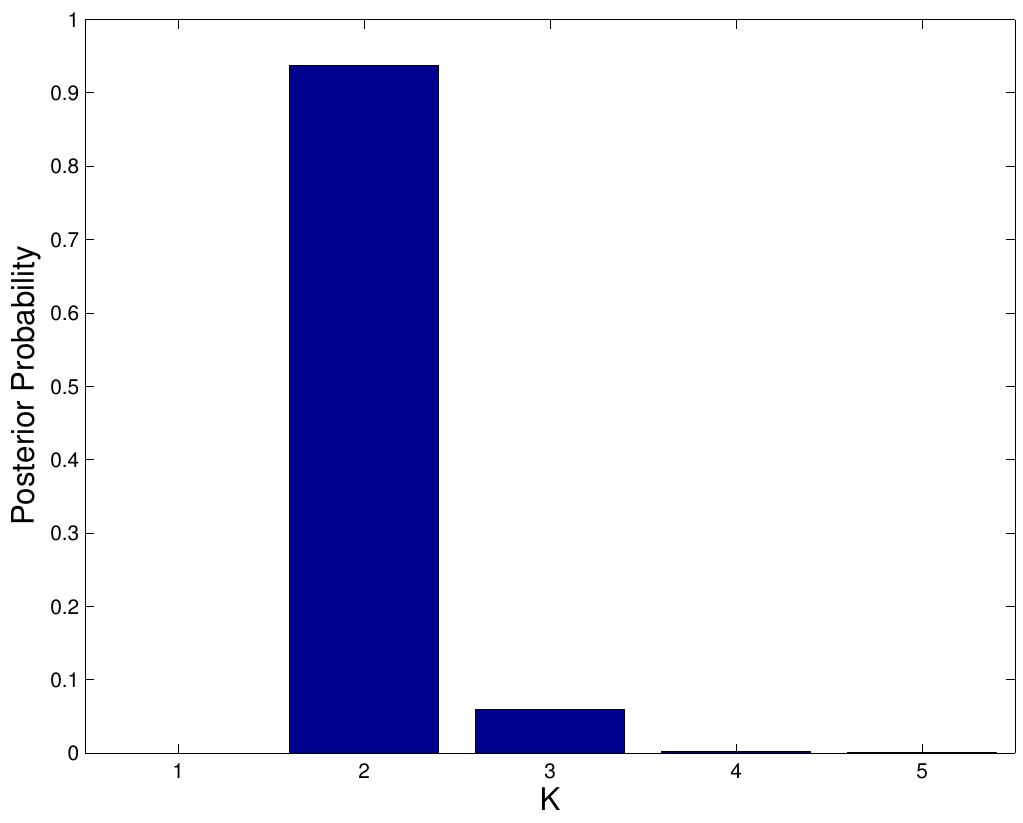} 
 \end{tabular} 
  \caption{\label{fig: best DPPM partition for Geyser}{Old Faithful Geyser data set (left), the optimal partition obtained by the DPPM  model $\lambda \bD \bA \bD^T$ (middle) and the empirical posterior distribution for the number of mixture components (right).}}
  }
 \end{figure*}

Figure \ref{fig: best DPPM partition for Geyser} shows the the optimal partition 
and the posterior distribution for the number of clusters. One can namely observe that the likely partition is provided with a number of cluster with high posterior probability (more than 0.9).
 
 Table \ref{tab: cpu for Geyser} shows the mean computer running time, measured in seconds, for the Gibbs inference of each DPPM models.
  \begin{table}[!ht]
   \centering
   \hspace*{-0.5cm}
   {\scriptsize
   \begin{tabular}{|c|c|c|c|c|c|c|c|c|}
    \hline
   $\text{Model}$&$\lambda \Identity$&$\lambda_k \Identity$&$\lambda \bA$ &$\lambda_k \bA$&$\lambda \bD \bA \bD^T$&$\lambda_k \bD \bA \bD^T$&$\lambda \bD_k \bA \bD_k^T$&$\lambda_k \bD_k \bA \bD_k^T$\\
   \hline
   \hline
   \text{CPU time (s)}& 953.86& 785.36 &  999.91 & 964.86 &  901.44 & 717.28&  1020&810.23 \\
   \hline
   
   \end{tabular}
   \caption{\label{tab: cpu for Geyser}{The DPPM Gibbs sampler mean CPU time (in seconds) for each parsimonious model on Old Faithful Geyser data set.}}
 }
  \end{table}

\subsubsection{Crabs data set}
%
The Crabs data set comprises $n=200$ observations  describing $d=6$ morphological measurements (Species, Frontal lip, Rearwidth, Length, Width Depth) on $50$ crabs each of two colour forms and both sexes, of the species Leptograpsus variegatus collected at Fremantle, W. Australia \citep{Campbell1974}. The Crabs are classified according to their sex ($K=2$).
%
%
%
We applied the proposed DPPM approach and the PGMM alternative to this data set (after PCA and standardization). For the PGMM the value of $K$ varied from 1 to 6. Table \ref{table:marginal likelihood crabs data} reports the log marginal likelihood values obtained by the PGMM the proposed DPPM approaches for the Crabs data set. 
 \begin{table}[!ht]
   \centering
  {\scriptsize
  \hspace*{-0.5cm}
          \begin{tabular}{|c|c|c|c|c|c|c|c|c|}
    \hline &\multicolumn{2}{c|}{DPPM} &\multicolumn{6}{c|}{PGMM}\\
   \hline
   $\text{Model}$ &$\hat{K}$ & $\log\text{ML}$& $K=1$ & $K=2$ & $K=3$ & $K=4$ & $K=5$ & $K=6$ \\
   \hline
   \hline 
  $\lambda \Identity$	        & 3 & -550.75       & -611.30   &  -615.73   &  \bf{-556.05}  & -860.95  & -659.93   &  -778.21 \\
  $\lambda_k \Identity$	        & 3 & -555.91       &-570.13   &  -549.06   &   -538.04 & -542.31  & -577.22   &  \bf{-532.40} \\
  $\lambda \bA$		            & 4 & -537.81       &-572.06   &  -539.17   &   -532.65   & -535.20   & -534.43   &  \bf{-531.19} \\
  $\lambda_k \bA$	            & 3 & -543.97       &-574.82   &  \bf{-541.27} &  -569.79    & -590.48  & -693.42   &  -678.95 \\
  $\lambda \bD \bA \bD^T$       & 4 & -526.87       & -554.64  &  -540.87  &   \bf{ -512.78}& -525.19  & -541.93   &  -576.27 \\
  $\lambda_k \bD \bA \bD^T$	    & 3 & -517.58       &-556.73   &  -541.88  &  \bf{-515.93}& -530.02  & -550.71   &  -595.38 \\
  $\lambda \bD_k \bA \bD_k^T$   & 4 & -549.78       & -573.80   &  -564.28  &  -541.67    & -547.45  & -547.13   &  \bf{-526.79} \\
  $\lambda_k \bD_k \bA \bD_k^T$ & 2 &\cellcolor{gray!25}\bf{ -499.54} & -557.69   &   \underline{\bf{-500.24}}  &    -700.44  &    -929.24   &   -1180.10  &    -1436.60\\
   \hline
   \end{tabular}
   \caption{\label{table:marginal likelihood crabs data}{\footnotesize 
   Log marginal likelihood values 
   for the Crabs data set.}}
   }
   \end{table}%
One can first see that  the best solution corresponding to the best model with the highest value of the log marginal likelihood is the one provided by the proposed DPPM and corresponds to the general model $\lambda_k \bD_k \bA \bD_k^T$ with different volume and orientation but equal shape. This model provides a partition with a number of clusters equal to the actual one $K=2$.
One can also see that the best solution for the PGMM approach is the one provided by the same model with a correctly estimated number of clusters.
On the other hand, one can also see that for this Crabs data set, the proposed DPPM models estimate the number of clusters between $2$ and $4$. 
This may be related to the fact that, for the Crabs data set, the data, in addition their sex, are also described in terms of their specie and the data contains two species. This may therefore result in four subgroupings of the data in four clusters, each couple of them corresponding to two species, and the solution of four clusters may be plausible for this data set.
However three PGMM models overestimate the number of clusters  and provide solutions with $6$ clusters. 
%
%
We can also observe that, in terms of Bayes factors, the model $\lambda_k \bD_k \bA \bD_k^T$ selected by the proposed DPPM for this data set, has a decisive evidence compared to all the other potential models. For example the value of $2 \log \text{BF}$ for this selected model, against to the most competitive one, which is in this case the model  $\lambda_k \bD \bA \bD^T$ equals $36.08$ and corresponds to a decisive  evidence of the selected model. 
   
The good performance of the DPPM compared the PGMM is also confirmed in terms of Rand index and misclassification error rate values. 
The optimal partition obtained by the proposed DPPM with the parsimonious model $\lambda_k \bD_k \bA \bD_k^T$ is the best defined one and corresponds to the highest  Rand index value of $0.8111$ and the lowest error rate of $10.5 \pm 1.98$. However, the partition obtained by the PGMM has a Rand index of $0.8032$ with an error rate of $11 \pm 2.07$. 



\begin{figure*}[!ht]
  \centering
  \hspace*{-0.5cm}
  {\footnotesize 
\begin{tabular}{ccc} 
\includegraphics[scale=.25]{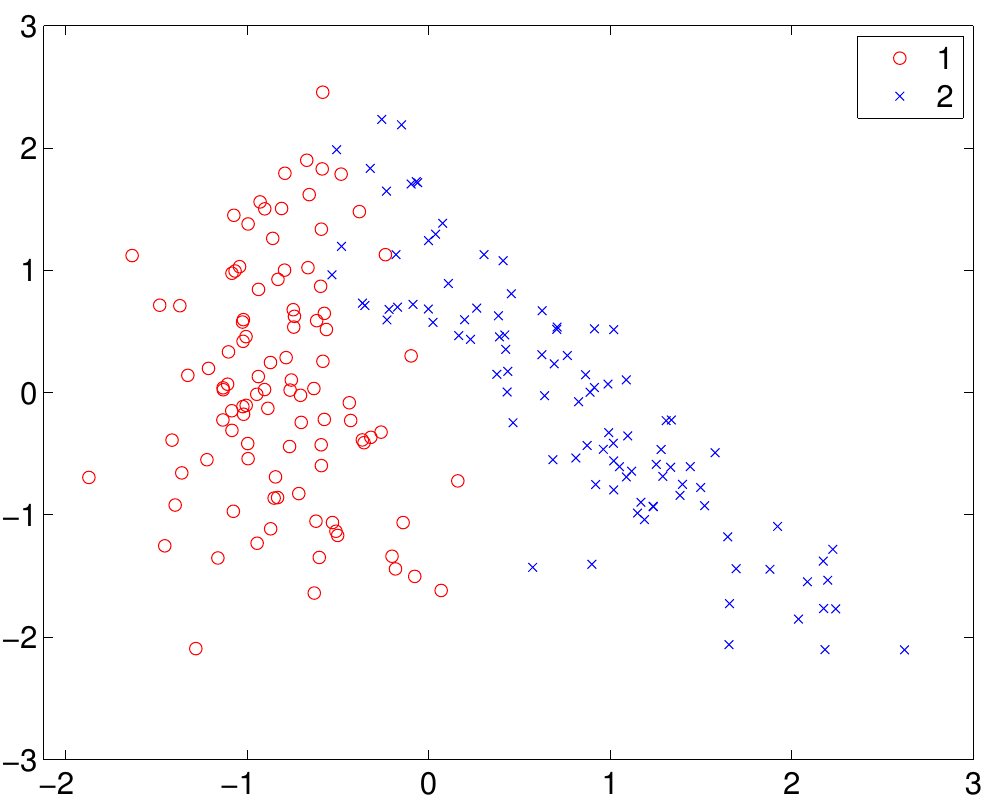}&
  \includegraphics[scale=.25]{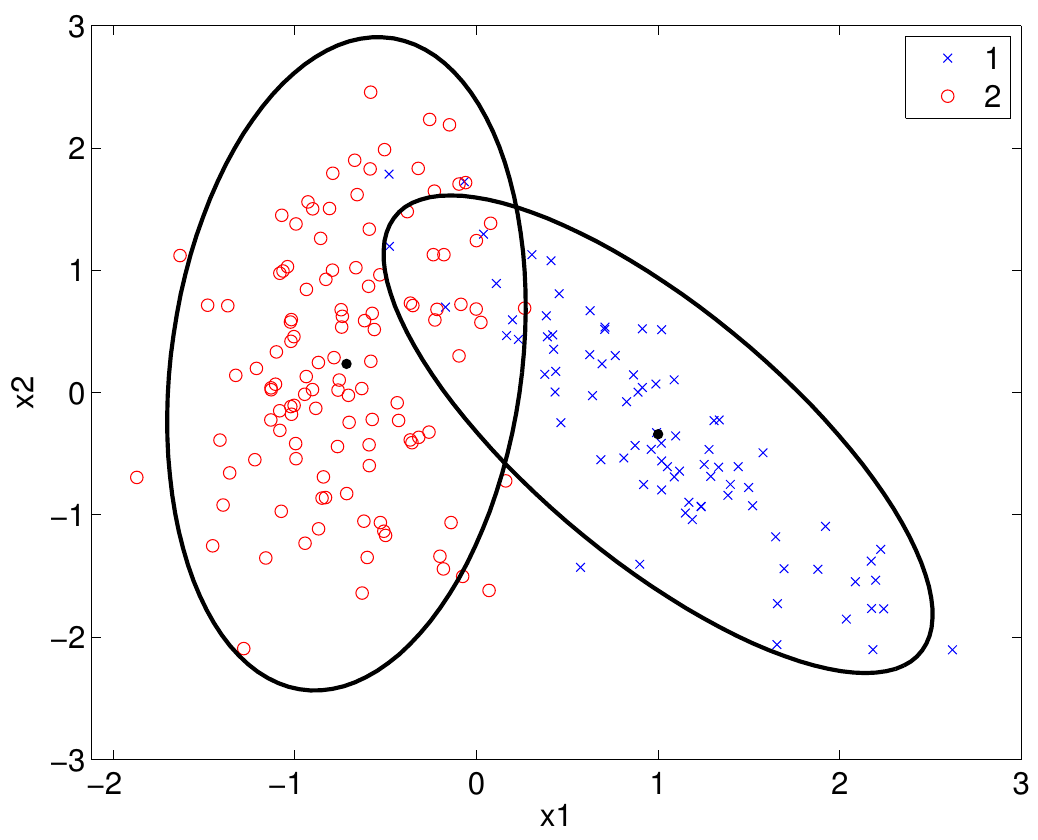}&
  \includegraphics[scale=.25]{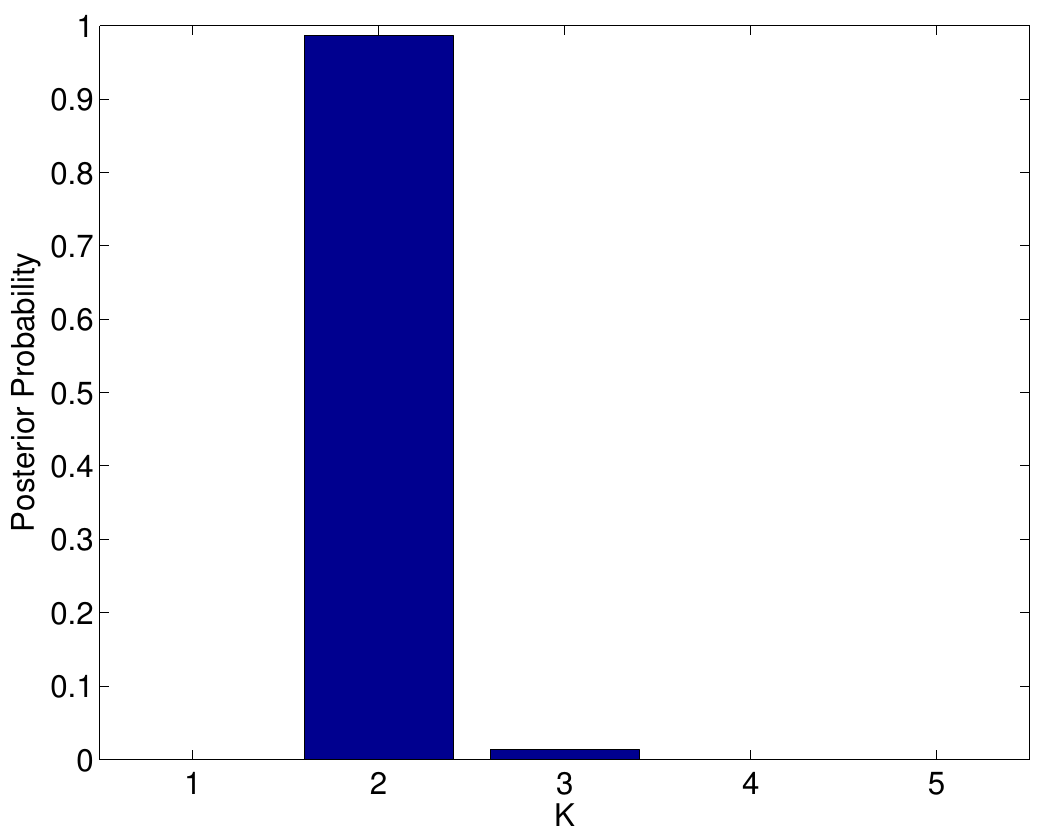}
  \end{tabular}
 \caption{\label{fig: best DPPM partition for Crabs}{Crabs data set in the two first principal axes and the actual partition (left), the optimal partition obtained by the DPPM  model $\lambda_k \bD_k \bA \bD_k^T$ (middle) and the empirical posterior distribution for the number of mixture components (right).}}
 }
\end{figure*}

Figure \ref{fig: best DPPM partition for Crabs} 
shows the optimal partition and the posterior distribution for the number of clusters for the Crabs data. 
One can observe that the provided partition is quite precise 
and is provided with a number of clusters equal to the actual one, and with a posterior probability very close to $1$.

Table \ref{tab: cpu for Crabs} shows the mean computer running time, measured in seconds, for the Gibbs inference of each DPPM models.
  \begin{table}[!ht]
   \centering
   \hspace*{-0.5cm}
    {\scriptsize
   \begin{tabular}{|c|c|c|c|c|c|c|c|c|}
    \hline
   $\text{Model}$&$\lambda \Identity$&$\lambda_k \Identity$&$\lambda \bA$ &$\lambda_k \bA$&$\lambda \bD \bA \bD^T$&$\lambda_k \bD \bA \bD^T$&$\lambda \bD_k \bA \bD_k^T$&$\lambda_k \bD_k \bA \bD_k^T$\\
   \hline
   \hline
   \text{CPU time (s)}&263.39 &318.06 &423.51 & 412.29& 399.91&399.50 &445.67 & 442.29\\
   \hline
   
   \end{tabular}
   \caption{\label{tab: cpu for Crabs}{The DPPM Gibbs sampler mean CPU time (in seconds) for each parsimonious model on Crabs dataset.}}
 }
  \end{table}
\vspace{-0.5cm}
\subsubsection{Diabetes data set}

The Diabetes data set was described and analysed in \citep{Diabetes_dataset79} consists of $n=145$ subjects, describing $d=3$ features: the area under a plasma glucose curve (glucose area), the area under a plasma insulin curve (insulin area) and the steady-state plasma glucose response (SSPG). This data has $K=3$ groups: the chemical diabetes, the overt diabetes and the normal (nondiabetic). 
%
We applied the proposed DPPM models and the alternative PGMM ones on this data set (the data was standardized). For the PGMM, the number of clusters varied from 1 to 8. 

Table \ref{table:marginal likelihood diabetes data} reports the log marginal likelihood values obtained by the two approaches for the Crabs data set. 
One can see that both the proposed DPPM and the PGMM estimate correctly the true number of clusters. However, the best model with the highest log marginal likelihood value is the one obtained by the proposed DPPM approach and corresponds to  the parsimonious model 
$\lambda_k \bD_k \bA \bD_k^T$ with the actual number of clusters ($K=3$).  
\begin{table}[!ht]
   \centering
   \hspace*{-1.3cm}
  {\scriptsize
          \begin{tabular}{|c|c|c|c|c|c|c|c|c|c|c|}
    \hline &\multicolumn{2}{c|}{DPPM} &\multicolumn{8}{c|}{PGMM}\\
   \hline
   $\text{Model}$ &$\hat{K}$ & $\log\text{ML}$& $K=1$ & $K=2$ & $K=3$ & $K=4$ & $K=5$ & $K=6$ & $K=7$ & $K=8$\\
   \hline
   \hline
  $\lambda \Identity$		&4&-573.73& -735.80&   -675.00&-487.65&-601.38&-453.77&-468.55&\bf{-421.33}&-533.97\\
  $\lambda_k \Identity$		&7&-357.18&-632.18&-432.02&-412.91&-417.91&-398.02&-363.12&\bf{-348.67}&-378.48\\
  $\lambda \bA$			&8& -536.82& -635.70&-492.61&-488.55&-418.51&-391.05&-377.37&-370.47&\bf{-365.56}\\
  $\lambda_k \bA$		&6&-362.03&-638.69&-416.27&-372.71&\bf{-358.45}&-381.68&-366.15&-385.73&-495.63\\
  $\lambda \bD \bA \bD^T$	&7&-392.67&-430.63&-418.96& -412.70&\bf{-375.37}&-390.06&-405.11&-426.92&-427.46\\
  $\lambda_k \bD \bA \bD^T$	&5&-350.29&-432.85&-326.49&-343.69&\bf{-325.46}& -355.90&-346.91&-330.11&-331.36\\
  $\lambda \bD_k \bA \bD_k^T$	&5&-338.41&-644.06&-427.66&-454.47&-383.53&-376.03&-356.09&-355.03&\bf{-349.84}\\
  $\lambda_k \bD_k \bA \bD_k^T$	&3& \cellcolor{gray!25}\bf{-238.62}&
  -433.61&-263.49&\underline{\bf{-248.85}}&-273.31&-317.81&-440.67& -453.70&-526.52\\
   \hline
   \end{tabular}
   \caption{\label{table:marginal likelihood diabetes data}{Obtained marginal likelihood values 
   for the Diabetes data set.}}
   }
   \end{table}
Also, the evidence of the model $\lambda_k \bD_k \bA \bD_k^T$ selected by the proposed DPPM for the Diabetes data set, compared to all the other models, is decisive. Indeed, in terms of Bayes factor comparison, the value of $2 \log \text{BF}$ for this selected model, against to the most competitive one, which is in this case the model  $\lambda \bD_k \bA \bD_k^T$ is $111.86$ and corresponds to a decisive evidence of the selected model.  
In terms of Rand index, the best defined partition is the one obtained by the proposed DPPM approach with the parsimonious model  $\lambda_k \bD_k \bA \bD_k^T$, which has the highest Rand index value of $0.8081$ which indicates that the partition is well defined, with a misclassification error rate of 
$17.24 \pm 2.47$. 
However, the best PGMM partition  $\lambda_k \bD_k \bA \bD_k^T$ has a Rand index of $0.7615$ with $22.06 \pm 2.51$ error rate.

 
 
\begin{figure*}[!ht]
 \centering
 {\footnotesize
 \begin{tabular}{ccc}
  \includegraphics[scale=.25]{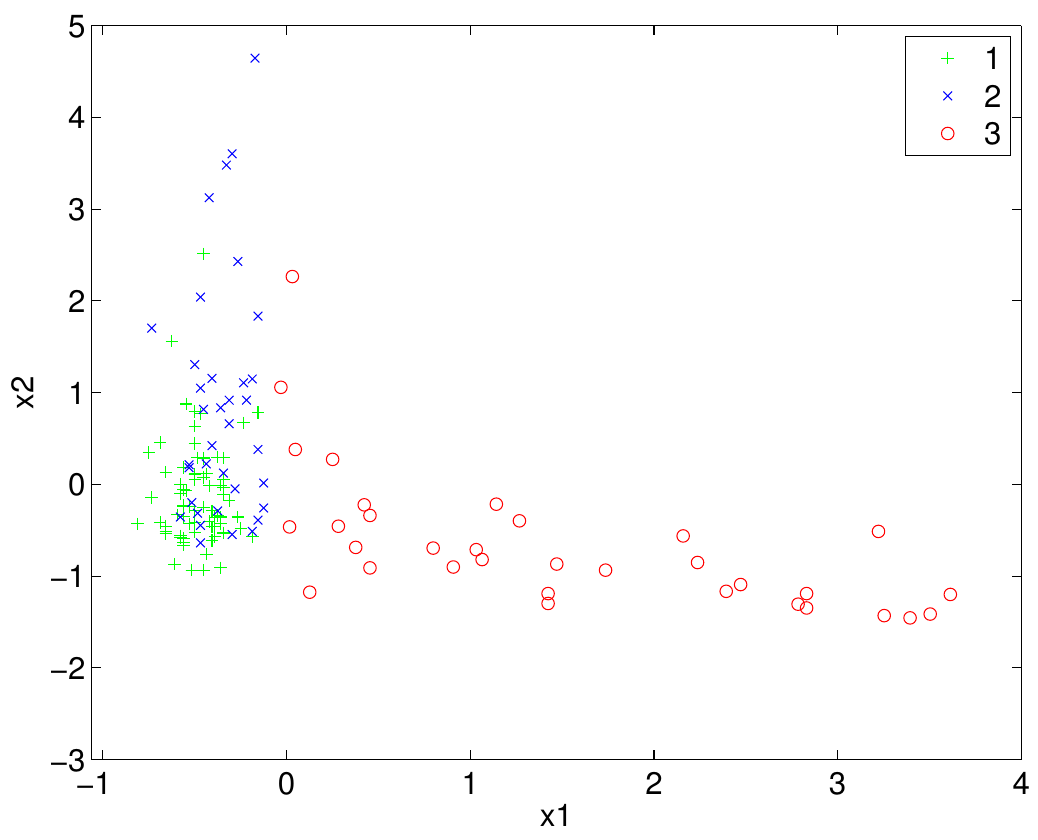}&
 \includegraphics[scale=.25]{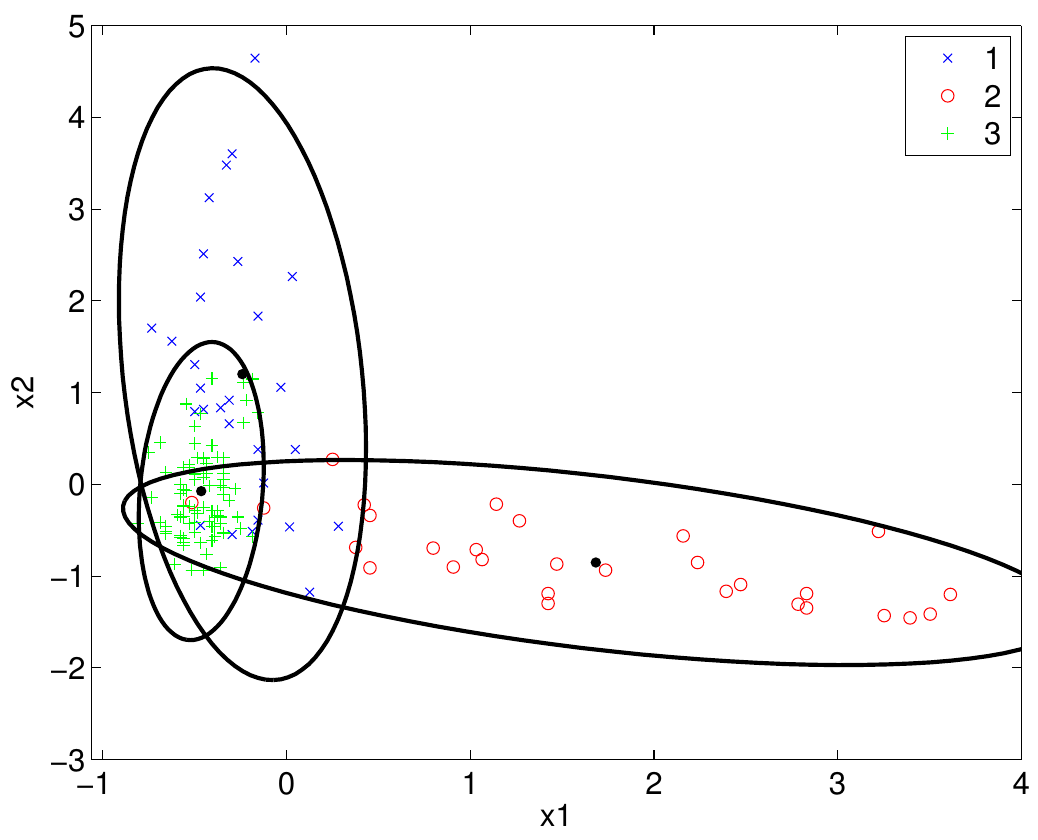}&
 \includegraphics[scale=.25]{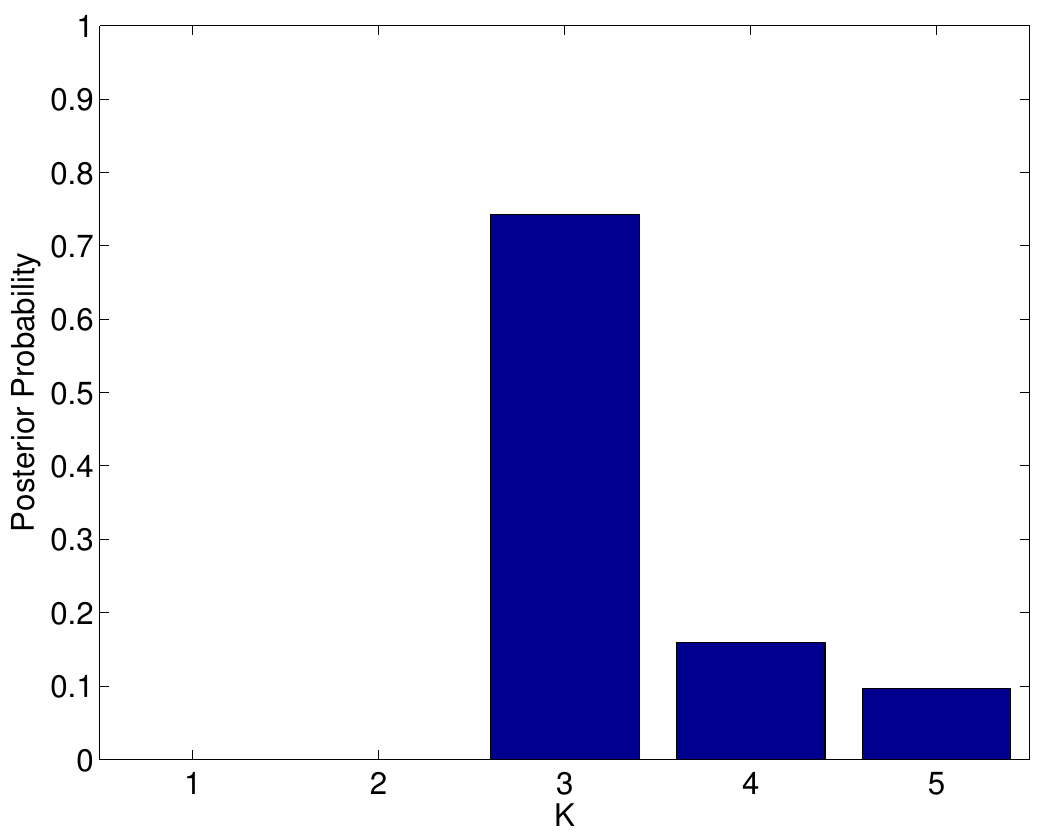}
  \end{tabular}
  \caption{\label{fig: best DPPM partition for Diabetes}{Diabetes data set in the space of the components 1 (glucose area) and 3 (SSPG) and the actual partition (left), the optimal partition obtained by the DPPM  model $\lambda_k \bD_k \bA \bD_k^T$ (middle) and the empirical posterior distribution for the number of mixture components (right).}}
 }
 \end{figure*}

 Figure (\ref{fig: best DPPM partition for Diabetes}) shows the optimal partition provided by the DPPM model $\lambda_k \bD_k \bA \bD_k^T$ and the distribution of the number of clusters $K$. We can observe that the partition is quite well defined (the misclassification rate in this case is $17.24 \pm 2.47$) and the posterior mode of the number of clusters equals the actual number of clusters ($K=3$).
 %
 
 Table \ref{tab: cpu for Diabetes} shows the mean computer running time, measured in seconds, for the Gibbs inference of each DPPM models.
  \begin{table}[!ht]
   \centering
   \hspace*{-0.5cm}
   {\scriptsize
   \begin{tabular}{|c|c|c|c|c|c|c|c|c|}
    \hline
   $\text{Model}$&$\lambda \Identity$&$\lambda_k \Identity$&$\lambda \bA$ &$\lambda_k \bA$&$\lambda \bD \bA \bD^T$&$\lambda_k \bD \bA \bD^T$&$\lambda \bD_k \bA \bD_k^T$&$\lambda_k \bD_k \bA \bD_k^T$\\
   \hline
   \hline
   \text{CPU time (s)}& 1471.7 &1335 & 1664&1386.8 & 1348.6 &715.01&1635 & 1454.4 \\
   \hline
   
   \end{tabular}
   \caption{\label{tab: cpu for Diabetes}{The DPPM Gibbs sampler mean CPU time (in seconds) for each parsimonious model on Diabetes data set.}}
 }
  \end{table}
 \subsubsection{Iris data set}
The well-known  Iris data set of \cite{fisher36} contains measurements for $n=150$ samples of Iris flowers covering three Iris species (setosa, virginica and versicolor) ($K=3$) with $50$ samples for each specie. Four features were measured for each sample ($d=4$): the length and the width of the sepals and petals, in centimetres. 
We applied PGMM models and the proposed DPPM models on this data set. For the PGMM models, the number of clusters $K$ was tested in the range $[1; 8]$. 

Table \ref{table:marginal likelihood iris data} reports the obtained log marginal likelihood values. 
We can see that the best solution  is the one of the proposed DPPM and corresponds to the model $\lambda_k \bD_k \bA \bD_k^T$, which has the highest log marginal likelihood value. One can also see that the other models provide partitions with two, three or four clusters and thus do not overestimate the number of clusters. However, the solution selected by the PGMM approach corresponds to a partition with four clusters, and some of the PGMM models overestimate the number of clusters. 
\begin{table}[!ht]
  \centering
{\scriptsize
\hspace*{-1.3cm}
         \begin{tabular}{|c|c|c|c|c|c|c|c|c|c|c|} 
   \hline &\multicolumn{2}{c|}{DPPM} &\multicolumn{8}{c|}{PGMM}\\ 
  \hline
  $\text{Model}$ &$\hat{K}$ & $\log\text{ML}$ & $K=1$ & $K=2$ & $K=3$ & $K=4$ & $K=5$ & $K=6$ & $K=7$ & $K=8$ \\
  \hline
  \hline 
$\lambda \Identity$		&4&-415.68&-1124.9& -770.8& -455.6&-477.67&-431.22&-439.35&\bf{-423.49}&-457.59\\
 $\lambda_k \Identity$		&3&-471.99&-913.47&-552.2&\bf{-468.21}&-488.01&-507.8&-528.8&-549.62&-573.14\\
 $\lambda \bA$			&3&-404.87&-761.44&-585.53&-561.65&-553.41&-546.97&-539.91&-535.37&\bf{-530.96}\\
 $\lambda_k \bA$		&3&-432.62&-765.19&\bf{-623.89}&-643.07&-666.76&-688.16&-709.1&-736.19&-762.75\\
 $\lambda \bD \bA \bD^T$	&4&-307.31&-398.85&-340.89&-307.77&\bf{-286.96}& -291.7&-296.56&-300.37&-299.69\\
 $\lambda_k \bD \bA \bD^T$	&2&-383.72& -401.61&-330.55&-297.50&\underline{\bf{-279.15}}&-282.83&-296.24&-304.37&-306.81\\
 $\lambda \bD_k \bA \bD_k^T$	&4&-576.15&-1068.2&-761.71&-589.91&-529.52& -489.9&-465.37&\bf{-444.84}&-457.86\\
 $\lambda_k \bD_k \bA \bD_k^T$	&2&\cellcolor{gray!25}\bf{-278.78}&-394.68&\bf{-282.86}&-451.77&-676.18&-829.07&-992.04&-1227.2&-1372.8\\
  \hline
  \end{tabular}
  }
  \caption{\label{table:marginal likelihood iris data}{\footnotesize Log marginal likelihood values
  for the Iris data set.}}
\end{table}

  %
\begin{figure*}[!ht]
\centering
{\footnotesize 
\hspace*{-0.5cm}
\begin{tabular}{ccc} 
   \includegraphics[scale=.25]{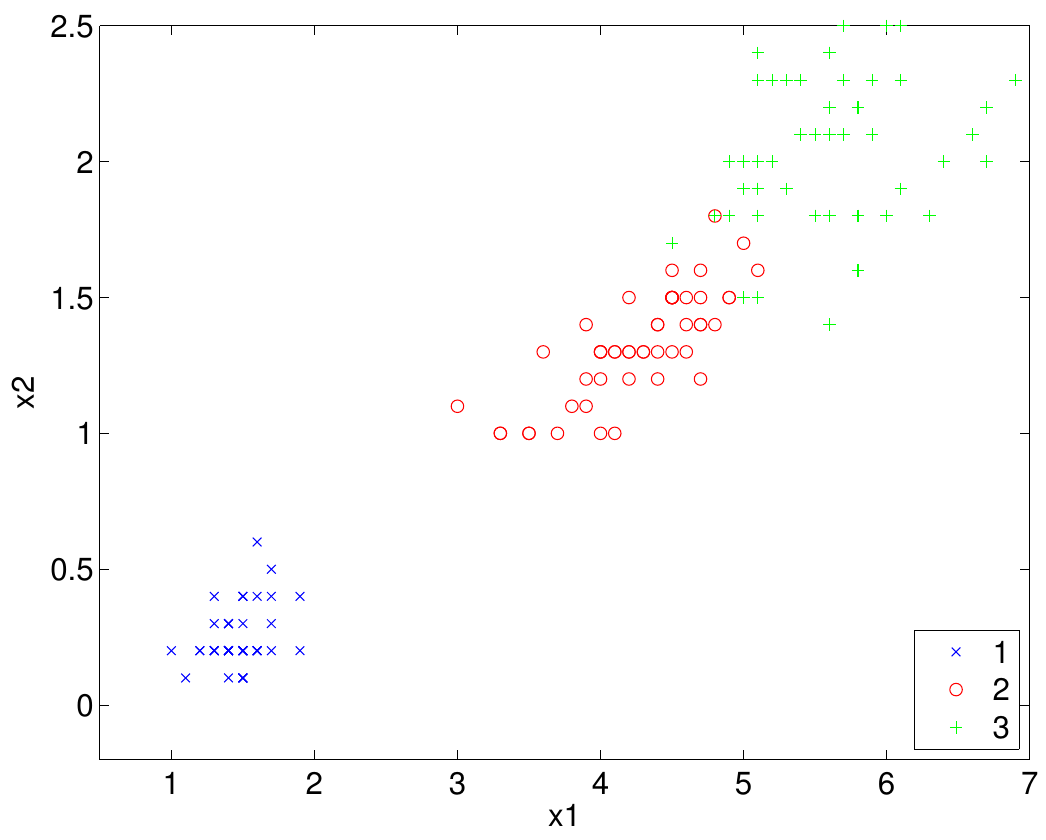} &
   \includegraphics[scale=.25]{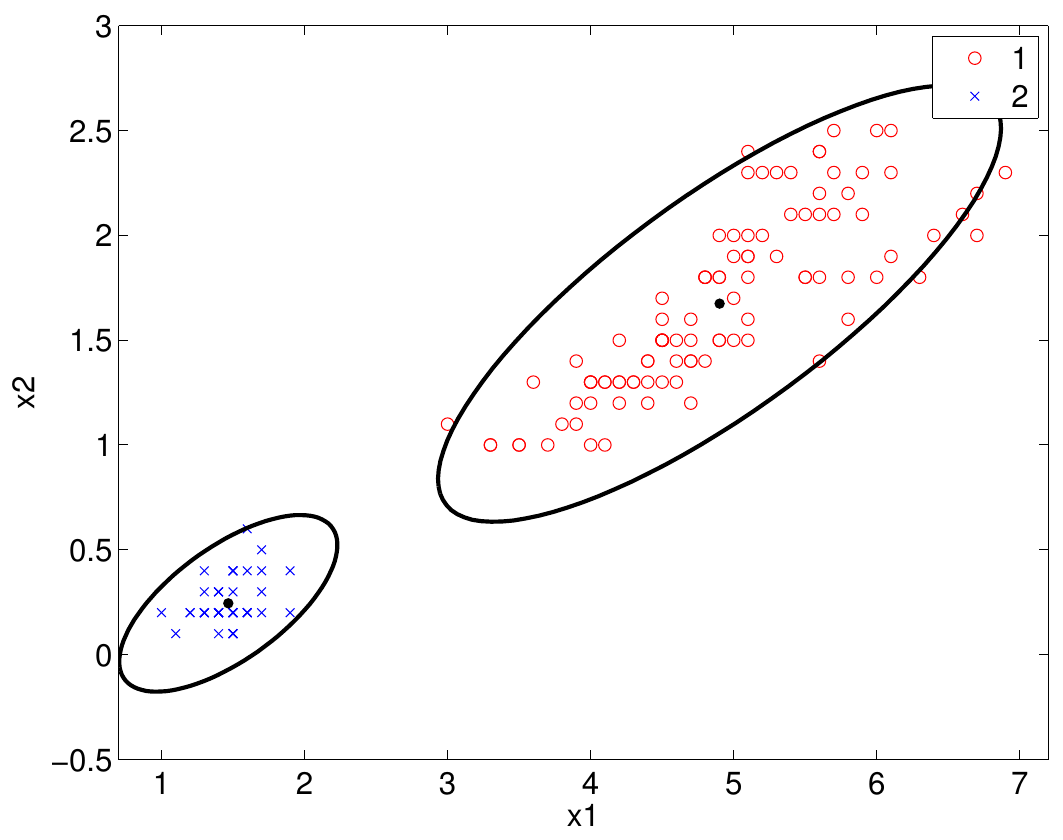} &
   \includegraphics[scale=.25]{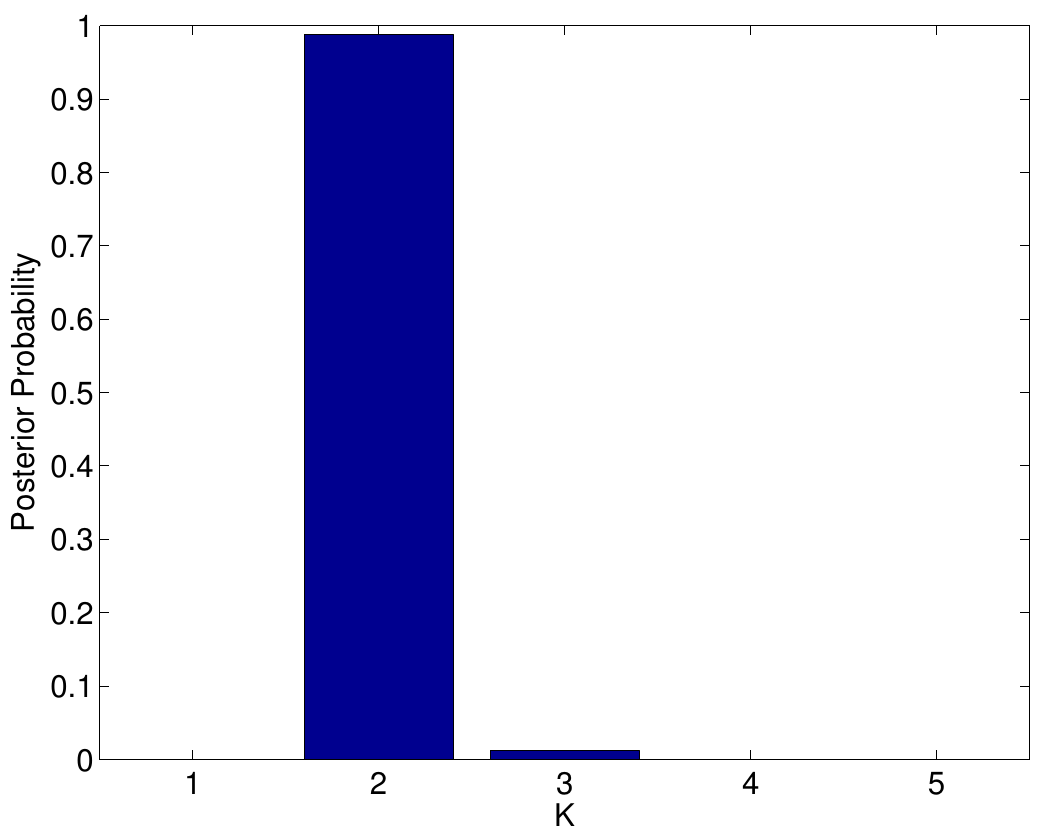} 
 \end{tabular} 
\caption{\label{fig: best DPPM partition for Iris}{Iris data set in the space of the components 3 (petal length) and 4 (petal width) (left), the optimal partition obtained by the DPPM  model $\lambda_k \bD_k \bA \bD_k^T$ (middle) and the empirical posterior distribution for the number of mixture components (right).}}
  }
 \end{figure*} 
We also note that, the best partition found by the proposed DPPM, 
while it contains two clusters, is quite well defined, and 
has a Rand index of $0.7763$.  

Table \ref{tab: cpu for Iris} shows the mean computer running time, measured in seconds, for the Gibbs inference of each DPPM models.
  \begin{table}[!ht]
   \centering
   \hspace*{-0.5cm}
   {\scriptsize
   \begin{tabular}{|c|c|c|c|c|c|c|c|c|}
    \hline
   $\text{Model}$&$\lambda \Identity$&$\lambda_k \Identity$&$\lambda \bA$ &$\lambda_k \bA$&$\lambda \bD \bA \bD^T$&$\lambda_k \bD \bA \bD^T$&$\lambda \bD_k \bA \bD_k^T$&$\lambda_k \bD_k \bA \bD_k^T$\\
   \hline
   \hline
   \text{CPU time (s)}& 144.04& 261.34 & 342.48 & 352.81& 293.91 & 382.0401 & 342.85 &196.66\\
   \hline
   
   \end{tabular}
   \caption{\label{tab: cpu for Iris}{The DPPM Gibbs sampler mean CPU time (in seconds) for each parsimonious model on Iris data set.}}
 }
  \end{table}
  

The evidence of the selected DPPM models, compared to the other ones, for the four real data sets, is significant. This can be easily seen in the tables showing the log marginal likelihood values. Consider the comparison between the selected model, and the more competitive for it, for the four real data. As it can be seen in Table \ref{table:bf values for real data}, which reports the values of $2\log\text{BF}$ of the best model against the second best one, that the evidence of the selected model, according to Table \ref{tab:bf_interpretation} is strong for Old Faithful geyser data, and very decisive for Crabs, Diabetes and Iris data. Also, the model selection by the proposed DPPM for these latter three data sets, is made with a greater evidence, compared to the PGMM approach. 
\begin{table}[!ht]
{\tiny
\centering
\hspace*{-1cm}
\begin{tabular}{|c|c|c|c|c|}
\hline 
Data set & Old Faithful Geyser & Crabs & Diabetes & Iris \\
\hline
\hline
DPPM & $\lambda \bD \bA \bD^T$ vs $\lambda_k \bD_k \bA \bD_k^T$ & $\lambda_k \bD_k \bA \bD_k^T$ vs $\lambda_k \bD \bA \bD^T $  & $\lambda_k \bD_k \bA \bD_k^T$ vs $\lambda \bD_k \bA \bD_k^T$  & $\lambda_k \bD_k \bA \bD_k^T$ vs $\lambda \bD \bA \bD^T$\\
\hline 
$2\log\text{BF}$ & 5  & 36.08  & 199.58 & 57.06\\ 
\hline
\hline
PGMM & $\lambda_k \bD \bA \bD^T$ vs $\lambda \bD \bA \bD^T$ & $\lambda_k \bD_k \bA \bD_k^T$ vs $\lambda \bD \bA \bD^T $  & $\lambda_k \bD_k \bA \bD_k^T$ vs $\lambda_k \bD \bA \bD^T$ & $\lambda_k \bD \bA \bD^T$ vs $\lambda_k \bD_k \bA \bD_k^T$\\
\hline
$2\log\text{BF}$ & 14.96  & 25.08  & 153.22 & 7.42\\ 
\hline
\end{tabular}
\caption{\label{table:bf values for real data}Bayes factor values for the selected model against the more competitive for it, obtained by the PGMM and the proposed DPPM for the real data sets.}}
\end{table} 


\subsection{Scaled application on real-world bioacoustic data}
\label{sec:experiments dppm whale song data}
In this section, we will apply the DPPM models on a further real dataset in the framework of a challenging problem of humpback whale song decomposition. 
The objective is the unsupervised decomposition of these bioacoustic data. 
Humpback whale songs are long cyclical sequences produced by males during the reproduction season which follows their migration from high-latitude to low-latitude waters. 
Singers of one geographical population share parts of the same song. 
This leads to the idea of dialect \citep{Helweg1998}. 
Different hypotheses of these songs were emitted \citep{Medrano94, Frankel95, Scott84, garland2011dynamic, Mercado1998bioacoustic}, even as used as sonar \citep{FrazerSonarWhale2000, AuSonarWhale2001}.

\subsection*{Data description}
The data consist in whale song signals in the 
framework of unsupervised analysis of bioacoustic data. %
%
%
%
This humpback whale song recording has been produced at few meters distance from the whale in La Reunion - Indian Ocean, by the "Darewin"
regroup in 2013, at a Frequency Sample of 44.1kHz, 32 bits, mono, wav format.

They consist of MFFC features of 8.6 minutes that have been extracted using Spro 5.0, with
pre-emphasis: 0.95, hamming window, fft on 1024 points (nearly 23ms),
frameshift 10 ms, 24 Mel channels, 12 MFCC coefficients and energy and
their delta and acceleration, CMS (mean normalisation) and variance
normalization, for a total of 39 dimensions as detailed in the SABIOD
NIPS challenge : http://sabiod.univ-tln.fr/nips4b/challenge2.html
where the signal and the features are available.

\begin{figure*}[!ht]
\centering
{\footnotesize 
\begin{tabular}{cc}
\includegraphics[width=6.5cm]{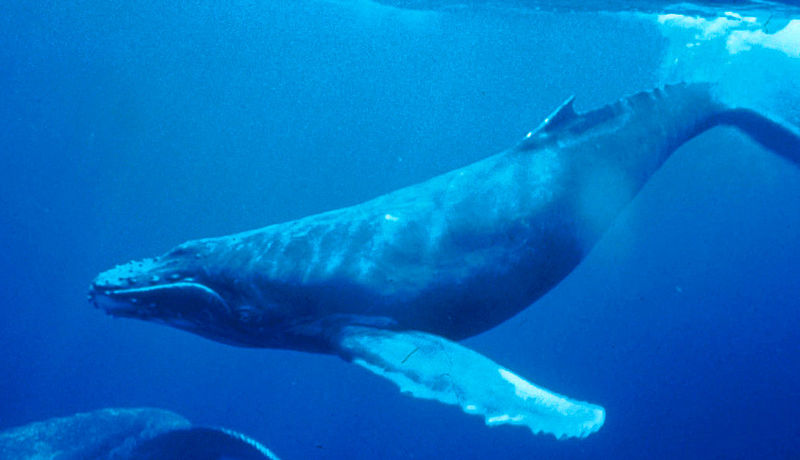}&
\includegraphics[width=7.5cm]{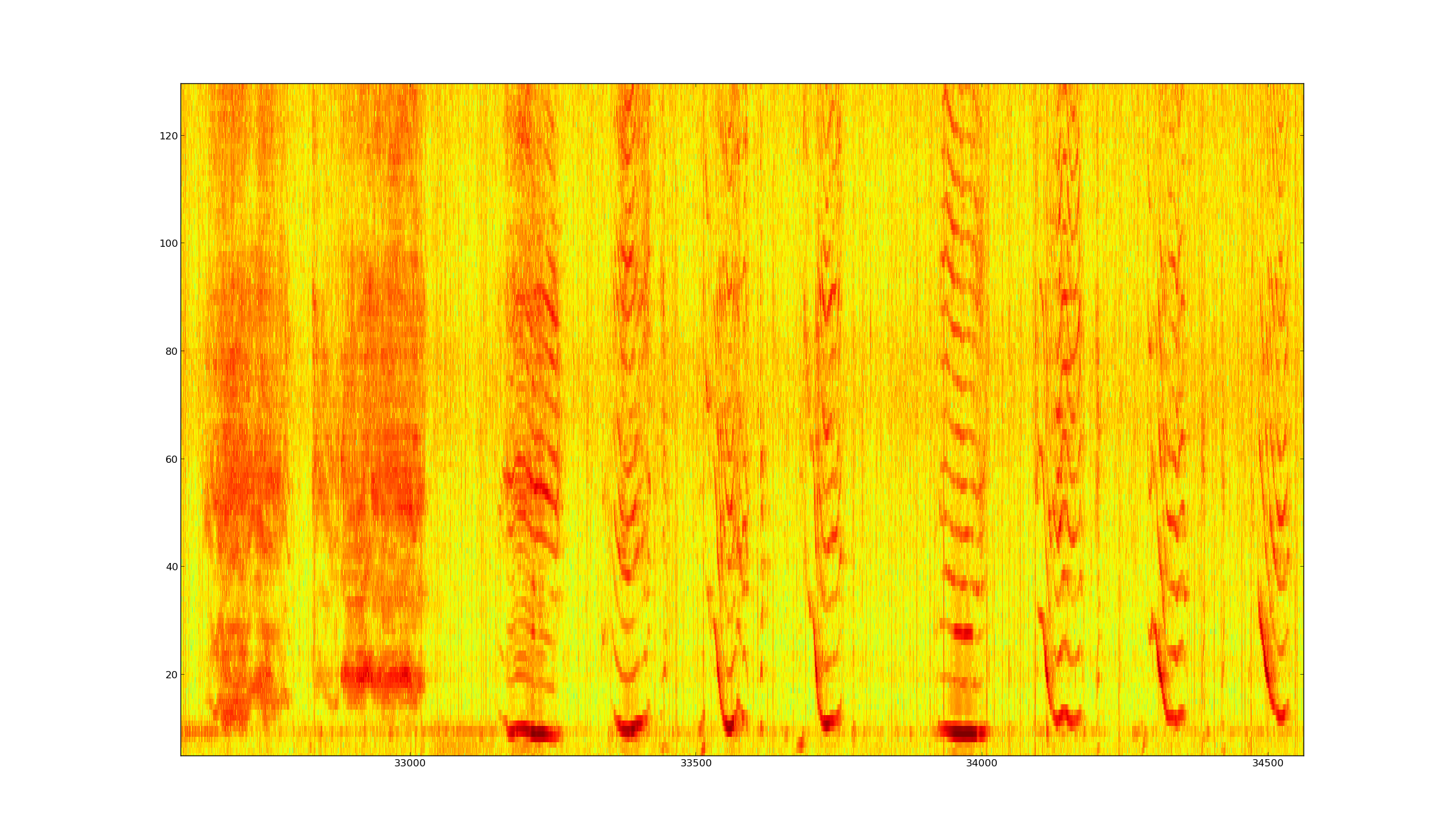}
\end{tabular}
  \caption{\label{fig: whale song representation}{On left, the Humpback whale and on right, the spectrum of around $20$ seconds of the given song of Humpback Whale.}}
}
\end{figure*}

A spectrum of this whale of around $20$ seconds of the given song can be seen in Figure \ref{fig: whale song representation}.
The data comprises 51336 observations with 39 features. 
A dimension reduction pretreatment with a PCA technique was made. 
%
We therefore choose to retain 13 features of the data, since it was sufficient to capture more then 95\% of the cumulative percentage of the variance. 

The analysis of such complex signals that aims at discovering the call units (which can be considered as a kind of whale alphabet), can be seen as a problem of unsupervised call units classification as in \cite{PaceEtAl2010}. 
Another analysis of the humpback whale song by clustering approach can be found in \cite{ClusterHumbackWhales2008}. 
The authors in \cite{ClusterHumbackWhales2008} implemented a segmentation algorithm based on Payne's principle to extract sound units of a whale song. 
In their application, six song units (pattern intonations) were found. 
We therefore reformulate the problem of whale song decomposition as an unsupervised data classification problem. 
Contrary to the approach used in  \cite{PaceEtAl2010}, in which the number of states (call units in this case) has been fixed manually, 
or \cite{ClusterHumbackWhales2008} where the unsupervised algorithm K-means was performed for automatic classification and then automatically define the optimal number of classes by 
maximizing the Davies Bouldin criterion. 
Here, we first apply the proposed DPPM models to learn the complex bioacoustic data, to find the classes (states) of the whale song, and  automatically infer the number of classes (states) from the data. 

\subsubsection*{Unsupervised decomposition of whale songs with the proposed DPPM models}
We applied our proposed DPPM approach, 
into the challenging problem of Whale song decomposition
NIPS4B challenge \citep{BartcusEtAl_NIPS2013}. 

The Gibbs sampling runs 10 times with $4000$ samplers and a burn-in period equal to $10\%$, by selecting the one with the highest MAP. 
Covering the three families, from the simplest one, which are the spherical models ($\lambda \bI$ and $\lambda_k \bI$), 
the diagonal models ($\lambda \bA$ and $\lambda_k \bA$), to the more complex general models ($\lambda \bD \bA \bD^T$, $\lambda_k \bD \bA \bD^T$ and $\lambda_k \bD_k \bA_k \bD_k^T$) 
are applied in this application. 

\begin{figure*}[!h]
\centering
{\small
\begin{tabular}{ccc}
\includegraphics[width=4cm]{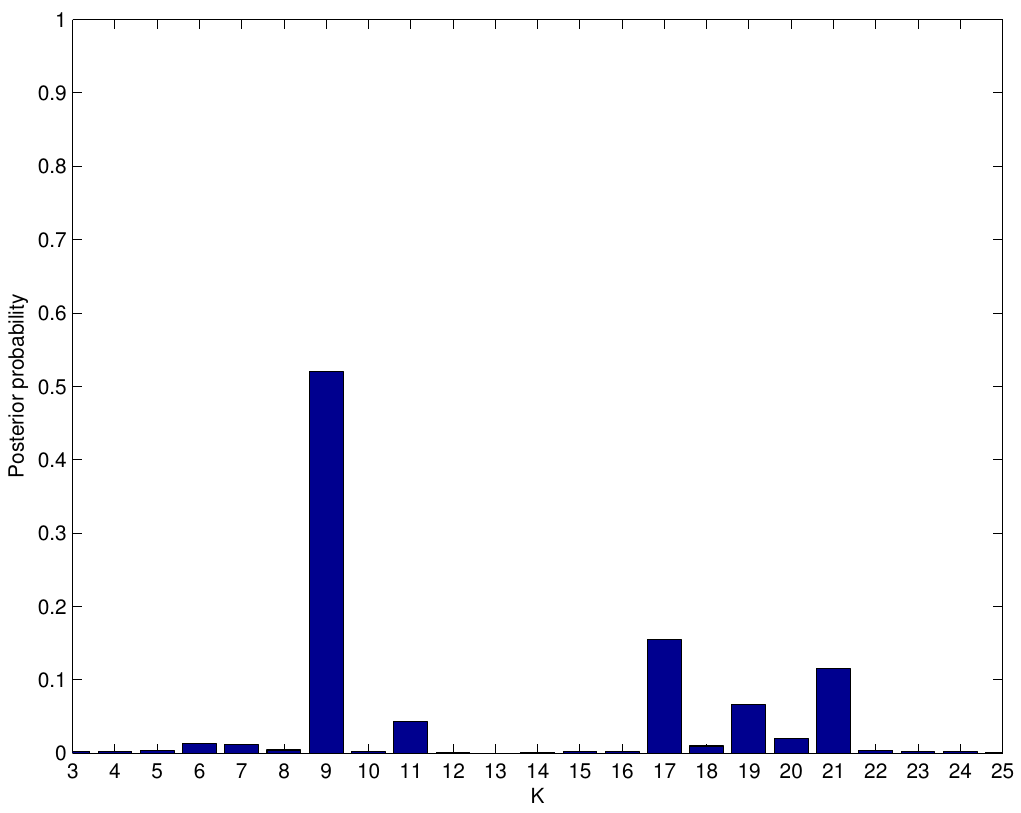}&

\includegraphics[width=4cm]{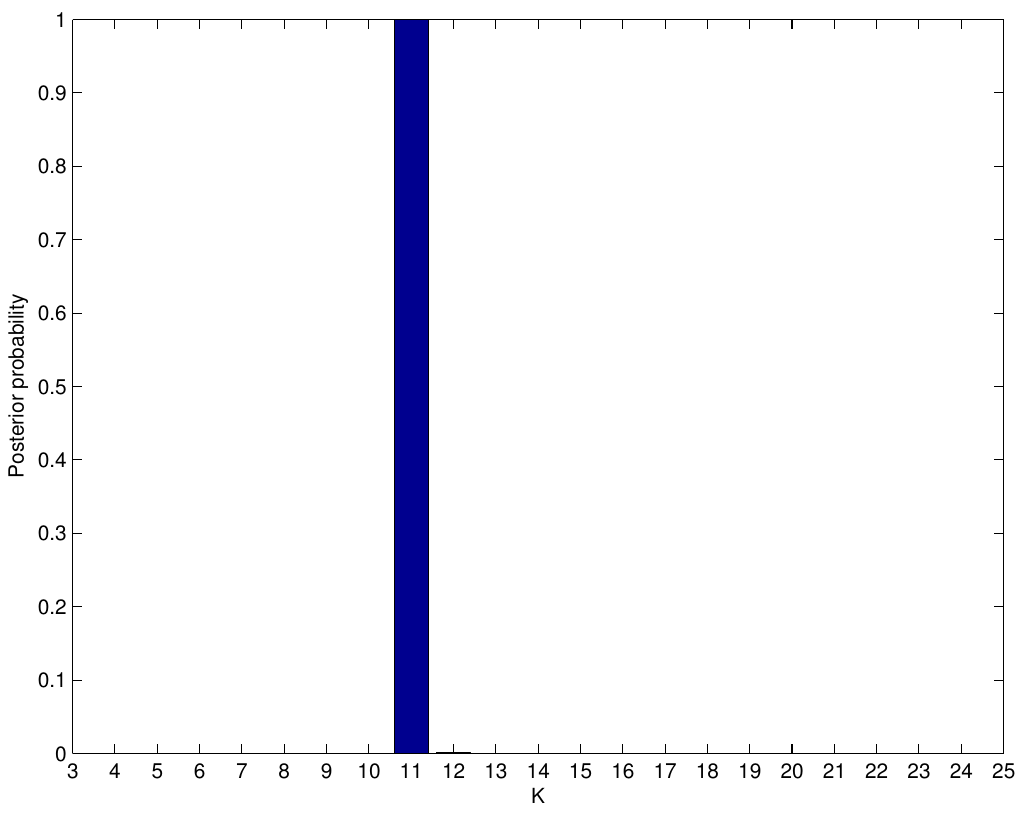}& 

\includegraphics[width=4cm]{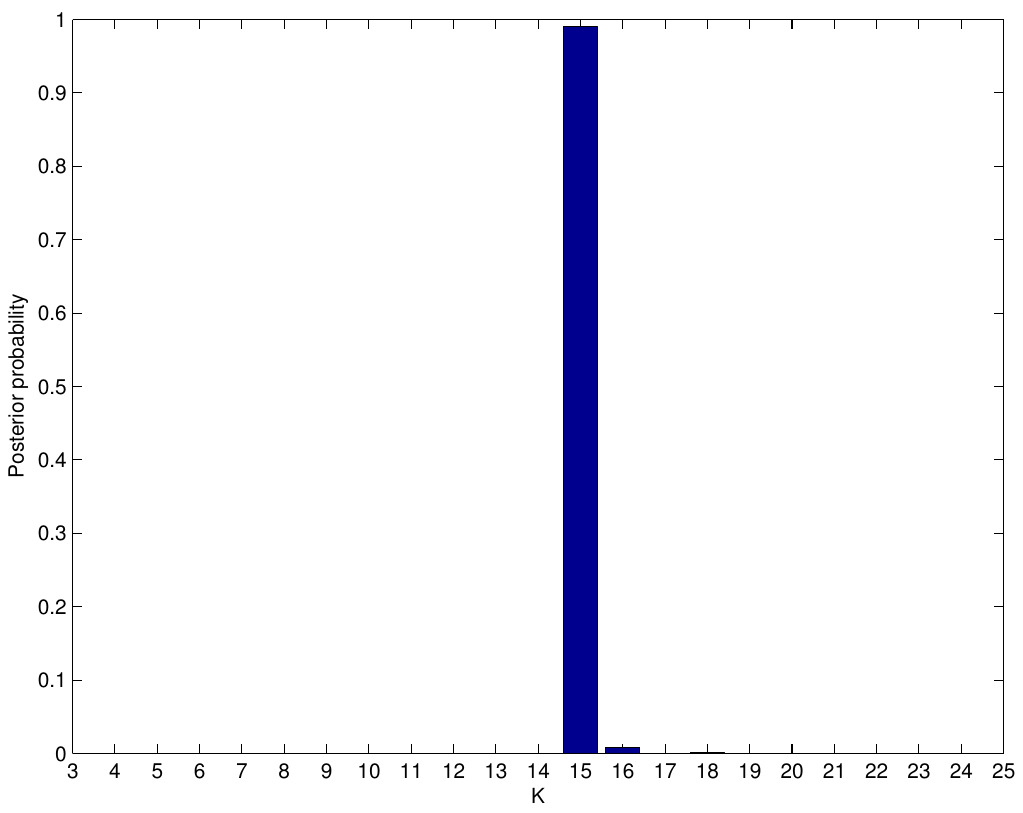}\\
$\lambda \bI$ & $\lambda_k \bA$ & $\lambda_k \bD_k \bA_k \bD_k^T$
\end{tabular}}
\caption{\label{fig:hist_K_record illustration} Posterior distribution of the number of components obtained by the proposed DPPM approach, for the whale song data.}
 \end{figure*}

In Figure \ref{fig:hist_K_record illustration} we show the posterior distributions of the numbers of components provided by the Gibbs sampler for the spherical model $\lambda \bI$, 
the diagonal model $\lambda_k \bA$ and the general model $\lambda_k \bD_k \bA_k \bD_k^T$. We can see that model $\lambda \bI$ retrieves $9$ clusters, 
the model $\lambda_k \bA$ retrieves $11$ clusters and model $\lambda_k \bD_k \bA_k \bD_k^T$ retrieves $15$ clusters. 

\begin{figure}[!h]
   \centering
 \includegraphics[width=7.1cm]{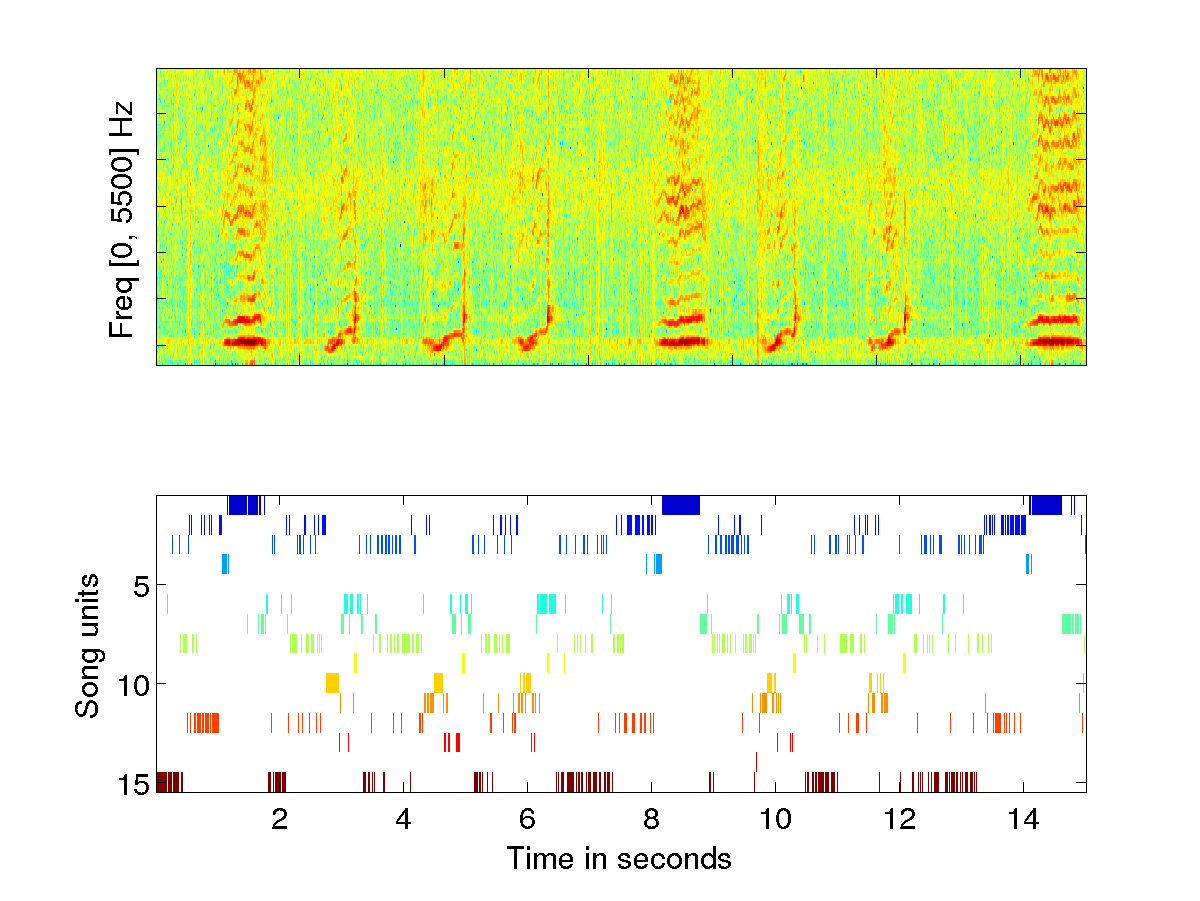}
\hspace*{-.9cm}
 \includegraphics[width=7.1cm]{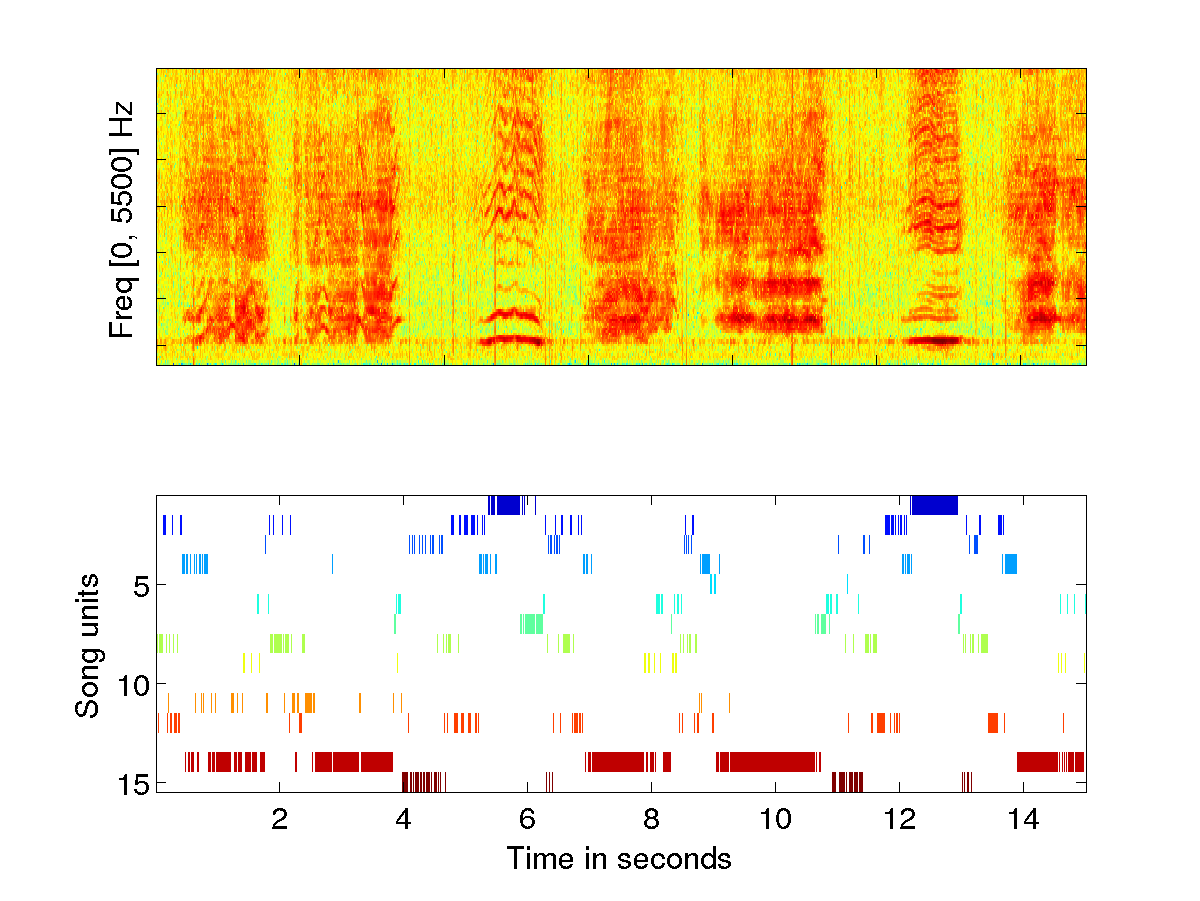}
 %
\caption{\label{fig: dppm full model whale song decomposition 3} Obtained song units by applying or DPM model with the parametrization $\lambda_k \bD_k \bA_k \bD_k^T$ (general) to two different signals with top: the spectrogram of the part of the signal starting at 280 seconds and it's corresponding partition, and bottom those for the part of signal starting at 295 seconds.} 
\end{figure}

Because of the length of 8.6 minutes of the signal,  
for a more detailed information, we show separate parts of $15$ seconds of the whole signal of the humpback whale. 
Some examples of the humpback whale song with $15$ seconds duration each are presented. 
Figure \ref{fig: dppm full model whale song decomposition 3} we show the two different signals with top, the signal starting at 280 seconds and it's corresponding partition obtained by the proposed DPPM model $\lambda_k \bD_k \bA_k \bD_k^T$ (general), 
and bottom those for the part of the signal starting at 295 seconds.%

Next, we illustrate the obtained results for the two proposed DPPM models, 
that corresponds to the parsimonious spherical model $\lambda \bI$ with equal cluster volumes and 
the parsimonious diagonal model $\lambda_k \bA$ with different cluster volumes. 
As for the general model $\lambda_k \bD_k \bA_k \bD_k^T$, 
we show separate parts of $15$ seconds duration of the whole signal of the humpback whale song in order to visualize the signal in a more detail.  

\begin{figure}[!h]
   \centering
 \includegraphics[width=7.1cm]{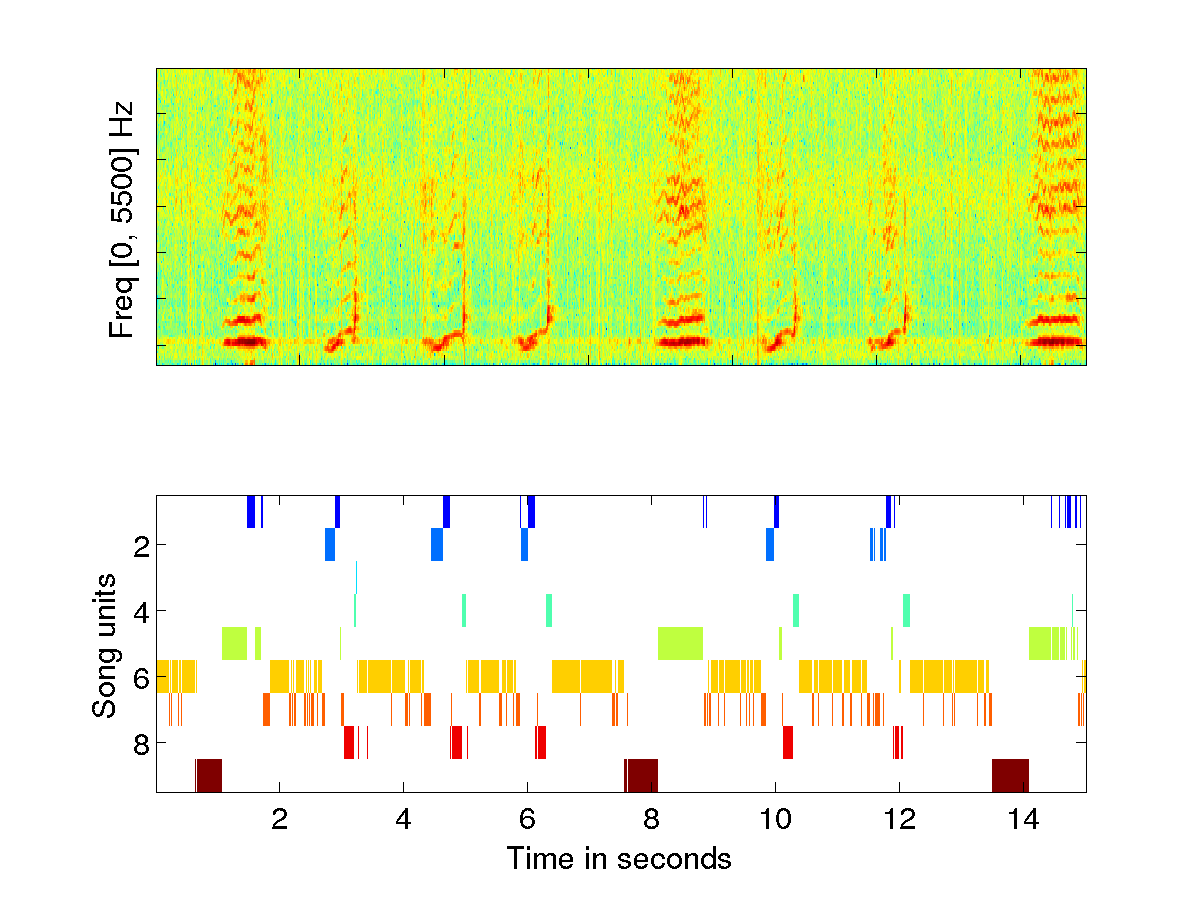}
\hspace*{-.9cm}
 \includegraphics[width=7.1cm]{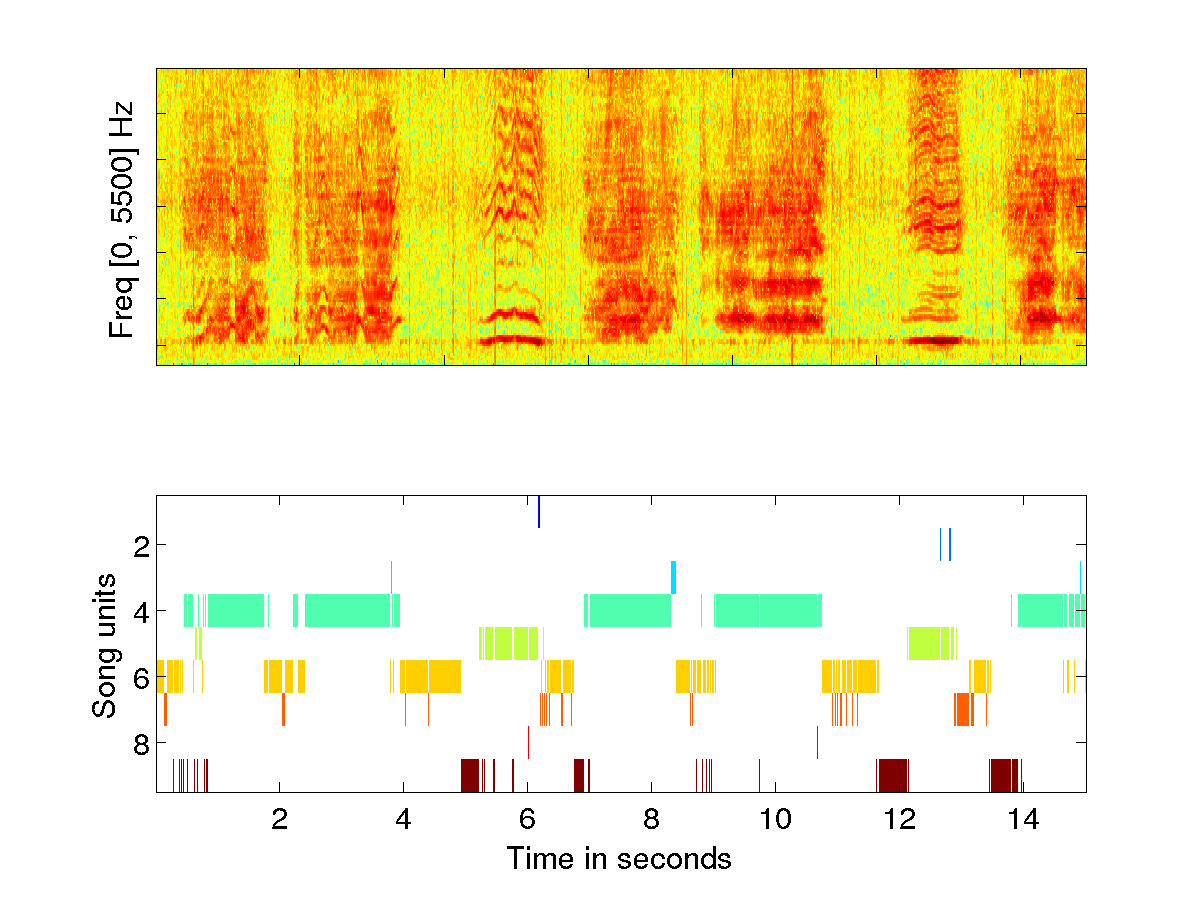}
\caption{\label{fig: dppm lI model whale song decomposition 3} Obtained song units by applying or DPPM model with the parametrization $\lambda \bI$ (spherical) to two different signals with top: the spectrogram of the part of the signal starting at 280 seconds and it's corresponding partition, and bottom those for the part of signal starting at 295 seconds.} 
\end{figure}

Finally, Figure \ref{fig: dppm lI model whale song decomposition 3} shows two different signals with top, the signal starting at 280 seconds and it's 
corresponding partition obtained by the proposed DPPM model $\lambda \bI$ (spherical), 
and bottom those for the part of the signal starting at 295 seconds.

The spherical $\lambda \bI$ model fit well the whale song data set with $9$ song units. 
In this situation, 
it is noticed that the sixth state represents the silence, that can be filled with state $7$ and $8$. 
The state $4$ is a very noisy and broad sound. 

\begin{figure}[!h]
   \centering
 \includegraphics[width=7.1cm]{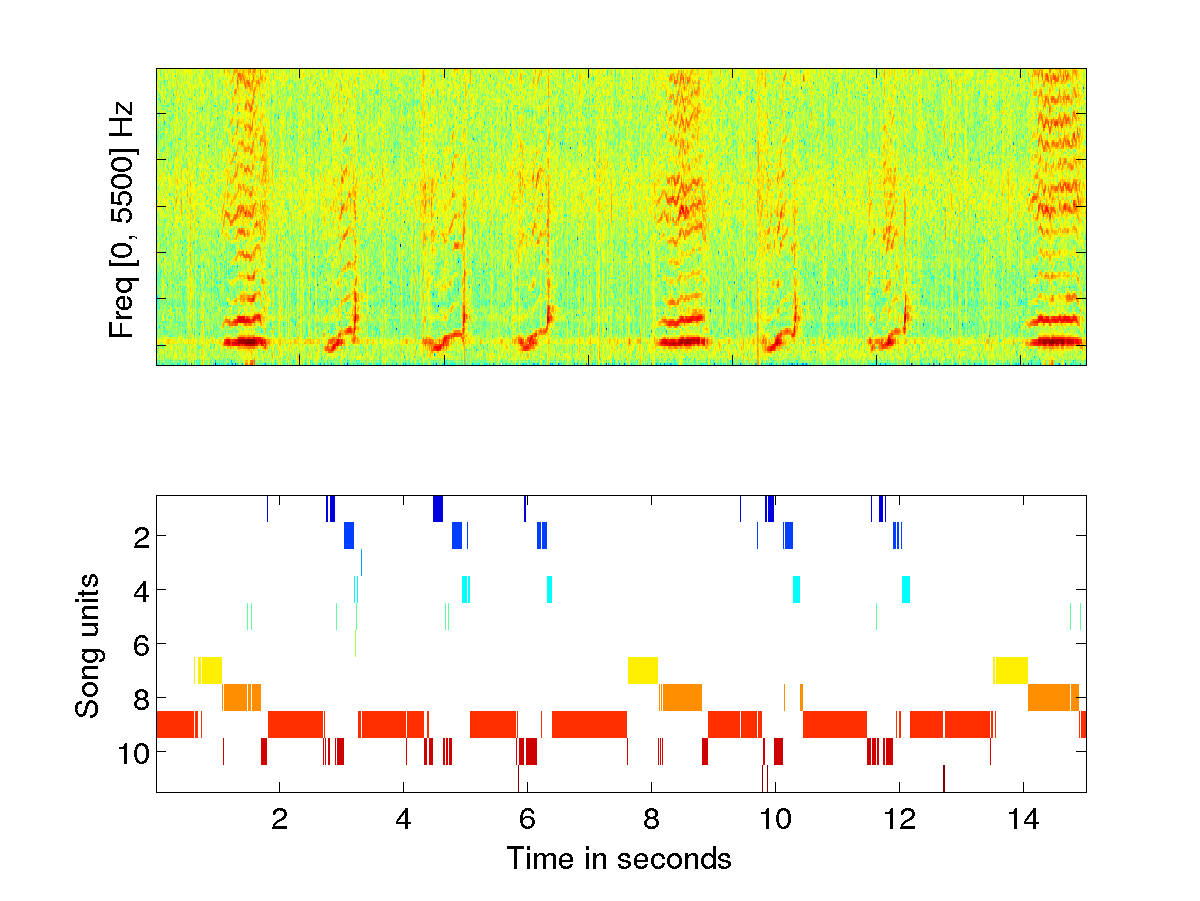}
\hspace*{-.9cm}
\includegraphics[width=7.1cm]{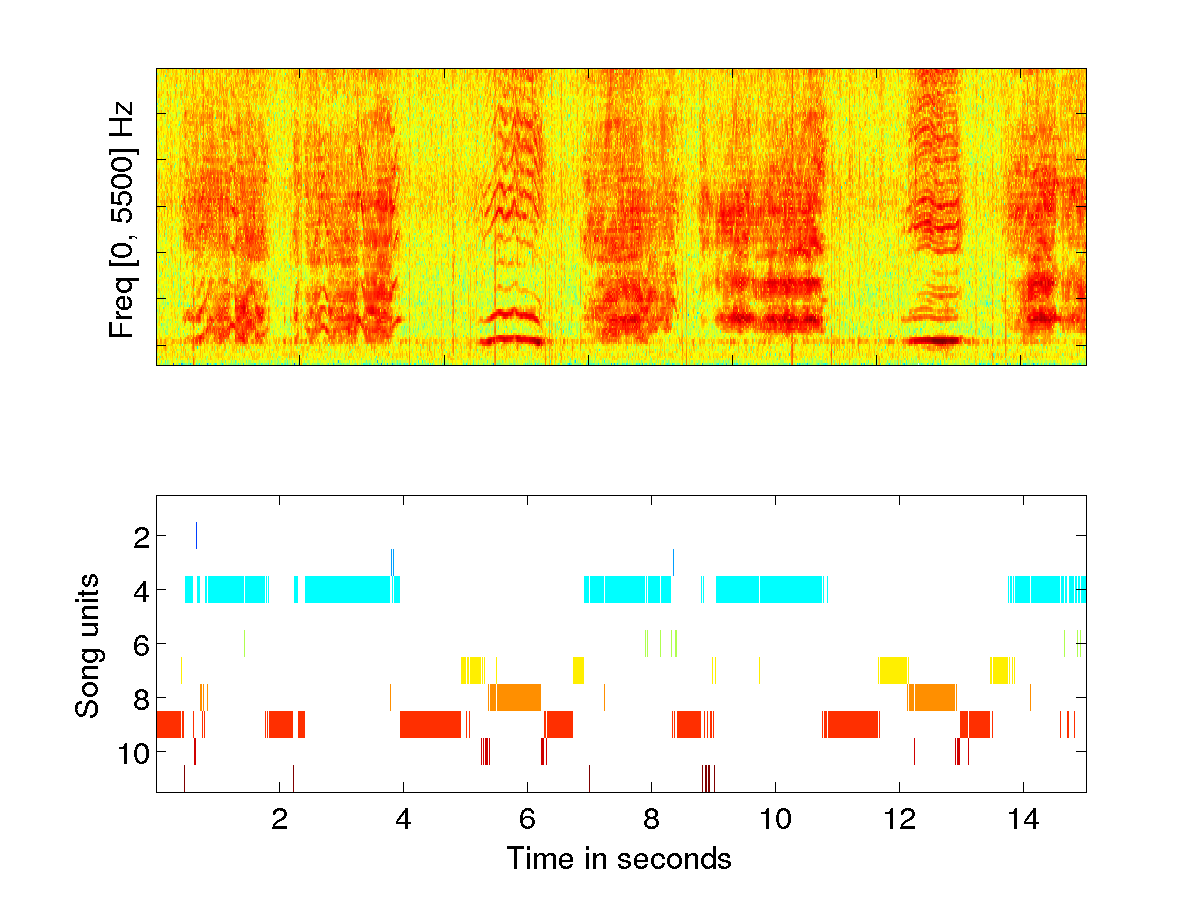}
\caption{\label{fig: dppm lkA model whale song decomposition 3} Obtained song units by applying or DPPM model with the parametrization $\lambda_k \bA$ (diagonal) to two different signals with top: the spectrogram of the part of the signal starting at 280 seconds and it's corresponding partition, and bottom those for the part of signal starting at 295 seconds.} 
\end{figure}

Figure \ref{fig: dppm lkA model whale song decomposition 3}, shows the signal starting with 280 seconds and it's corresponding obtained partition (top), and those for the part of the signal starting with 295 seconds (bottom). 

The DPPM diagonal model, with different cluster volumes, that corresponds to the covariance matrix decomposition $\lambda_k \bA$ 
fit well the data with $11$ song units. 
It can clearly be seen that the state $9$ is the silence. State $1$, $2$, $8$ and $11$ is the up and down sweeps. 
The seventh state is also the silence that generally ends the ninth state. The state $4$ is a very noisy and broad sound. 
These obtained results highlight the potential parsimonious Bayesian non-parametric modelling for the unsupervised segmentation of the studied bioacoustic data. 

\section{Conclusion}
 \label{sec: conclusion}
In this paper we presented Bayesian nonparametric parsimonious mixture models for clustering. It is based on an infinite Gaussian mixture with an eigenvalue decomposition of the cluster covariance matrix and a Dirichlet Process, or by equivalence a Chinese Restaurant Process prior. 
This allows deriving several flexible models and avoids the problem of model selection encountered in the standard maximum likelihood-based and Bayesian parametric Gaussian mixture.
We also proposed a Bayesian model selection an comparison framework to automatically select, the best model, with the best number of components, by using Bayes factors. 

Experiments on simulated data highlighted that the proposed DPPM represent a good nonparametric alternative to the standard parametric Bayesian and non-Bayesian finite mixtures. They simultaneously and accurately estimate accurate partitions with the optimal number of clusters also inferred from the data. 
We also applied the proposed approach on real data sets. The obtained results show the interest of using the  Bayesian parsimonious clustering models and the potential benefit of using them in practical applications. We applied the models on the challenging problem of humpback whale song decomposition. 
Despite the fact that the dataset are by nature sequential, and DPPMs models assume an exchangeability property, 
the models arrive to fit quiet satisfying partition of the data. This application opens a perspective on the extension of the previously discussed DPPMs models, 
from the i.i.d case to sequential data. Hence this may provide a good perspective 
for further integrating the parsimonious DPM models into a Markovian framework. 

A future work  related to this proposal may concern other parsimonious models such us those proposed by \cite{BiernackiL14}  based on a variance-correlation decomposition of the group covariance matrices, which  are stable and visualizable and have  desirable properties. 

Until now we have only considered the problem of clustering. A perspective of this work is to extend it to the case of model-based co-clustering \citep{Govaert_Nadif-Co-ClusteingBook} with block mixture models, which consists in simultaneously cluster individuals and variables, rather that only individuals. 
The nonparametric formulation of these models may represent a good alternative to select the number of latent blocks or co-clusters.
 
\appendix
\section{Prior and Posterior distributions for the model parameters}
 \label{appendix} 
 Here we provide the prior and  posterior distributions (used in the Gibbs sampler) for the mixture model parameters for each of the developed DPPM models.   
%
First, recall that $\bz=(z_1,\ldots,z_n)$ denotes a vector of class labels  
where $z_i$ is the class label of $\bx_i$. Let $z_{ik}$ be the indicator binary variable such that $z_{ik}=1$ if $z_{i}=k$ (i.e when $\bx_i$ belongs to component $k$). Then, let $n_k = \sum_{i=1}^n z_{ik}$ represents the number of data points belonging to cluster (or component) $k$. Finally, let 
$\bar{\bx}_k = \frac{\sum_{i=1}^n z_{ik} \bx_i}{n_k}$ be the empirical mean vector of cluster $k$, and 
  $W_k = \sum_{i=1}^n z_{ik} (\bx_i-\bar{\bx}_k) (\bx_i-\bar{\bx}_k)^T$ its scatter matrix. 
\subsection{Hyperparameters values}
In our experiments for the multivariate parsimonious models, we choose the prior hyperparameters $H$ as follows: 
the mean of the data $\bsmu_0$, 
the shrinkage $\kappa_n = 0.1$, 
the degrees of freedom $\nu_0 = d+2$, 
the scale matrix $\Lambda_0$ equal to the covariance of the data, 
and for the spherical models, the hyperparameter  $s_0^2$ was taken as the greatest eigenvalue of $\Lambda_0$.



\subsection{Spherical models}
\paragraph{(1)  Model $\lambda \bI$} For this spherical model, the covariance matrix, for all the mixture components, is parametrized as $\lambda \bI$ and hence is described by the scale parameter $\lambda>0$, which is common for all the mixture components. For this spherical model, the prior over the covariance matrix is defined through the prior over $\lambda$, for which we used a conjugate prior density, that is an inverse Gamma. For the mean vector for each of Gaussian components, we used a conjugate multivariate normal prior. The resulting prior density   is therefore a normal inverse Gamma conjugate prior: 
\begin{eqnarray}
  \bsmu_k |\lambda & \sim & \cN(\bsmu_0, \lambda \bI / \kappa_n) \ \forall k=1,\ldots,K\\
  \lambda  & \sim& \cI\cG (\nu_0/2, s_0^2/2) \nonumber
\end{eqnarray}where 
$(\bsmu_0, \kappa_n)$ are the hyperparamerets for the multivariate normal over $\bsmu_k$
and
$(\nu_0, s_0^2)$ are those for the inverse Gamma over $\lambda$.
Therefore, the resulting posterior is a multivariate Normal inverse Gamma and the sampling from this posterior density is performed as follows:
{\footnotesize
\begin{eqnarray}
\bsmu_k|\bX, \bz, \lambda, H  &\sim& \cN (\bsmu_n, \lambda \bI/(n_k+\kappa_n)) \nonumber \\
\lambda|\bX, \bz, H &\sim& \cI\cG (\frac{\nu_0+n_k}{2}, \frac{1}{2} \{ s_0^2 + \sum\limits_{k=1}^{K} \tr(W_k) + \sum\limits_{k=1}^{K} \frac{n_k \kappa_n}{n_k+\kappa_n} (\bar{\bx}_k - \bsmu_0)^T (\bar{\bx}_k - \bsmu_0) \} ) \nonumber
\end{eqnarray}
}where the posterior mean $\bsmu_n$ is equal to  $\frac{n_k\bar{\bx}_k + \kappa_n\bsmu_0}{n_k + \kappa_n}$.

\paragraph{(2) Model $\lambda_k \bI$} 
This other spherical model parametrized $\lambda_k \bI$ is also described by the scale parameter $\lambda_k>0$ which is different for all the mixture components.
As for the previous spherical model, a normal inverse Gamma conjugate prior is used. 
In this situation the scale parameter $\lambda_k$ will have different priors and respectively posterior distributions for each mixture component.
The resulting prior density for this spherical model is a normal inverse Gamma conjugate prior: 
{\small 
\begin{eqnarray}
  \bsmu_k|\lambda_k &\sim& \cN(\mu_0, \lambda_k \bI / \kappa_n) \ \forall k=1,\ldots,K \nonumber \\
  \lambda_k &\sim& \cI\cG (\nu_k/2, s_k^2/2) \ \forall k=1,\ldots,K \nonumber
\end{eqnarray}
}where 
$(\bsmu_0, \kappa_n)$ are the hyperparamerets for the multivariate normal over $\bsmu_k$
and
$(\nu_k, s_k^2)$ are those for the inverse Gamma over $\lambda_k$. 
The set of hyperparameters $\nu_k =\{ \nu_1,\ldots, \nu_k\}$ and $s_k=\{s_1\ldots s_k\}$ are chosen to be equal, throw all the components of the mixture, to $\nu_0$ and respectively $s_0^2$.
Analogously, the resulting posterior is a normal inverse Gamma and the sampling for the model parameters $(\bsmu_1,\ldots,\bsmu_K, \lambda_1,\ldots,\lambda_K)$ is performed as follows:
{\footnotesize
\begin{eqnarray}
\bsmu_k|\bX, \bz, \lambda_k, H &\sim& \cN (\bsmu_n, \lambda_k\bI/(n_k+\kappa_n)) \nonumber \\
\lambda_k|\bX, \bz, H &\sim& \cI\cG (\frac{\nu_k+d n_k}{2}, \frac{1}{2} \{ s_k^2 + \tr(W_k) + \frac{n_k \kappa_n}{n_k+\kappa_n} (\bar{\bx}_k - \bsmu_0)^T (\bar{\bx}_k - \bsmu_0) \} ). \nonumber
\end{eqnarray}
}

\subsection{Diagonal models}
\paragraph{(3)  Model $\lambda \bA$}
The diagonal parametrization $\lambda \bA$ of the covariance matrix is described by the volume $\lambda$ (a scalar term) and a diagonal matrix $\bA$. The parametrization $\lambda \bA$ therefore corresponds to a diagonal matrix whose diagonal terms are $a_j, \ \forall j=1,\ldots d$. 
The prior normal inverse Gamma conjugate prior density is given as follows:
\begin{eqnarray}
  \bsmu_k|\bsSigma_k &\sim& \cN(\bsmu_0, \bsSigma_k / \kappa_n) \ \forall k=1,\ldots,K \nonumber \\
  a_j &\sim& \cI\cG (r_j/2, p_j/2) \ \forall j=1\ldots d \nonumber
\end{eqnarray} where the set of parameters $r_j, p_j$ are considered to be equal $\forall j=1\ldots d$ to $\nu_0$ and respectively $s_k^2$.
The resulting posterior for the model parameters takes the following form:
{\footnotesize
\begin{eqnarray}
  \bsmu_k|\bX, \bz, \bsSigma_k, H  &\sim& \cN (\bsmu_n, \bsSigma_k/(n_k+\kappa_n)) \nonumber \\
  a_j|\bX, \bz, H &\sim& \cI\cG (   \frac{n_k+\nu_k + K(d+1)-2}{2},      \frac{\text{diag}(\sum_{k=1}^{K} \frac{n_k \kappa_n}{n_k+\kappa_n} (\bar{\bx}_k - \bsmu_0) (\bar{\bx}_k - \bsmu_0)^T + W_k + \Lambda_k )}{2}  ) \nonumber 
 \end{eqnarray}
 }where the posterior mean $\bsmu_n=\frac{n_k\bar{\bx}_k + \kappa_n\bsmu_0}{n_k + \kappa_n}$.

\paragraph{(4)  Model $\lambda_k \bA$}
This diagonal model, analogous to the previous one, but with different  volume $\lambda_k>0$ for each component of the mixture, takes the parametrization $\lambda_k \bA$.
In this situation, the normal prior density  for the mean remains the same and the inverse Gamma prior density for the volume parameter $\lambda_k$ is given as follows:
\begin{eqnarray}
  \lambda_k &\sim& \cI\cG (r_k/2, p_k/2) \ \forall j=1\ldots K \nonumber
\end{eqnarray} where the set of hyperparamerets for the scale parameter $\lambda_k$, $r_k=\{r_1,\ldots,r_K\}$ and $p_k=\{p_1,\ldots,p_k\}$ are considered to be equal, for all mixture components, to respectively $\nu_0$ and $s_k^2$.
The resulting posterior distributions over the parameters of the model are given as follows:
{\footnotesize
\begin{eqnarray}
 \bsmu_k|\bX, \bz, \bsSigma_k, H &\sim& \cN (\bsmu_n, \bsSigma_k/(n_k+\kappa_n)) \nonumber \\
 a_j|\bX, \bz, \lambda_k, H &\sim& \cI\cG (   \frac{n_k+\nu_k + Kd+1}{2},      \frac{\text{diag}(\sum_{k=1}^{K} \lambda_k^{-1}  (\frac{n_k \kappa_n}{n_k+\kappa_n} (\bar{\bx}_k - \bsmu_0) (\bar{\bx}_k - \bsmu_0)^T + W_k + \Lambda_k)  )}{2}  ) \nonumber \\
 \lambda_k|\bX, \bz, \bA, H &\sim& \cI\cG ( \frac{r_k+n_kd}{2}, \frac{p_k + \tr(\bA^{-1}  ( \frac{n_k \kappa_n}{n_k+\kappa_n} (\bar{\bx}_k - \bsmu_0) (\bar{\bx}_k - \bsmu_0)^T + W_k + \Lambda_k ) )}{2} ).\nonumber 
\end{eqnarray}
}
\vspace{-0.5cm}
\subsection{General models}
\paragraph{(5) Model $\lambda \bD \bA \bD^T$}
The first general model has the $\lambda \bD \bA \bD^T$ parametrization,
where the covariance matrices have the same volume $\lambda>0$, orientation $\bD$ and shape $\bA$ for all the components of the mixture.
This is equivalent, in the literature, to the model where the covariance $\bsSigma$ is considered equal throw all the components of the mixture.
The resulting conjugate normal inverse Wishart prior over the parameters $(\bsmu_1, \ldots, \bsmu_K, \bsSigma)$ is given as follows:
{\small \begin{eqnarray}
  \bsmu_k|\bsSigma &\sim& \cN(\bsmu_0, \bsSigma / \kappa_n) \ \forall k=1,\ldots,K \nonumber \\
  \bsSigma &\sim& \cI\cW (\nu_0, \Lambda_0) \nonumber
\end{eqnarray}}where $(\bsmu_0,\kappa_n)$ are the hyperparameters for the multivariate normal prior over $\bsmu_k$
and $(\nu_0, \Lambda_0)$ are hyperparameters for the inverse Wishart prior $(\cI\cW)$ over the covariance matrix $\bsSigma$ that is common to all the components of the mixture. 
The posterior of the model parameters $(\bsmu_1,\ldots,\bsmu_K, \bsSigma)$ for this general model is given by:
{\footnotesize
\begin{eqnarray}
  \bsmu_k|\bX, \bz, \lambda_k, H &\sim& \cN (\bsmu_n, \bsSigma/(n_k+\kappa_n)) \nonumber \\
  \bsSigma|\bX, \bz, H &\sim& \cI\cW (\nu_0+n_k, \Lambda_0 +\sum\limits_{k=1}^{K} \{ W_k +  \frac{n_k \kappa_n}{n_k+\kappa_n} (\bar{\bx}_k - \bsmu_0) (\bar{\bx}_k - \bsmu_0)^T \} ). \nonumber
\end{eqnarray}}

\paragraph{(6) Model $\lambda_k \bD \bA \bD^T$}
The second parsimonious model from the general family has the parametrization $\lambda_k \bD \bA \bD^T$, 
where the volume $\lambda_k$ of the covariance differs from one mixture component to another, but the orientation $\bD$ and the shape $\bA$ are the same for all the mixture components. This parametrization can thus be simplified as 
$\lambda_k \bsSigma_0$, where the parameter $\bsSigma_0 = \bD\bA\bD^T$. 
This general model has therefore a Normal prior distribution over the mean, an inverse Gamma prior distribution over the scale parameter $\lambda_k$ and 
an inverse Wishart prior distribution over the matrix $\bsSigma_0$ that controls the orientation and the shape for the mixture components.
The conjugate prior for the mixture parameters $(\bsmu_1,\ldots,\bsmu_K,\lambda_1,\ldots,\lambda_K, \bsSigma_0)$ are thus given as follows:
{\footnotesize
\begin{eqnarray}
  \bsmu_k|\lambda_k,\bsSigma_0  &\sim& \cN(\bsmu_0, \lambda_k \bsSigma_0/\kappa_n) \ \forall k=1,\ldots, K \nonumber \\
  \lambda_k &\sim& \cI\cG (r_k/2, p_k/2) \ \forall k=2,\ldots, K \nonumber \\
  \bsSigma_0 &\sim& \cI\cW (\nu_0,\Lambda_0) \nonumber
\end{eqnarray}}where $\lambda_1$ is supposed to be equal to $1$ (to make the model identifiable),  the hyperparameters $\{r_1,\ldots,r_K\}$ and $\{p_1\ldots p_K\}$ are supposed to be equal to respectively $\nu_0$ and  $s_k^2$ for each of the mixture components.
The resulting posterior over the parameters $(\bsmu_1,\ldots,\bsmu_K, \lambda_1,\ldots,\lambda_K, \bsSigma_0)$ of this model is given as follows:
{\footnotesize
\begin{eqnarray} 
\bsmu_k|\bX, \bz, \lambda_k, \bsSigma_0, H &\sim& \cN (\bsmu_n, \lambda_k \bsSigma_0/(n_k+\kappa_n)) \nonumber \\
\lambda_k|\bX, \bz, H &\sim& \cI\cG ( \frac{r_k+n_k d}{2}, \frac{1}{2} \{ p_k+\tr(W_k \bsSigma_0^{-1}) + \frac{n_k\kappa_n}{n_k+\kappa_n} (\bar{\bx}_k - \bsmu_0)^T \bsSigma_0^{-1} (\bar{\bx}_k - \bsmu_0) \} ) \nonumber \\ 
\bsSigma_0|\bX, \bz, H &\sim& \cI\cW (\nu_0+n_k, \Lambda_0 +\sum\limits_{k=1}^{K} \{ \frac{W_k}{\lambda_k} +  \frac{n_k \kappa_n}{\lambda_k(n_k+\kappa_n)} (\bar{\bx}_k - \bsmu_0)^T (\bar{\bx}_k - \bsmu_0) \}). \nonumber
\end{eqnarray}
}

\paragraph{(7) Model $\lambda \bD_k \bA \bD_k^T$}
This other general model $\lambda \bD_k \bA \bD_k^T$ is parametrized by  the scalar parameter (the volume) $\lambda$ and the shape diagonal matrix $\bA$. 
This model parametrization can therefore be summarized to the $\bD_k \bA \bD_k^T$ parametrization, by including $\lambda$ in a resulting diagonal matrix $\bA$, whose diagonal elements $a_1,\ldots, a_d$. 
The prior density over the mean is normal, the one over the orientation matrix $\bD_k$  is inverse Wishart, and the one over each of the diagonal elements $a_j, \ \forall j=1\ldots d$ of the matrix $\bA$ is an inverse Gamma. The conjugate prior for this general model is therefore as follows:
\begin{eqnarray}
  \bsmu_k|\bsSigma_k &\sim& \cN(\bsmu_0, \bsSigma_k / \kappa_n) \ \forall k=1,\ldots,K \nonumber \\
  a_j &\sim& \cI\cG (r_j/2, p_j/2) \ \forall j=1\ldots d \nonumber
\end{eqnarray}
The  hyperparameters $r_j$ and $p_j$ for the $\lambda \bA$, are considered to be the same $\forall j=1\ldots d$ and are respectively equal to $\nu_0$ and $s_k^2$.
 The resulting posterior for the model parameters takes the following form:
{\footnotesize
\begin{eqnarray}
  \bsmu_k|\bX, \bz, \bsSigma_k, H  &\sim& \cN (\bsmu_n, \bsSigma_k/(n_k+\kappa_n)) \nonumber \\
  a_j|\bX, \bz, H &\sim& \cI\cG (   \frac{n_k+\nu_k + K(d+1)-2}{2},      \frac{\text{diag}(\sum_{k=1}^{K} \bD_k^T (\frac{n_k \kappa_n}{n_k+\kappa_n} (\bar{\bx}_k - \bsmu_0) (\bar{\bx}_k - \bsmu_0)^T + W_k + \Lambda_k) \bD_k )}{2}). \nonumber 
 \end{eqnarray}}
 The parameters, that controls the orientation of the covariance, $D_k$, have the same inverse Wishart posterior distribution as the general covariance matrix:
{\footnotesize
\begin{equation}
  \bD_k|\bX, \bz, H \sim \cI\cW ( n_k+\nu_k, \Lambda_k + W_k +  \frac{n_k \kappa_n}{n_k+\kappa_n} (\bar{\bx}_k - \bsmu_0) (\bar{\bx}_k - \bsmu_0)^T ) \nonumber
 \end{equation}}And as mentioned above the covariance matrix $\bsSigma_k$ for this model will be formed as $\text{diag}(a_j)\bD_k$.
 
\paragraph{(8) Model $\lambda_k \bD_k \bA \bD_k^T$}
The third considered parsimonious model for the general family, is the one with the parametrization $\lambda_k \bD_k \bA \bD_k^T$ of the covariance matrix, 
and is analogous to the previous model, but for this one, the scale $\lambda_k$ of the covariance (the cluster volume) differs for each component of the mixture. 
The prior over each of the scale parameters $\lambda_1 \ldots \lambda_K$ is an inverse Gamma prior :
\begin{eqnarray}
  \lambda_k &\sim& \cI\cG (r_k/2, p_k/2) \ \forall k=1,\ldots, K. \nonumber  
\end{eqnarray}The set of hyperparameters $r_k=\{r_1,\ldots r_K\}$ and $p_k=\{p_1,\ldots p_K\}$ are considered equal between the components of the mixture and are taken equal to respectively  
$\nu_0$ and $s_k^2$.
The resulting posterior distributions over the parameters of the model are given as follows:
{\footnotesize
\begin{eqnarray}
 \bsmu_k|\bX, \bz, \bsSigma_k, H &\sim& \cN (\bsmu_n, \bsSigma_k/(n_k+\kappa_n)) \nonumber \\
 a_j|\bX, \bz, \lambda_k, \bD_k, H &\sim& \cI\cG (   \frac{n_k+\nu_k + Kd+1}{2},      \frac{\text{diag}(\sum_{k=1}^{K} \lambda_k^{-1} \bD_k^T (\frac{n_k \kappa_n}{n_k+\kappa_n} (\bar{\bx}_k - \bsmu_0) (\bar{\bx}_k - \bsmu_0)^T + W_k + \Lambda_k) \bD_k )}{2}  ) \nonumber \\
 \bD_k|\bX, \bz, H &\sim& \cI\cW ( n_k+\nu_k, \Lambda_k + W_k +  \frac{n_k \kappa_n}{n_k+\kappa_n} (\bar{\bx}_k - \bsmu_0) (\bar{\bx}_k - \bsmu_0)^T ) \nonumber \\
 \lambda_k|\bX, \bz, \bD_k, \bA_k, H &\sim& \cI\cG ( \frac{r_k+n_kd}{2}, \frac{p_k + \tr( \bD_k \bA^{-1} \bD_k^T ( \frac{n_k \kappa_n}{n_k+\kappa_n} (\bar{\bx}_k - \bsmu_0) (\bar{\bx}_k - \bsmu_0)^T + W_k + \Lambda_k ) )}{2}).\nonumber 
\end{eqnarray}
}

\paragraph{(9) Model $\lambda_k \bD_k \bA_k \bD_k^T$}
Finally, the more general model is the standard one with $\lambda_k \bD_k \bA_k \bD_k^T$ parametrization.
This model is also known as the full covariance model $\bsSigma_k$.
The volume $\lambda_k$, the orientation $\bD_k$, and the shape $\bA_k$ differ for each component of the mixture. 
In this situation, the prior density for the mean is normal and the one for the covariance matrix is an inverse Wishart, which leads to  the following conjugate normal inverse Wishart prior density:
{\small
\begin{eqnarray}
  \bsmu_k|\bsSigma_k &\sim& \cN(\bsmu_0, \bsSigma_k / \kappa_n) \ \forall k=1,\ldots,K \nonumber \\
  \bsSigma_k &\sim& \cI\cW (\nu_k, \Lambda_k)  \ \forall k=1,\ldots,K \nonumber 
\end{eqnarray}}where $(\bsmu_0,\kappa_n)$ and $(\nu_k, \Lambda_k)$ are respectively 
the hyperparamerets for respectively  normal prior density  over the mean  and  
the inverse Wishart prior density over the covariance matrix.
The resulting posterior over the model parameters $(\bsmu_1,\ldots,\bsmu_k, \bsSigma_1,\ldots,\bsSigma_k)$ is given as follows:
{\footnotesize
\begin{eqnarray}
  \bsSigma_k|\bX, \bz,H &\sim& \cI\cW ( n_k+\nu_k, \Lambda_k + W_k +  \frac{n_k \kappa_n}{n_k+\kappa_n} (\bar{\bx}_k - \bsmu_0) (\bar{\bx}_k - \bsmu_0)^T). \nonumber
\end{eqnarray}
}

\renewcommand{\baselinestretch}{1.3}
\normalsize
\setlength{\bibsep}{0.1pt}

\section*{References}
\bibliographystyle{apalike}
{
\bibliography{Refs}
}
\end{document}